\documentclass{article}

\usepackage{setspace}
\usepackage{microtype}
\usepackage{graphicx}
\usepackage{wrapfig}
\usepackage{subfig}
\usepackage{booktabs} 
\usepackage[hidelinks,colorlinks]{hyperref}
\usepackage{float}
\usepackage[ruled,linesnumbered,algo2e]{algorithm2e}
\usepackage{enumitem}

\usepackage{icml2024}

\usepackage{amsmath}
\usepackage{amssymb}
\usepackage{mathtools}
\usepackage{amsthm}
\usepackage{mathrsfs}

\usepackage{xcolor}
\usepackage[capitalize,noabbrev]{cleveref}
\newcommand{\bee}{\begin{equation}\begin{aligned}}
\newcommand{\ee}{\end{aligned}\end{equation}}
\newcommand\abs[1]{\left | #1 \right |}
\newcommand{\norm}[1]{\left\lVert#1\right\rVert}

\newcommand{\sbr}[1]{\left (#1\right )}
\newcommand{\mbr}[1]{\left [#1\right ]}
\newcommand{\lbr}[1]{\left \{#1\right\}}
\usepackage{tikz}
\newcommand*\circled[1]{\tikz[baseline=(char.base)]{
            \node[shape=circle,draw,inner sep=0.5pt] (char) {#1};}}

\usepackage{amsmath,amsfonts,bm}

\def\Tableref#1{Table~\ref{#1}}

\def\Lemref#1{Lemma~\ref{#1}}
\def\Thref#1{Theorem~\ref{#1}}

\def\eqref#1{equation (\ref{#1})}
\def\Eqref#1{Equation (\ref{#1})}

\def\Probref#1{Problem (\ref{#1})}

\def\1{\bm{1}}

\def\Tau{\mathcal{T}}

\def\vtheta{{\bm{\theta}}}
\def\vOmega{{\bm{\Omega}}}
\def\vDelta{{\bm{\Delta}}}

\def\vlam{{\bm{\lambda}}}
\def\vbeta{{\bm{\beta}}}
\def\vphi{{\bm{\phi}}}

\def\vf{{\bm{f}}}

\def\vh{{\bm{h}}}

\def\vv{{\bm{v}}}

\def\vx{{\bm{x}}}
\def\vy{{\bm{y}}}
\def\vz{{\bm{z}}}

\def\mJ{{\bm{J}}}

\def\mY{{\bm{Y}}}

\DeclareMathAlphabet{\mathsfit}{\encodingdefault}{\sfdefault}{m}{sl}
\SetMathAlphabet{\mathsfit}{bold}{\encodingdefault}{\sfdefault}{bx}{n}

\def\gB{{\mathcal{B}}}

\def\gH{{\mathcal{H}}}

\newcommand{\R}{\mathbb{R}}

\newcommand{\pf}{\mathcal{T}}
\newcommand{\mTheta}{ \mathbf{\Theta} }
\newcommand{\DomTheta}{ {{[0, \frac{\pi}{2}]}^{m-1}} }

\newcommand{\tb}{\textbf}

\newcommand{\sLam}{{\mathbf{\Lambda}}}

\DeclareMathOperator*{\argmin}{arg\,min}

\newtheorem{theorem}{Theorem}[section]
\newtheorem{proposition}[theorem]{Proposition}
\newtheorem{lemma}[theorem]{Lemma}
\newtheorem{corollary}[theorem]{Corollary}
\theoremstyle{definition}
\newtheorem{definition}[theorem]{Definition}
\newtheorem{ass}[theorem]{Assumption}

\theoremstyle{remark}
\newtheorem{remark}[theorem]{Remark}

\usepackage[textsize=tiny]{todonotes}

\def \ewidth{0.23}
\def \hwidth{0.15}
\def \qwidth{0.13}

\icmltitlerunning{UMOEA/D}
\begin{document}
\twocolumn[
\icmltitle{UMOEA/D: A Multiobjective Evolutionary Algorithm for Uniform Pareto Objectives based on Decomposition}
\icmlsetsymbol{equal}{*}
\begin{icmlauthorlist}
\icmlauthor{Xiaoyuan Zhang}{dep1}
\icmlauthor{Xi Lin}{dep1}
\icmlauthor{Yichi Zhang}{dep2}
\icmlauthor{Yifan Chen}{dep3}
\icmlauthor{Qingfu Zhang}{dep1}
\end{icmlauthorlist}

\icmlaffiliation{dep1}{Department of Computer Science, City University of Hong Kong}
\icmlaffiliation{dep2}{Department of Biostatistics and Bioinformatics, Duke University}
\icmlaffiliation{dep3}{Departments of Computer Science and Mathematics, Hong Kong Baptist University}


\vskip 0.3in
]

\printAffiliationsAndNotice{}  

\begin{abstract}
    Multiobjective optimization (MOO) is prevalent in numerous applications, in which a Pareto front (PF) is constructed to display optima under various preferences. Previous methods commonly utilize the set of Pareto objectives (particles on the PF) to represent the entire PF. However, the empirical distribution of the Pareto objectives on the PF is rarely studied, which implicitly impedes the generation of diverse and representative Pareto objectives in previous methods. To bridge the gap, we suggest in this paper constructing \emph{uniformly distributed} Pareto objectives on the PF, so as to alleviate the limited diversity found in previous MOO approaches. We are the first to formally define the concept of ``uniformity" for an MOO problem. We optimize the maximal minimal distances on the Pareto front using a neural network, resulting in both asymptotically and non-asymptotically uniform Pareto objectives. Our proposed method is validated through experiments on real-world and synthetic problems, which demonstrates the efficacy in generating high-quality uniform Pareto objectives and the encouraging performance exceeding existing state-of-the-art methods. 
    The detailed model implementation and the code are scheduled to be open-sourced upon publication.
\end{abstract}

\section{Introduction}
\begin{figure}[t]
    \centering
    \begin{subfloat}[\scriptsize MOEA/D on RE21 (obj=2)]{\includegraphics[width=\ewidth \textwidth]{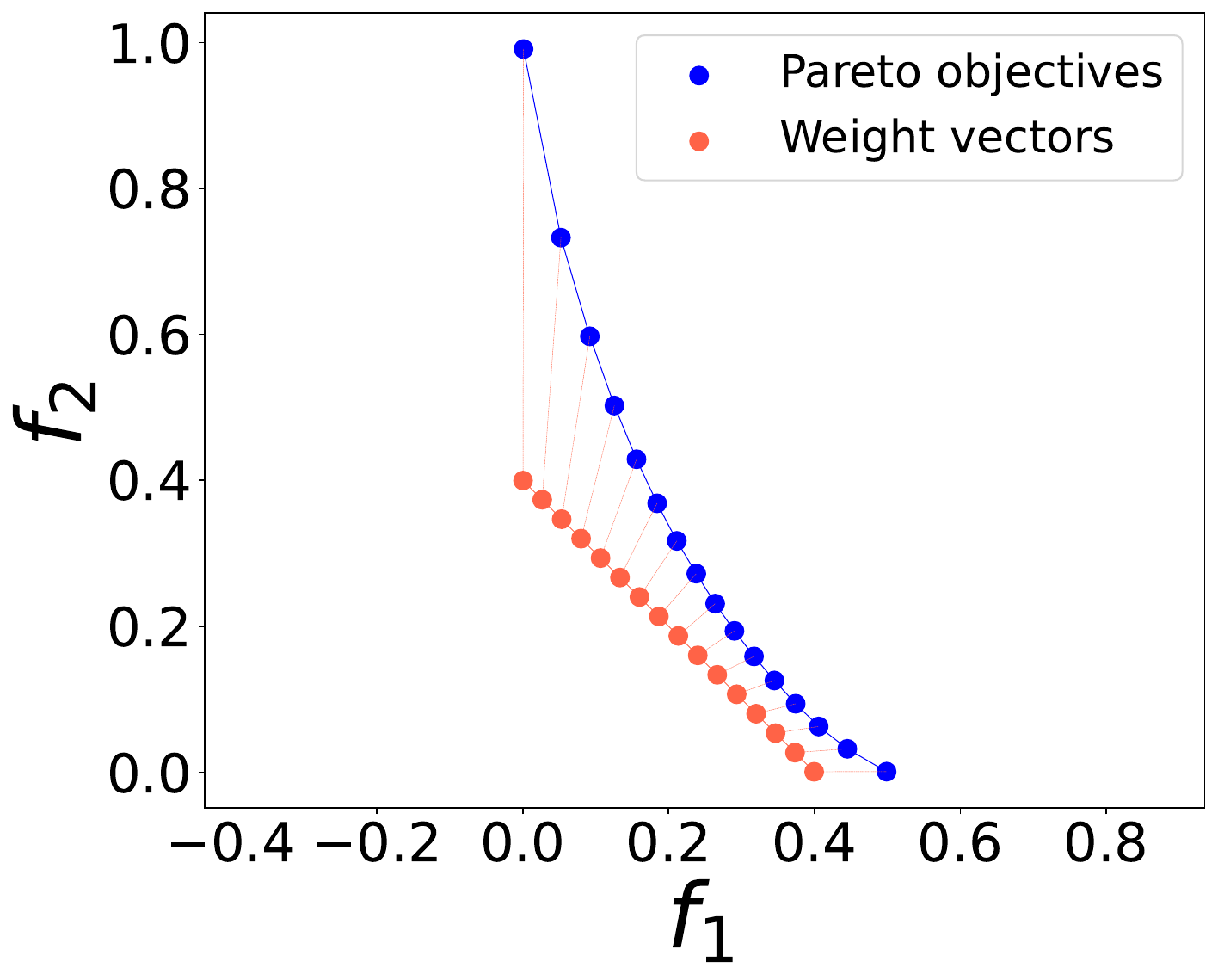}}
    \end{subfloat}
    \hfill
    \begin{subfloat}[\scriptsize UMOEA/D on RE21]{\includegraphics[width=\ewidth \textwidth]{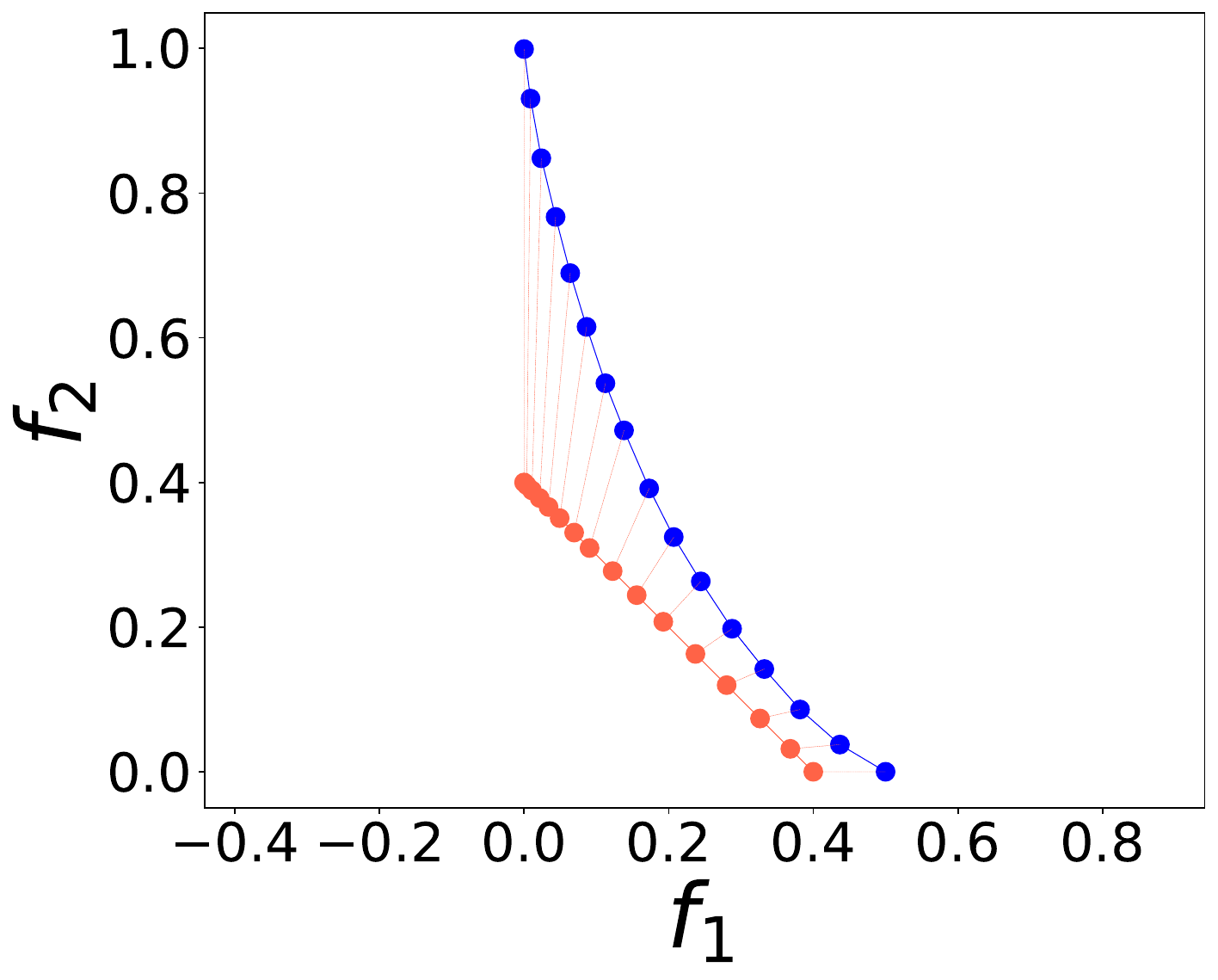}}
    \end{subfloat} \\
    \vspace{-10pt}
    \begin{subfloat}[\scriptsize MOEA/D on RE41 (obj=4)]{\includegraphics[width=\ewidth \textwidth]{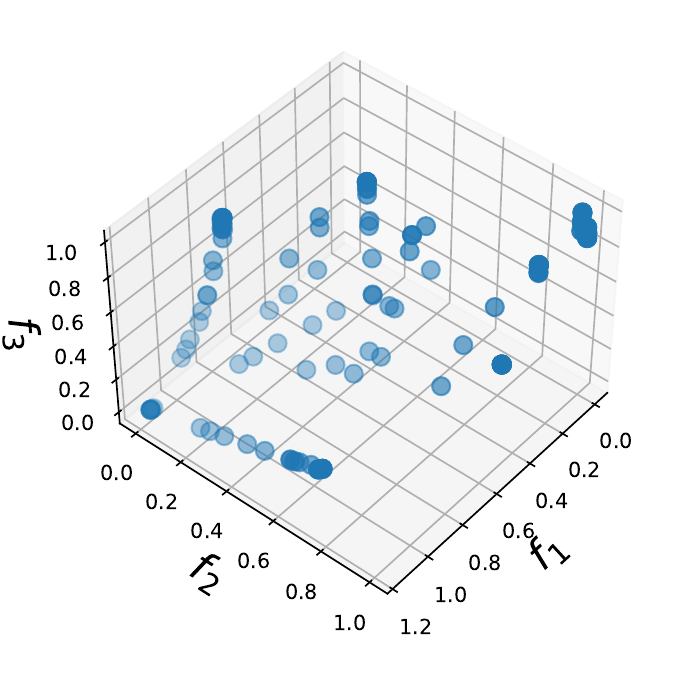}}
    \end{subfloat}
    \hfill
    \begin{subfloat}[\scriptsize UMOEA/D on RE41]{\includegraphics[width=\ewidth \textwidth]{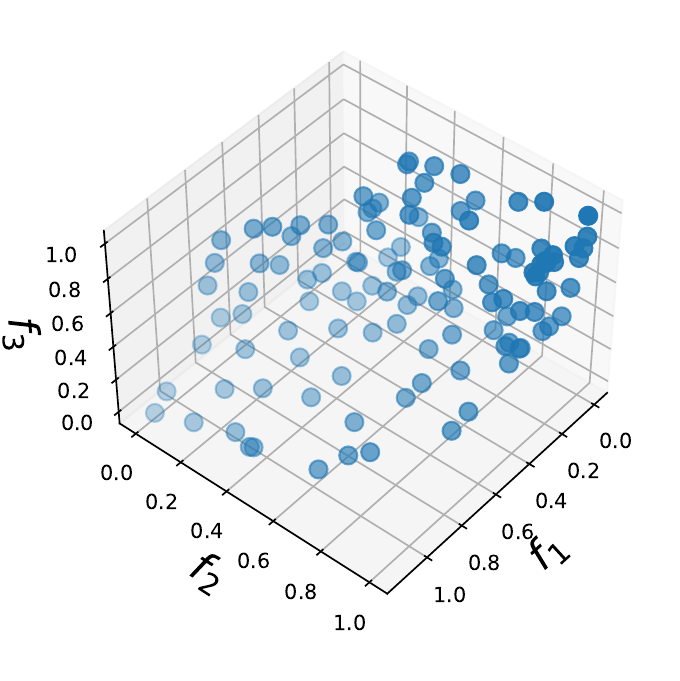}}
    \end{subfloat}
    \caption{(a)/(c): MOEA/D produces non-uniform and potentially duplicated solutions. 
    (b)/(d): UMOEA/D generates uniform objectives on the PF by searching for a set of weight vectors. RE41 is a 4-objective problem, and its objective space is projected onto the first three objectives for visualization. }
    \label{fig:illus_uniform}
\end{figure}

Real-world applications such as trustworthy machine learning \cite{zhao2022inherent,liang2021pareto}, autonomous agent planning \cite{xu2020prediction,hayes2022practical}, and industrial design \cite{schulz2017interactive,wang2011multi,tanabe2020easy,xu2021multi} often involve multiobjective optimization (MOO) problems. The most important insight from MOO is that due to the conflicting nature between different objectives, simultaneously achieving optima for all objectives is difficult. For example, for a fairness classification system, it is essential for a decision maker to balance a trade-off between conflicting objectives, e.g., accuracy and fairness, to make a trustworthy decision \cite{zhao2022inherent,ruchte2021scalable,xian2023fair}.  

The concept of Pareto optimality is thereby introduced to resolve such conflict. A solution $\vx$ is called Pareto optimal if it cannot be dominated by any other solution $\vx'$ in the decision space: more specifically, we will have if $f_i(\vx) \leq f_i(\vx')$, for some $i \in [m]$, then there must exist $j \in [m]$ such that $f_j(\vx) > f_j(\vx')$; Here $\vf(\vx)$ is an $m$-objective vector function and $[m]=\{1,\ldots,m\}$ \cite{miettinen1999nonlinear,ehrgott2005multicriteria}. An illustrative example of Pareto objectives for a two-objective problem can be seen in \Cref{fig:illus_uniform}(a) or (b), where one objective (e.g. $f_1$) cannot be decreased without deteriorating the other objective (e.g. $f_2$). The collection of all Pareto solutions is dubbed as the Pareto set (PS), and its corresponding objectives form the PF. 

In the past few decades, a large amount of MOO algorithms \cite{zhang2007moea,zhang2008rm,li2014evolutionary,lin2019pareto,liu2021profiling,bernreuther2023multiobjective} have been proposed for constructing a finite set of solutions (termed as a ``population'' in MOO) to approximate the PF. Among them, multiobjective evolutionary algorithms (MOEAs) are the most popular due to their capability to avoid poor local optima and obtain a set of solutions in a single run \cite{pymoo, caramia2020multi}.

\subsection{Challenges for constructing uniform PF}

One pursuit for contemporary MOEAs research is to efficiently generate Pareto objectives \emph{uniformly} distributed on the PF; which are considered effectively capturing the entire PF since they represent diverse optimal trade-offs among multiple objectives.

To achieve that, a flurry of efforts have been made to generate uniform weight vectors \cite{ref_dirs_energy,liu2021tri,das1998normal,elarbi2019approximating} aiming to approximate uniform Pareto objectives consequently. However, it has been noted in recent work~\citep{rhee2017space,liu2021profiling}, that uniform weights do not necessarily yield uniform objectives. Illustrative examples are shown in \Cref{fig:illus_uniform}, where Figures~1(a) and (c) on the left reflect MOEA/D (multiobjective evolutionary algorithm based on decomposition) \cite{zhang2007moea} adopts uniform weight vectors but obtains non-uniform and possibly duplicate Pareto objectives. (In contrast, our method, as shown in Figures~1(b) and (d), manages to generate uniform Pareto objectives for both a two-objective and a four-objective problem.)

A straightforward way to address the issue of non-uniform Pareto objectives is to search for a set of weight factors that corresponding to uniform Pareto objectives.
The first notable work in this line is MOEA/D-AWA~\citep[MOEA with adaptive weight adjustment by decomposition]{qi2014moea}. Subsequently, numerous follow-up works developed similar weight adjustment algorithms~\citep{de2018moea, siwei2011multiobjective, wu2017adaptive, dong2020moea, jiao2021two}. 
In general, these methods employ simple models to estimate the underlying PF \cite{wu2017adaptive, dong2020moea} and use heuristic strategies to adjust weights, such as removing the most crowded solutions and adding the most sparse ones \cite{de2018moea}. However, those methods perform poorly in certain scenarios due to the reliance on heuristic strategies.

\subsection{Our contributions}

In this paper, we propose UMOEA/D, the first principled method to generate \underline{Uniform} Pareto objectives under the \underline{MOEA/D} (D denotes decomposition) framework \cite{zhang2007moea}. 
We propose formal definitions of ``uniformity'' on the PF. We demonstrate that maximizing the minimal pairwise distances in a finite set yields both asymptotic and non-asymptotic uniformity guarantees. 
To numerically strength the search process, we employ a neural network to approximate the shape of the PF. 
The empirical results show that our method can generate high-quality uniform Pareto objectives in a reasonable time for both synthetic and real-world industrial design problems with a large number of locally optimal solutions. Those problems are challenging to gradient-based MOO methods \cite{mahapatra2020multi, liu2021profiling}. UMOEA/D outperforms state-of-the-art methods in terms of both solution quality and running time.

In summary, the contribution of this paper is three-fold. 
\begin{enumerate}[itemsep=-0.2em, topsep=0.0em, leftmargin=1.0em]
    \item We study the deficiency of MOEA/D in achieving uniform Pareto objectives, which rooted in the non-linearity of the ``weight-to-objective'' function $\vh(\cdot)$ and investigate the characteristics of this function.
    
    \item We propose UMOEA/D, an NN-based scheme to obtain uniform Pareto objectives by modeling $\vh(\cdot)$ to minimize a rigorously defined uniformity score,
    along with provable asymptotic and non-asymptotic results.

    \item We conducted extensive experiments on both synthetic and real-world MOO problems. Results demonstrate that UMOEA/D outperforms state-of-the-art methods in terms of both the uniformity of solutions and runtime efficiency.
\end{enumerate}

\section{Multiobjective preliminaries} \label{sec:pre:moo}
In this section, we provide a concise overview of key concepts in MOO. For clarity, throughout this paper, bold lower letters denote vectors (e.g. $\vv$), bold upper letters denote a set of vectors (e.g., $\mY$). A multiobjective problem (MOP) of $m$ conflicting objectives can be expressed as:
\begin{equation} \label{eqn:original}
    \min_{\vx \in \mathscr{X} \subset \mathbb{R}^n} \vf(\vx)=\left( f_1(\vx), \ldots, f_m(\vx) \right),
\end{equation}
where the $\min$ operator is formally reloaded and different from the minimization operator in the scalar case.

Notably, for an MOO problem, it is difficult to compare two solutions and the concepts of domination and Pareto optimal solutions are thereby introduced.
We say a solution $\vx^{(a)}$ \textbf{dominates} $\vx^{(b)}$ if there exists an index $i \in [m]$ such that $f_i(\vx^{(a)}) < f_i(\vx^{(b)})$ and $\forall j \in [m] \setminus \{i\}, f_j(\vx^{(a)}) \leq f_j(\vx^{(b)})$. A solution $\vx$ is a \textbf{Pareto solution} if no other solution $\vx^\prime \in \mathscr{X}$ dominates it. The set of all Pareto solutions is denoted as the \textbf{Pareto set} $\mathtt{PS}$, and its image set $\mathcal{T}$, $\mathcal{T} = (f \circ \mathtt{PS})$ is called the \textbf{ Pareto front} (PF). The dominance of a solution $\vx^{(a)}$ over another solution $\vx^{(b)}$ implies that $\vx^{(a)}$ is strictly better than $\vx^{(b)}$, which indicates that $\vx^{(b)}$ can be dropped in MOO. For a Pareto solution $\vx$, $\vy=\vf(\vx)\in \mathbb{R}^m$ is called a \textbf{Pareto objective}. $\delta$-dominance \cite{zuluaga2016varepsilon} extends the concept of dominance, where the solution $\vx^{(a)}$ $\delta$-dominates $\vx^{(b)}$ if $(\vf(\vx^{(a)}) - \delta)$ dominates $\vf(\vx^{(b)})$.
Additionally, a solution $\vx$ is called \textbf{weakly Pareto optimal} if it cannot be strictly dominated by other solutions ($f_i(x') < f_i(x)$ for all $x' \in \mathscr{X}$).

It is very difficult to optimize all $m$ objectives directly due to their conflicting nature. A more practical approach is to convert the objective vector $\vf(\vx)$ into a single objective problem through an aggregation function $g(\cdot, \vlam)$ with a specific weight $\vlam$. Many aggregation methods have been proposed in the past few decades, and this work focuses on the modified Tchebycheff (``mtche'' in short) aggregation function ~\cite{ma2017tchebycheff}:
\begin{align}
    g^{\text{mtche}}(\vy, \vlam) = \max_{i \in [m]} \sbr{ \frac{y_i - z_i}{\lambda_i} }: \quad \vDelta_{m-1} \mapsto \R^n,
    \label{eqn:g-mtche}
\end{align}
where $\vDelta_{m-1}$ denotes the $m$-1 weight simplex, defined as $\vDelta_{m-1}=\{\vy | \sum_{i=1}^m y_i=1, \; y_i \geq 0. \}$. Additionally, let $\vz$ be an ideal point such that $z_i \leq f_i(\vx)$ for all $\vx \in \mathscr{X}$.

\begin{wrapfigure}{R}{0.26 \textwidth}
    \centering
    \vspace{-15pt}
    \includegraphics[width=0.26 \textwidth]{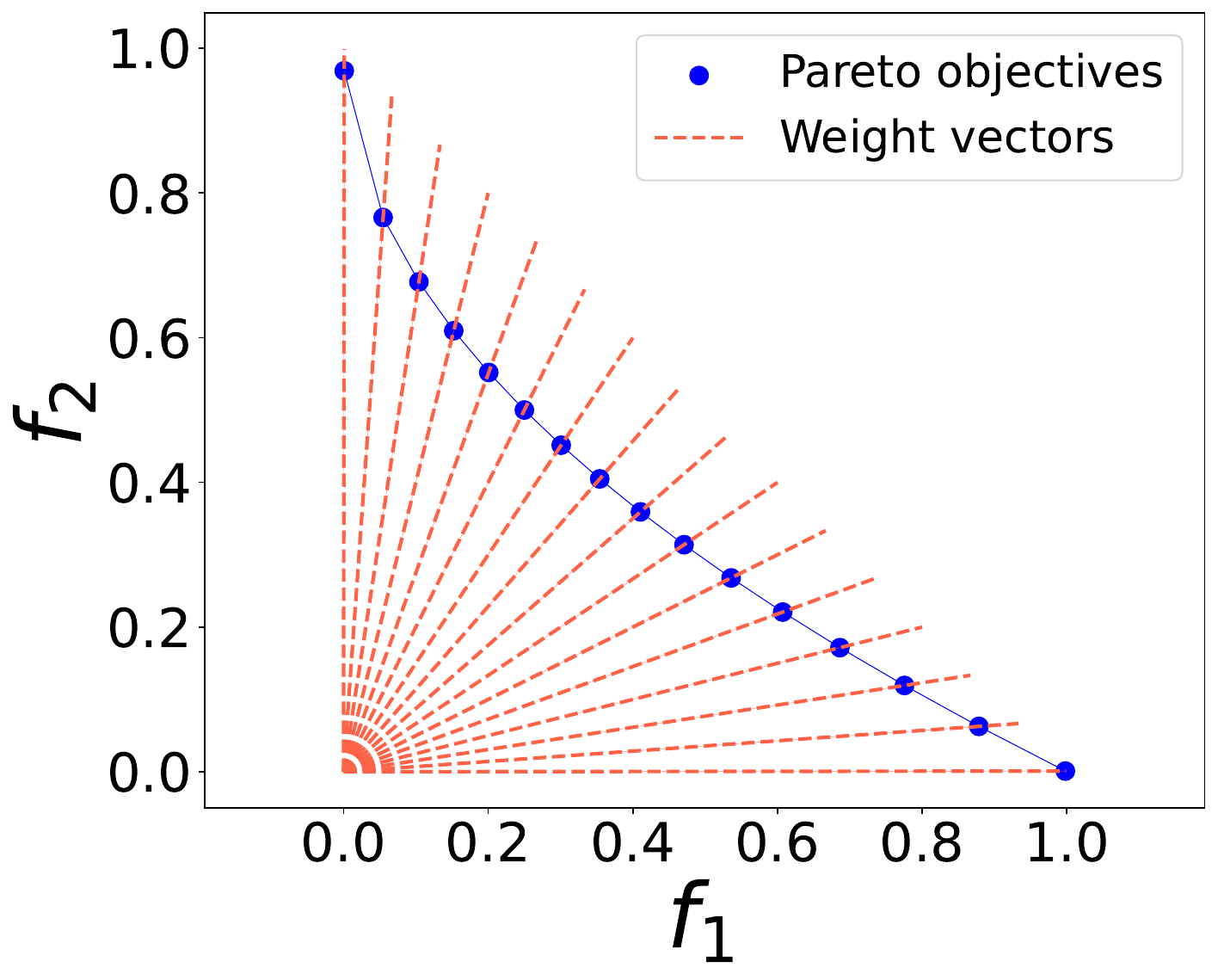}
    \caption{Exact Pareto solutions are intersections between PF and weight vectors.}
    \label{fig:illus_exact}
    \vspace{-5pt}
\end{wrapfigure}

One attractive property of this aggregation function is that minimizing it produces $\vlam$-exact Pareto solutions \cite{mahapatra2020multi} under mild conditions \cite{zhang2023hypervolume}, namely, for any given $\vlam$, the optimal Pareto objective $\vy^*$, i.e., the minimize of in equation (\ref{eqn:g-mtche}), follows the pattern $\sbr{\frac{y_i^* - z_i}{\lambda_i}} = C,$ for all $i \in [m]$, where $C$ is a positive constant (vector $\vy - \vz$ is parallel to $\vlam$).

\section{Related work} \label{app:sec:related_work}
This section reviews two closely related lines of research. The first is Pareto set learning, which utilizes a single neural model to represent the entire Pareto set. The second is gradient-based methods that aim to find a set of Pareto optimal solutions. 

\subsection{Pareto set learning} \label{app:sec:psl}
Pareto set learning (PSL) aims to learn the entire Pareto set through a single neural model $x_\vbeta(\cdot): \vDelta_{m-1} \mapsto \mathbb{R}^n$, where $\vbeta$ are the trainable vectors \cite{navon2020learning, lin2022pareto,chen2022multi}. A PSL model $x_\vbeta(\cdot)$ is typically trained by optimizing the following loss function:
\begin{equation}\label{eqn:psl}
    \min_\vbeta \; \mathtt{psl\_loss}(\vbeta) = \min_\vbeta \mathbb{E}_{\vlam \sim \text{Unif}( \vDelta_{m-1} )}  \left [ g( \vf \circ x_\vbeta(\vlam), \vlam) \right ],
\end{equation}
where $g(\cdot)$ is some aggregation functions. The gradient involved in \Cref{eqn:psl} is computed by the chain rule: $\nabla_\vbeta \; \mathtt{psl\_loss}(\vbeta) = \mathbb{E}_{\vlam \sim \text{Unif}( \vDelta_{m-1} )} \frac{\partial g}{\partial \vf} \frac{\partial \vf}{\partial \vx} \frac{\partial \vx}{\partial \vbeta}$. Since \Cref{eqn:psl} is trained by stochastic gradient descent. Previous PSL methods could fail to find the globally optimal model $x_\vbeta(\cdot)$ when the objective $\vf(\cdot)$ has many local optimas. Another shortcoming of PSL is that the model can become excessively large when the decision space $\mathbb{R}^n$ is large, despite the effective dimension of a Pareto set being only $m$-1 \cite{zhang2008rm}. This results in inefficient utilization of the model.

\subsection{Gradient-based MOO methods} \label{sec:rw:grad}
Gradient-based MOO methods are efficient for optimizing conflicting objectives by utilizing gradient information. However, these methods may get stuck in local Pareto solutions, also known as Pareto stationary solutions. Some notable works in this area include:
\begin{enumerate}
    \item MOO-SVGD \cite{liu2021profiling}: This approach employs Stein Variational Gradient Descent to obtain a diverse set of Pareto solutions.
    \item EPO \cite{mahapatra2020multi}, COSMOS \cite{ruchte2021scalable}, WC-MGDA \cite{momma2022multi}, and OPT-in-Pareto \cite{ye2022pareto}: These methods focus on finding a single Pareto solution that satisfies specific user requirements.
\end{enumerate}
However, gradient-based MOO methods often struggle to produce globally optimal solutions. For instance, in the ZDT4 testing problem \cite{deb2006multi}, numerous local optimal solutions are present, making it challenging to identify them using gradient information (cf \Cref{sec:gradmoo} for a concrete example). The proposed method is an evolutionary-based approach that does not rely on gradient information to directly optimize objective functions. This characteristic makes it more suitable for finding global optimal solutions and avoids the limitations of gradient-based methods.

\section{The proposed UMOEA/D method}
In this section, we introduce UMOEA/D, a method for recovering uniform Pareto objectives within the MOEA/D framework. In the MOEA/D framework, decision makers choose a set of diverse weight vectors with the goal of obtaining a diverse set of weight vectors. Various approaches have been proposed to generate diverse weight vectors \cite{deb2019generating,blank2020generating,das1998normal,tan2013moea}. A key observation in this paper is that, despite the diversity of the chosen weight vectors, there exists a \emph{bottleneck} towards achieving the uniform Pareto objectives that decision makers are truly interested in. In the next subsection, we will study the underlying weight-to-objective function, which represents the bottleneck in achieving the desired uniform Pareto objectives. 

\subsection{Weight-to-objective function and induced Pareto objective distribution} \label{sec_41}
We study the weight-to-objective function with the `mtche' aggregation function (\Cref{eqn:g-mtche}). $\vh(\vlam): \vOmega \mapsto \mathbb{R}^m$, which maps a weight $\vlam$ from a weight space to objective space. Here $\vh(\vlam)$ is defined as. 

\begin{equation}\label{eqn:h}
    \vy = \vh(\vlam) = \arg \min_{\vy' \in \mathscr{Y}} g^{\text{mtche}}(\vy', \vlam).
\end{equation}
In \Cref{eqn:h}, $\mathscr{Y}$ represents the full feasible objective space, which is the result of mapping the decision space $\mathscr{X}$ through the vector mapping function $\vf$. $\mathscr{Y} = \{\vf(x) | x \in \mathscr{X}\}$. 
To ensure the well-definedness of $\vh(\vlam)$, the optimal solution $\vy$ that minimizes the problem in \Cref{eqn:h} must be unique. This uniqueness can be achieved by utilizing various methods, including selecting the optimal solution based on lexicographic order. For simplicity, we assume that the original MOP (\Cref{eqn:original}) does not have any weakly Pareto optimal solutions, which is commonly adopted \cite{roy2023optimization,bergou2021complexity}. Under this assumption, the optimal solution is proven to be unique (cf \Cref{cor:unique} in \Cref{app:sec:moo}).
We next build the diffeomorphism of $\vh(\vlam)$ under the following `full mapping' assumption. 
\begin{ass}[Full mapping] \label{ass:full}
    For each $\vlam$, there exist a scalar $k$, such that $k \vlam + \vz$ is a Pareto objective for the original MOP (\Cref{eqn:original}), where $\vz$ is a reference point used in \Cref{eqn:g-mtche}. 
\end{ass}
This assumption means that for any user preference $\vlam$, there exist a Pareto objective at the ray of vector $\vlam$, which is a common case for example such as famous DTLZ 1-4 testing problems \cite{deb2002scalable}. This assumption directly yields the diffeomorphism of $\vh(\vlam)$ when objectives $f_i$'s are differentiable. 
\begin{lemma}[Diffeomorphism of $\vh(\vlam)$] \label{lem:diff}
    Under the \Cref{ass:full}, if the objectives $f_i$'s are differentiable, then the weight-to-objective function $\vy = \vh(\vlam)$ is a diffeomorphism.
\end{lemma}
The proof is left in \Cref{sec:missing}. Now we are ready to describe the distribution on the PF. We let $\mathbf{\Lambda}_N = \{\vlam^{(1)},\ldots,\vlam^{(N)}\}$ to represent a uniform weight set solving the following weight generating problem as proposed in \cite{blank2020generating},
\begin{align*}
   \max_{\sLam_N \subset \vOmega} \min_{ \vlam^{(i)}, \vlam^{(j)} \in \sLam_N} \rho(\vlam^{(i)}, \vlam^{(j)}).
\end{align*}
The asymptotic Pareto objectives distribution is given by the following theorem:

\begin{theorem} \label{thm:obj_dist}
    The category distribution $\widetilde{\mY}_N$ over a set of Pareto objectives $\mY_N = \{\vy^{(1)}, \ldots, \vy^{(N)}\}$ converges in distribution to $\vh \circ \text{Unif}(\vOmega)$, denoted as $\widetilde{\mY}_N \xrightarrow{\text{d}} \vh \circ \text{Unif}(\vOmega)$, where $\vy^{(i)} = \vh(\vlam^{(i)})$, and  the probability density function (pdf) of $\widetilde{\mY}_N$ converges to $\frac{1}{A} \text{Det}(\mJ(\vh^{-1}))$. Here $A = \gH_{m-1}(\vOmega)$ represents the Hausdorff measure for a $(m-1)$-dimensional manifold, and $\mJ$ is the Jacobian matrix
\end{theorem}
The term $\text{Unif}(\vOmega)$ denotes the uniform distribution over the set $\vOmega$. The proof is in \Cref{app:sec:proof_1}.

We observe from \Cref{thm:obj_dist} that generating uniform Pareto objectives is challenging due to the non-linearity of the function $\vh(\vlam)$. To illustrate this point, we present concrete results for widely used multiobjective ZDT \cite{deb2006multi} and DTLZ \cite{deb2002fast} problems when the weight space is selected as simplexes.

\begin{enumerate}
    \item ZDT1: $y_1 = k \lambda_1$, $y_2 = k (1 - \sqrt{\lambda_1})$, where $k = \frac{2-\lambda_1 - \sqrt{-3\lambda_1^2+4 \lambda_1}}{2{(\lambda_1-1)}^2}$.
    \item ZDT2: $y_1 = k \lambda_1$, $y_2 = k (1-\lambda_1^2)$, where $k = \frac{\lambda_1-1 + \sqrt{5 \lambda_1^2-2\lambda_1+1}}{2\lambda_1^2}$.
    \item DTLZ1: $f_i = 0.5 \lambda_i$, for $i=1,2,3$.
\end{enumerate}

From these examples, we observe that uniformity in the weight space does not always imply uniformity in the Pareto objective space, except in specific circumstances when $\vh(\vlam)$ is an affine mapping, as is the case for DTLZ1.

\subsection{UMOEA/D framework} \label{sec:mms}
As discussed in the previous section, achieving a uniform Pareto objective is challenging due to the non-linearity of the weight-to-objective function $\vh(\vlam)$. To attain uniformity on the PF, two key steps need to be taken. First, it is necessary to formally define what is ``uniformity" for a MOP. Second, it is crucial to model $\vh(\vlam)$, and therefore, we are able find those particular weight vectors resulting in uniform Pareto objectives. 

We begin by analyzing the concept of uniformity in the context of multiobjective optimization problems. We provide two precise definitions of uniformity, starting with the asymptotic definition:

\begin{definition}[Asymptotic uniformity on PF, adapted from \citep{borodachov2007asymptotics}[Eq. 1.10]]
We say a serials of set $\mY_N$ to be asymptotically uniform on $\pf$ (the PF) with respect to the Hausdorff measure $\mathcal{H}_{m-1}$ if, for any subset $\gB \subset \pf$ with a boundary having a $\gH_{m-1}$-measure of zero, the following condition holds:
\begin{equation}
    \lim_{N \rightarrow \infty} \frac{\mathtt{Card}(\mY_N \cap \gB)}{\mathtt{Card}(\mY_N)} = \frac{\mathcal{H}_{m-1}(\mathcal{B})}{\mathcal{H}_{m-1}(\pf)},
\end{equation}
where $\mathtt{Card}$ represents the cardinality function. For example, $\mathtt{Card}(\mY_N \cap \gB)$ indicates the number of points in $\mY_N$ that belong to $\gB$.
\end{definition}

Intuitively, this definition implies that as the number of solutions in set $\mY_N$ tends to infinity, for any subset $\gB$ with a boundary having a $\gH_{m-1}$-measure of zero, the ratio of solutions lying in set $\gB$ is equal to the ratio of the cardinality of $\gB$ to the cardinality of the entire PF. In other words, it suggests that the solutions in $\mY_N$ must be uniformly distributed on the PF. The second definition is for the non-aymtotic case, 
\begin{definition}[Non-asymptotic uniformity on PF]
    We say a set $\mY_N$ is non-asymptotically uniform on $\pf$ with a parameter $\delta > 0$ when for any Pareto objective $\vy \in \pf$, there exist a objective $\vy' \in \mY_N$ such that, $\rho(\vy, \vy') \leq \delta$. 
\end{definition}
The non-asymptotic definition implies that for an arbitrary Pareto objective $\vy$, there exists a solution $\vy'$ in the discrete set $\mY_N$ such that $\vy'$ is an approximation of $\vy$ with an error of at most $\delta$. When the value of $\delta$ is small, the set $\mY_N$ serves as a suitable representation of the PF.

To generate uniformly distributed Pareto objectives, we propose to construct the objective configuration $\mY_N$ by solving the Maximal-Manifold-Separation (MMS) problem on the PF,
\begin{equation} \label{eqn:pack_pf}
    \max_{ \mY_N } \delta = \max_{ \mY_N } \left( \min_{ \vy^{(i) } \neq \vy^{(j)} \in \mY_N \subset \mathcal{T} } \rho(\vy^{(i)}, \vy^{(j)}) \right).
\end{equation}

Intuitively, \Cref{eqn:pack_pf} maximizes the minimum pairwise distances among all Pareto objectives. This objective ensures that densely populated solutions are spread apart, leading to a uniform distribution of Pareto objectives. We denote the optimal configuration of \Cref{eqn:pack_pf} as $\mY_N^*$ with a separation parameter of $\delta^*$. $\mY_N^*$ exhibits the following attractive uniform properties.

\begin{proposition} \label{prop:uniform}
    \circled{1} Under mild conditions, $\mY_N^*$ is non-asymptotically uniform on $\pf$ with a parameter of $\delta^*$. \circled{2} $\mY_N^*$ is asymptotically uniform on $\pf$ when $\pf$ is both compact and connected.
\end{proposition}

\begin{figure}[]
  \centering
  \includegraphics[width=0.5 \textwidth]{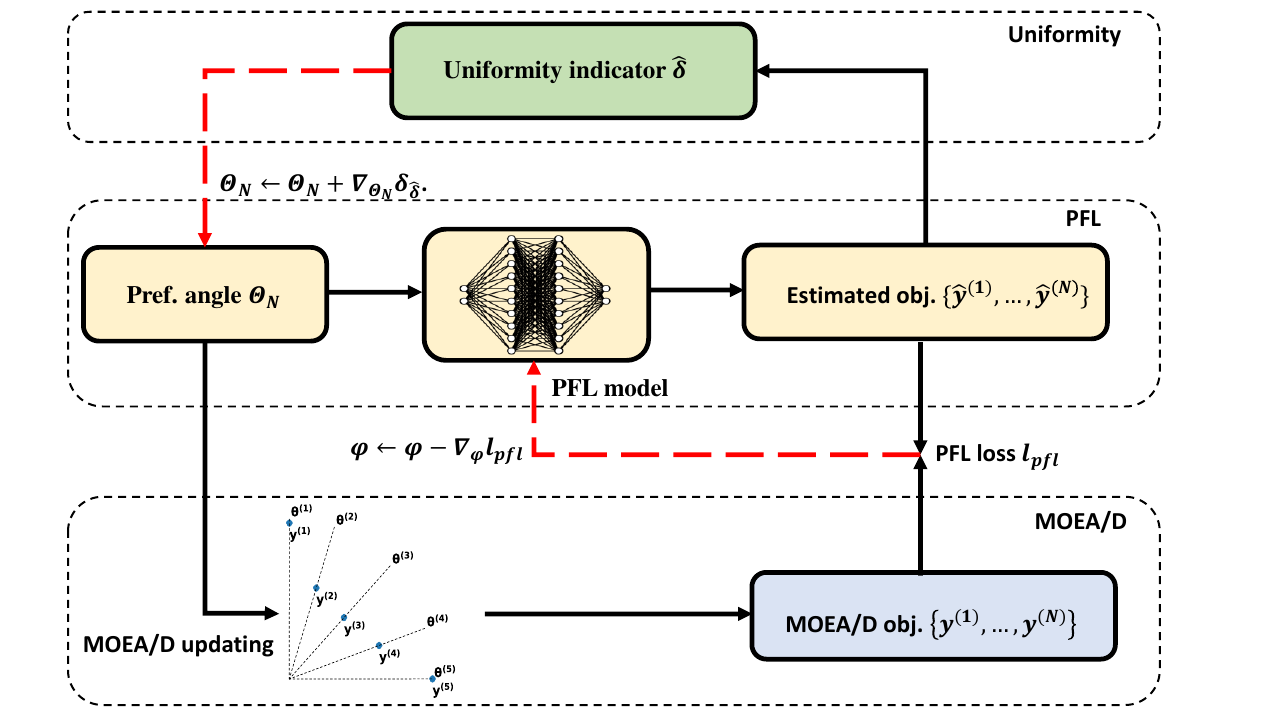}
  \caption{UMOEA/D consists of three parts: training the PFL model, updating new weights and MOEA/D. Firstly, Pareto optimal solutions are solved under initial weight vectors. Then, we use those weight vector and Pareto objective pairs to train a PFL model. Finally, new weight vectors are optimized by this model for achieving the most uniform Pareto objectives. Those three steps are performed iteratively.}
  \label{fig:framework}
\end{figure}

Please refer to \Cref{sec:uniform_proof} for the conditions and missing proofs. After those definitions, we introduce a compact framework for searching for uniform Pareto objectives in the MOEA/D framework. Since the true PF is unknown, we iteratively estimates the PF by learning the mapping from weight vectors to Pareto objectives. With the estimated PF, we then select weight vectors that exhibit the highest level of uniformity in their corresponding Pareto objectives. 
The framework comprises multiple components, and the computation flow is depicted in Figure \ref{fig:framework}.

\circled{1} We represent weight vectors using a set of weight angles $\mTheta_N = [\vtheta^{(1)}, \ldots, \vtheta^{(N)}]$. They are converted to each other by \Cref{app:eqn:pref_angle} and \Cref{app:eqn:pref_angle_inverse} in Appendix. One advantage of using weight angles is the simple constraint shape of each $\vtheta^{(i)} \in {[0, \frac{\pi}{2}]}^{m-1}$. In contrast, a weight vector is constrained on a simplex, which is more challenging to handle. A MOEA/D frame accept a set of weight angles and optimize for the Pareto objectives $\mY_N$. 
\circled{2} The Pareto Front Learning (PFL) module is then trained using the true objectives obtained from the output of MOEA/D. 
\circled{3} Subsequently, the weight angles are updated for achieving the most uniform Pareto objectives. 

More detailed descriptions of \circled{2} the PFL model and \circled{3} the weight update components are provided separately in the subsequent sections.
The practical algorithm is implemented as Algorithm \ref{alg_1} and \ref{alg_2} in \Cref{app:sec:alg}, where we also present time complexity analysis.

\subsection{Pareto front learning (PFL)} \label{sec:pfl}

\begin{table}
\caption{Comparison of model size between PSL and PFL. Note that for practical problems, decision variables $n$ is usually much larger than $m$.
} \label{tab:psl_pfl}
\centering
    \begin{tabular}{ll}
    \toprule
      Method  & Mapping function\\
      \midrule
    Pareto Set Learning (PSL) & $\vDelta_{m-1} \mapsto \mathbb{R}^{n}$     \\ 
    Pareto Front Learning (PFL) & ${[0, \frac{\pi}{2}]}^{m-1} \mapsto \mathbb{R}^m$      \\
    \bottomrule
    \end{tabular}
\end{table}

The PFL model, $\vh_\vphi(\cdot): \DomTheta \mapsto \R^m$, approximates the weight-to-objective function $\vh(\vlam)$ discussed in Section 4.1. It predicts the Pareto objective optimized by the modified Tchebycheff function under a specific weight angle $\vtheta$. $\vh_\vphi(\cdot)$ is trained by minimizing the mean square error (MSE) loss between the true Pareto objectives $\vy$ from MOEA/D and the estimated objectives $\vh_\vphi(\vtheta)$. In practice, the training time of the PFL model is fast (less than \texttt{1s}) and can be neglected compared to other operations in a multiobjective evolutionary algorithm.

We emphasize the necessity of introducing $\vh_\vphi(\cdot)$ rather than simply applying previous Pareto set learning methods, which learn the mapping from weight space to solution space $\R^n$ \cite{navon2020learning,lin2022pareto}. \circled{1} PSL simply uses gradient-based methods to optimize the PSL objective function defined in \Eqref{eqn:psl}. 
The induced locally optimal solutions make PSL fail on most MOEA benchmarks like ZDT, DTLZ problems. 
\circled{2} In an MOO problem, the number of decision variables $n$ can be arbitrarily large, while the efficient dimension of a Pareto set is only $m$-1 \cite{zhang2008rm}. PSL fails to detect it, resulting in the output of an $n$-D solution, leading to potentially large PSL model size. In contrast, the PFL model is constrained in the function space ${[0, \frac{\pi}{2}]}^{m-1} \mapsto \mathbb{R}^m$, which implies its complexity is independent of $n$.


\begin{wrapfigure}{R}{0.25\textwidth}
  \footnotesize
  \centering
  \includegraphics[width=0.25\textwidth]{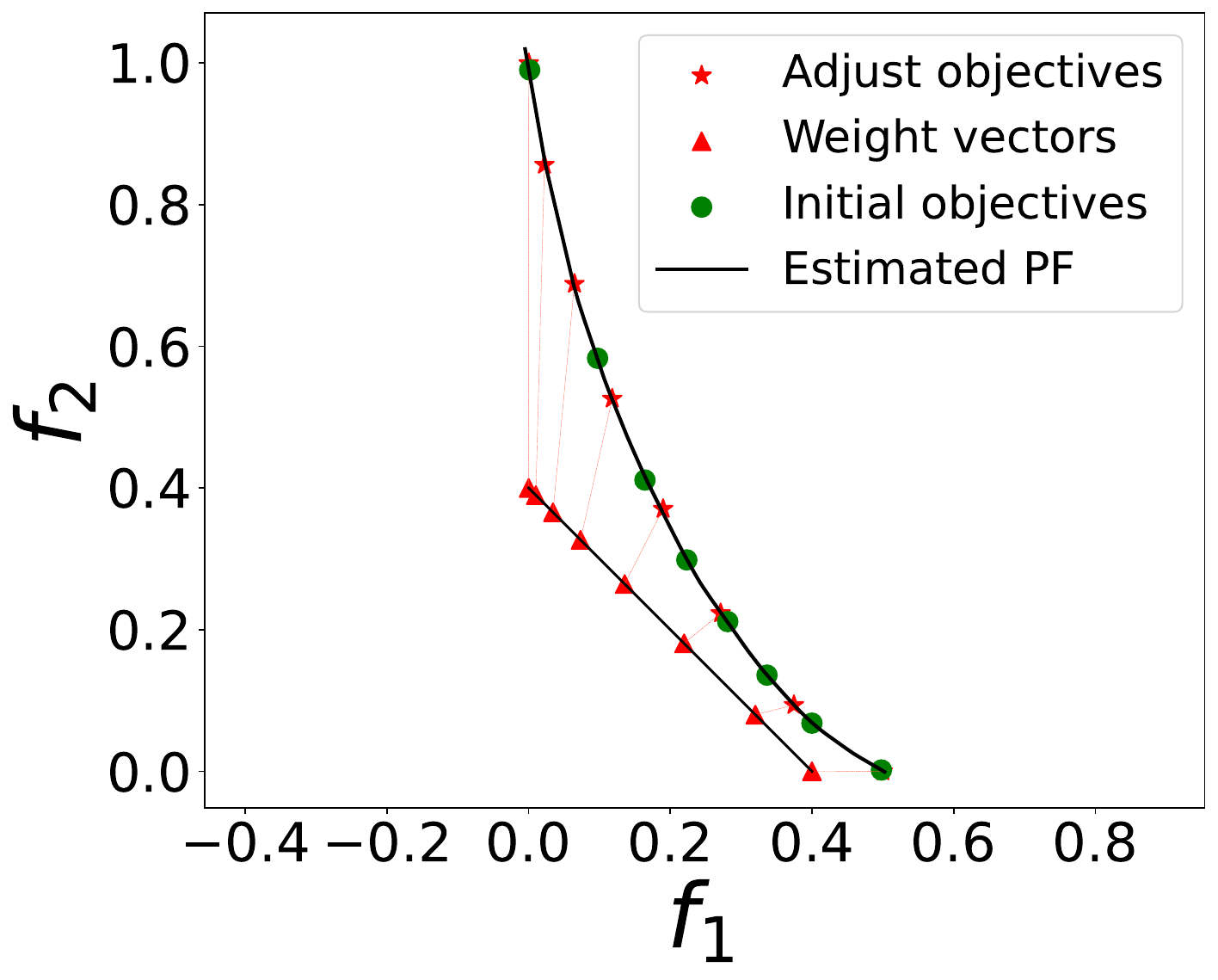}
  \caption{Weight adjustment.}
  \vspace{-10pt}
  \label{fig:pref_adjust}
\end{wrapfigure}

\subsection{Weight adjustment with a PFL model}
Solving the optimal configuration (\Cref{eqn:pack_pf}) on the PF with an arbitrary set size $N$, is generally a NP-hard problem \cite{borodachov2019discrete}. Therefore, we propose a parameterized problem to approximate problem (\Cref{eqn:pack_pf}). Notice that, when the PFL model $\vh_\vphi$ is trained, $\hat{\pf} = \{ \vh_\vphi(\vtheta) | \vtheta \in {[0, \frac{\pi}{2}]}^{m-1} \}$ is a good approximator of the true PF. We further show the average error of $\hat{\pf}$ and $\pf$ is bounded in the empirical risk and maximal pairwise distances in \Cref{rmk_estimate}. Solving the optimal weight factors which lead to uniform distribution on the estimated PF can be formulated as the following problem, 
\begin{equation} \label{eqn:pack_pf_hat}
    \max_{\mTheta_N} \hat{\delta} = \max_{\mTheta_N} \left( \min_{1 \leq i < j \leq N} \rho(\vh_\vphi(\vtheta^{(i)}), \vh_\vphi(\vtheta^{(j)})) \right)
\end{equation}
After this replacement, the parameterized optimization problem can be solved efficiently by projected gradient ascent method
$$\vtheta^{(i)} \leftarrow \mathrm{Proj}(\vtheta^{(i)} + \eta \nabla_{ \vtheta^{(i)} } \hat{\delta}), \quad i \in [N],$$
where $\eta$ is a learning rate and the $\mathrm{Proj}$ operator projects a weight angle back to its domain $[0, \frac{\pi}{2}]^{m-1}$. 

\Cref{fig:pref_adjust} demonstrates that a PFL model can effectively estimate the whole PF well with only a few numbers of Pareto objectives. The blue dots represent the original Pareto objectives optimized by MOEA/D, while the red stars are Pareto objectives after weight adjustment.

\subsection{PFL generalization bound} \label{sec:pfl:theory}
As discussed in \Cref{app:risk}, the population can be controlled by bounding this generalization error. 
We discuss the PFL generalization error, namely $\tilde{\epsilon} = |R(\tilde{\vh}) - \hat{R}(\tilde{\vh})|$ for an arbitrary diffeomorphic $\tilde{\vh}(\cdot)$, where $R(\cdot)/\hat{R}(\cdot)$ denote the population and empirical risks respectively. 

\begin{theorem}[PFL Generalization Bound]
\label{thm:generalization} We first make some regularity assumptions:
\begin{enumerate}[itemsep=-0.2em, topsep=0.0em, leftmargin=1.0em]
    \item(Function smoothness). Both $(\tilde{\vh} - \vh_*)(\cdot)$ and $\vh_*^{-1}(\cdot)$ are $L$ - and $L'$ -Lipschitz, respectively, ie $\forall \vtheta^{(1)}, \vtheta^{(2)} \in {[0,\frac{\pi}{2}]}^{m-1}$ and $\forall \vy^{(1)}, \vy^{(2)} \in \R^m$, 
    \bee
    \|(\tilde{\vh} - \vh_*)(\vtheta^{(1)}) - (\tilde{\vh}-\vh_*)(\vtheta^{(2)})\|&\leq L\|\vtheta^{(1)} - \vtheta^{(2)}\|,
    \\
    \|\vh^{-1}_*(\vy^{(1)}) - \vh^{-1}_*(\vy^{(2)})\| &\leq L'\| \vy^{(1)} - \vy^{(2)} \|, 
    \ee
    where $\vh_*$ denotes the true mapping function from weights to objectives.  
    \item(Function bound). $\| \tilde{\vh} - \vh_* \|_\infty \leq A$, $\norm{\vh_*^{-1}}_\infty \leq A'$. The function $\vh^{-1}$ exists according to \Cref{lem:diff}. 
    \item(Manifold property). We assume $\mathcal{T}$ is a differentiable, compact ($m$-1)-D manifold, a common assumption \cite{hillermeier2001generalized, roy2023optimization}. We also assume $\pf$ is connected which is common.   
\end{enumerate}
Under the above assumptions, for the risk $\tilde{\epsilon}= |R(\tilde{\vh}) - \hat{R}(\tilde{\vh})|$, we have the following bound,
\begin{equation}
    \tilde{\epsilon} \leq 2 \mathcal{H}_{m-1}(\mathcal{T}) A A' L L' \delta_v + 2 CA^2\sqrt{\mathcal{W}_1 (\mathcal{U}, \widetilde{\mY}_N ) + \delta_v},
\end{equation}
where $\mathcal{U}$ is the uniform distribution over $\Tau$, $\widetilde{\mY}_N$ is the empirical distribution of $\mY$, $\mathcal{W}_1(\cdot, \cdot)$ is the Wasserstein distance with the $\ell_1$ norm, $\delta_v$ represents the maximal diameter of Voronoi cells \footnote{For the formal definition of Voronoi cells and diameter of a set, please refer to Definitions \ref{app:def:diam} and \ref{app:def:vor}}, and $C$ is some universal constant representing the smoothness of $\Tau$ \cite{chae2020wasserstein}. We left the proof in \Cref{app:sec:proof_thm2}.
\end{theorem}
\begin{remark}
    In \Cref{thm:generalization}, the error bound for $\tilde{\epsilon}$ involves two quantities, the diameter of the Voronoi cell $\delta_v$ and Wasserstein distance $\mathcal{W}_1 (\mathcal{U}, \widetilde{\mY}_N)$. 
    The margin $\delta_v$ is controlled by maximizing the minimal separation distance.
    The decaying rate of $\mathcal{W}_1 (\mathcal{U}, \widetilde{\mY}_N)$ is impacted by not only the margin $\delta_v$, but also the manifold properties of the PF.
    However, by \Cref{prop:uniform}-\circled{2}, we still have $\mathcal{W}_1 (\mathcal{U}, \widetilde{\mY}_N) \to 0$ since $\widetilde{\mY}_N$ weakly converges to $\mathcal{U}$, and minimizing the margin $\delta_v$ is thus critical to the control of the generalization error $\tilde{\epsilon}$. 
\end{remark}
\begin{remark}[Average PF estimation error] \label{rmk_estimate}
     $R(\tilde{\vh})$ equals to $\int_\DomTheta \| \tilde{\vh}(\vtheta) - h_*(\vtheta) \| d\vtheta$, the average error of $\hat{\pf}$ and $\pf$. Hence, this average error can accordingly be upper bounded by $\tilde{\epsilon} + \hat{R}(\tilde{h})$.  
\end{remark}

\section{Experiments}
\subsection{Experiment settings}
We validate the effectiveness of the proposed method on various MOEA problems, including ZDT1,2,4,6 \cite{deb2006multi} (PFs of ZDT3,5 are disconnected thus not included), DTLZ 1-4 \cite{deb2002scalable}, and real-world problems, namely, four-bar truss design (RE21), reinforced concrete beam design (RE22), rocket injector design (RE37), car side impact design (RE41), and conceptual marine design (RE42) \cite{tanabe2020easy}. RE41 and RE42 are four-objective problems owning a very large objective space, and the previous method is very difficult to cover the PF efficiently. 

For presentation, we normalize the PF of RE37/RE41/RE42 to ${[0,1]}^3$. To test the ability to deal with objectives of different scales, the PFs of RE21 and RE22 are normalized to $[0, 0.5] \times [0,1]$. ZDTs and DTLZs possess numerous local optima that cannot be identified by gradient-based MOO methods (see \Cref{sec:gradmoo}). The RE problems are real-world problems with unknown PFs.

\begin{figure*}
    \centering
    
    \subfloat[\scriptsize AWA-1 RE41]{\includegraphics[width = \hwidth \textwidth]{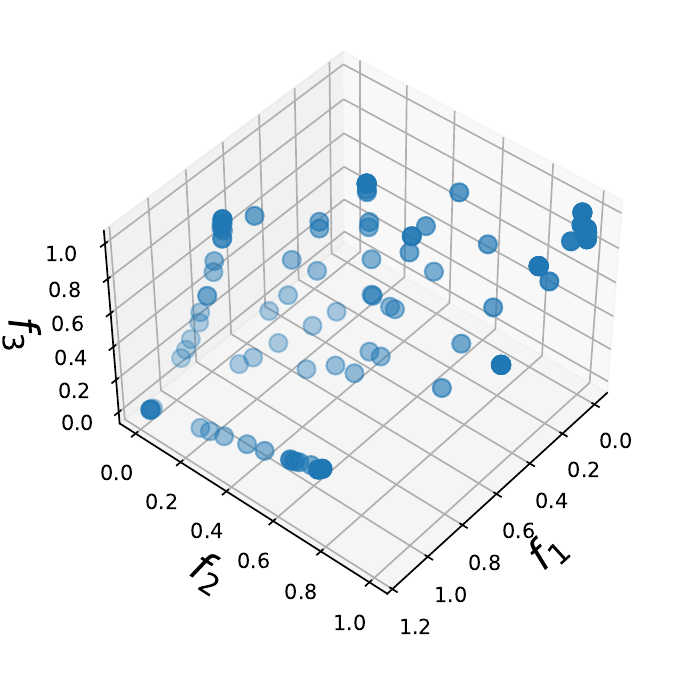}}
    \hfill
    \subfloat[\scriptsize AWA-2 RE41]{\includegraphics[width = \hwidth \textwidth]{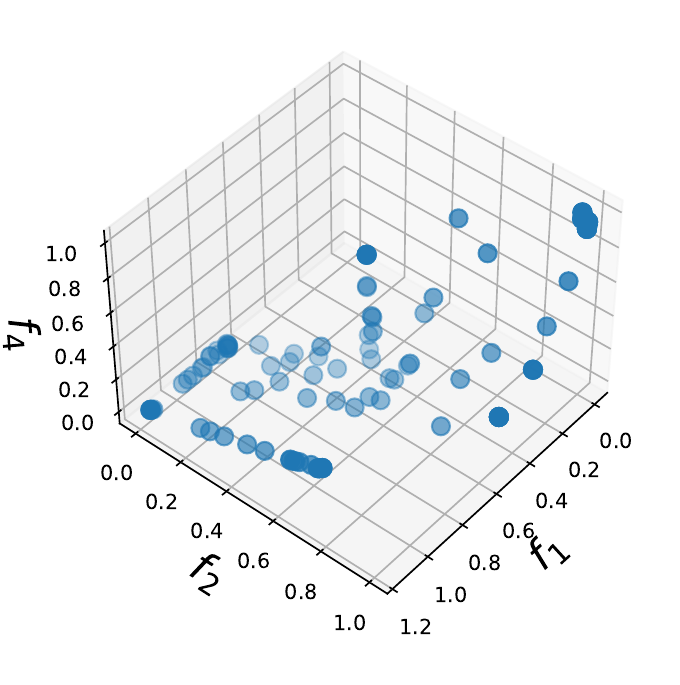}}
    \hfill
    \subfloat[\scriptsize AWA-3 RE41 ]{\includegraphics[width = \hwidth \textwidth]{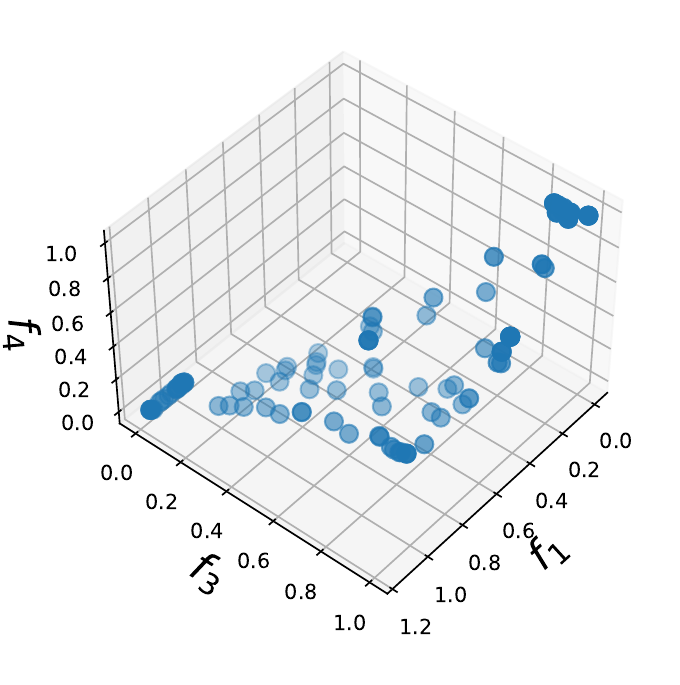}}
    \hfill
    \subfloat[\scriptsize AWA-4 RE41 ]{\includegraphics[width = \hwidth \textwidth]{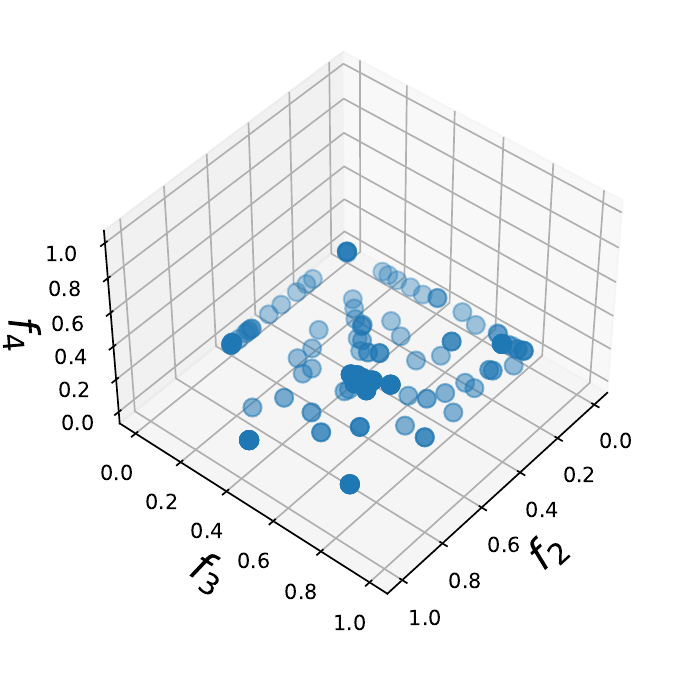}}
    \hfill
    \subfloat[\scriptsize AWA RE37]{\includegraphics[width = \hwidth \textwidth]{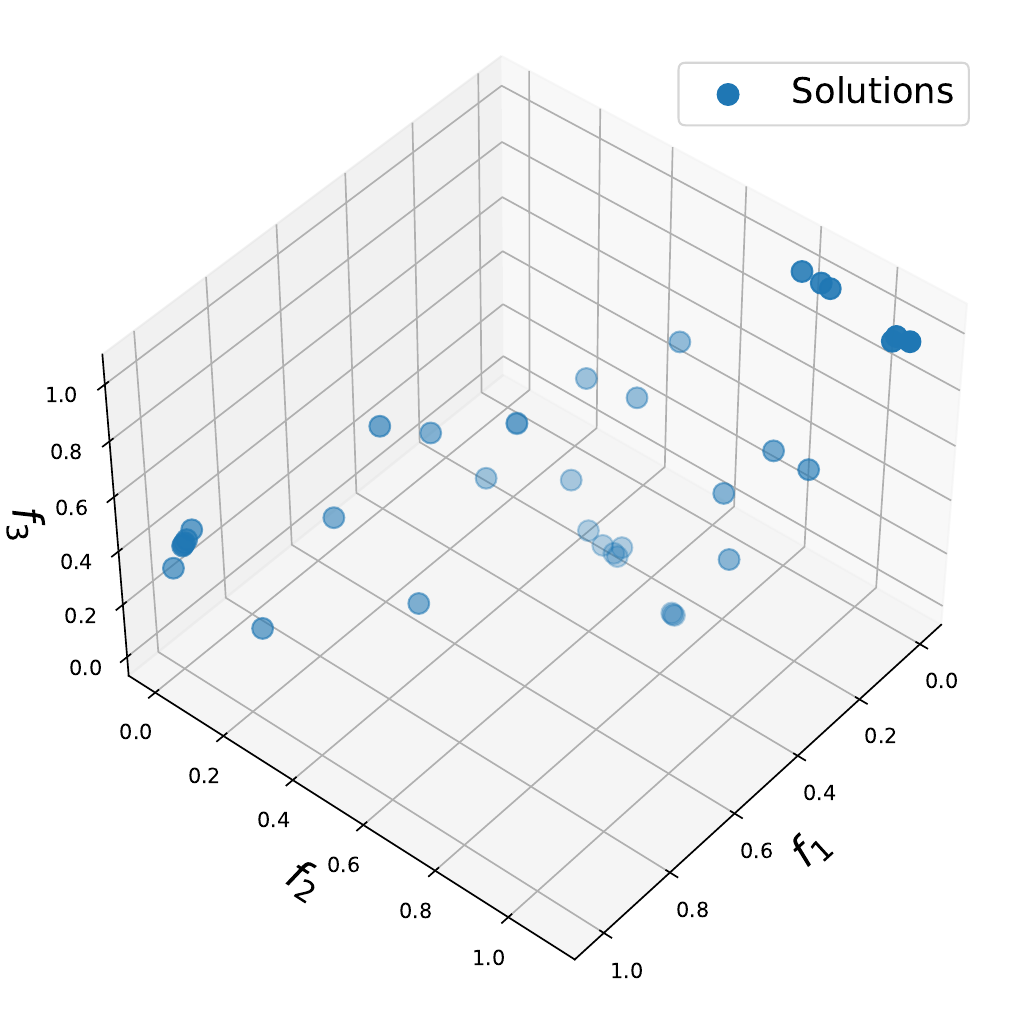}}
    \hfill
    \subfloat[\scriptsize AWA RE42-4]{\includegraphics[width = \hwidth \textwidth]{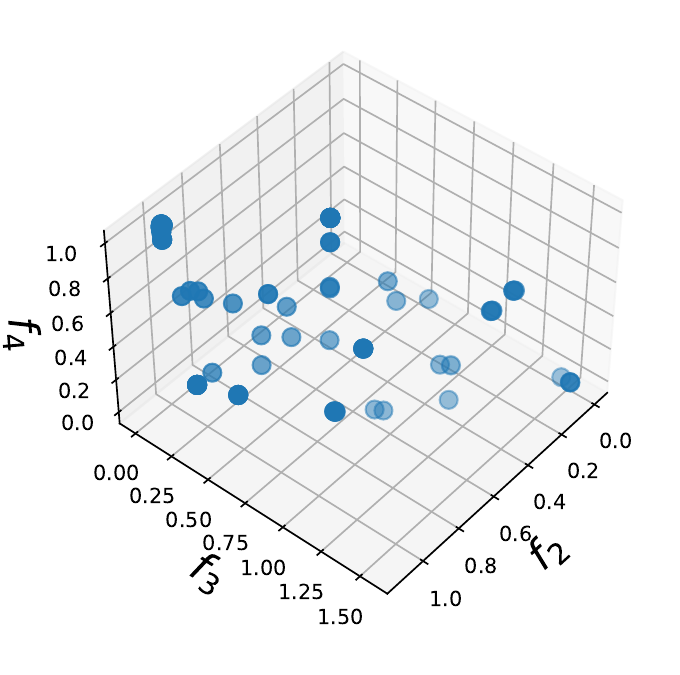}}
    \vspace{-10pt}
    \\
    \subfloat[\scriptsize SMS-1 RE41]{\includegraphics[width = \hwidth \textwidth]{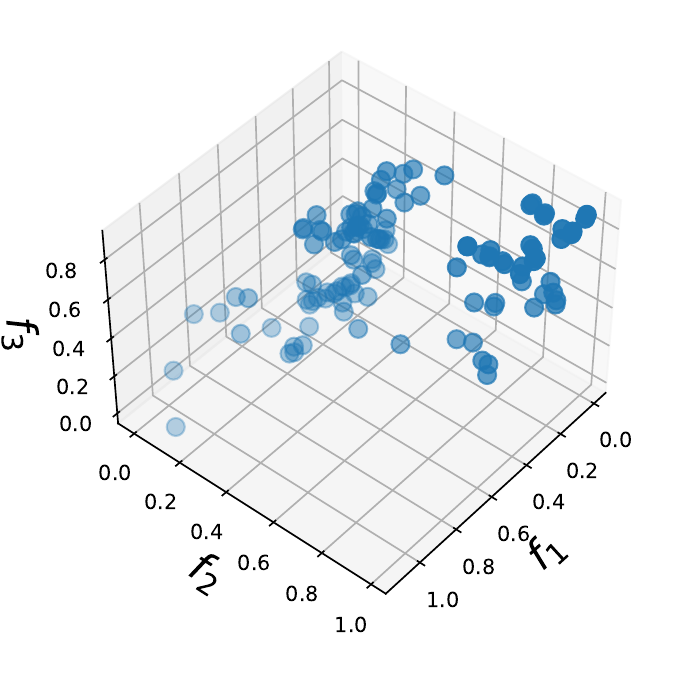}}
    \hfill
    \subfloat[\scriptsize SMS-2 RE41]{\includegraphics[width = \hwidth \textwidth]{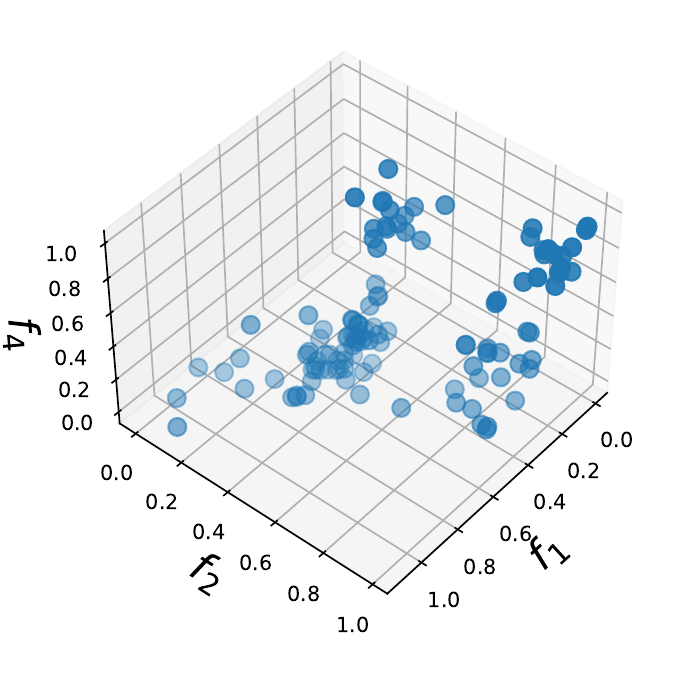}}
    \hfill
    \subfloat[\scriptsize SMS-3 RE41]{\includegraphics[width = \hwidth \textwidth]{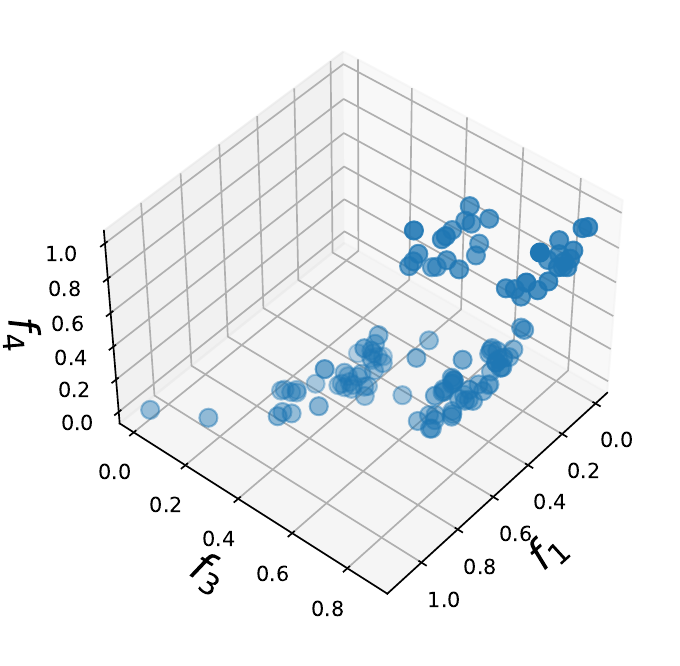}}
    \hfill
    \subfloat[\scriptsize SMS-4 RE41]{\includegraphics[width = \hwidth \textwidth]{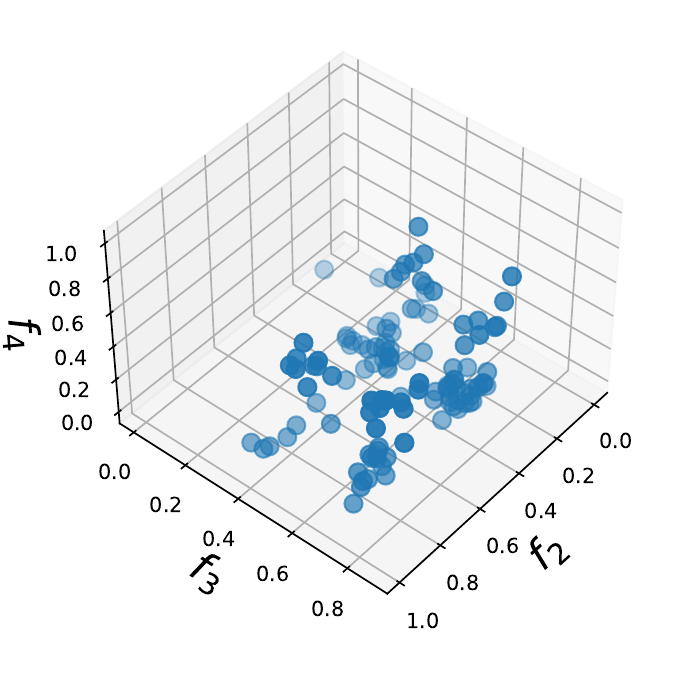}}
    \hfill
    \subfloat[\scriptsize SMS RE37 ]{\includegraphics[width = \hwidth \textwidth]{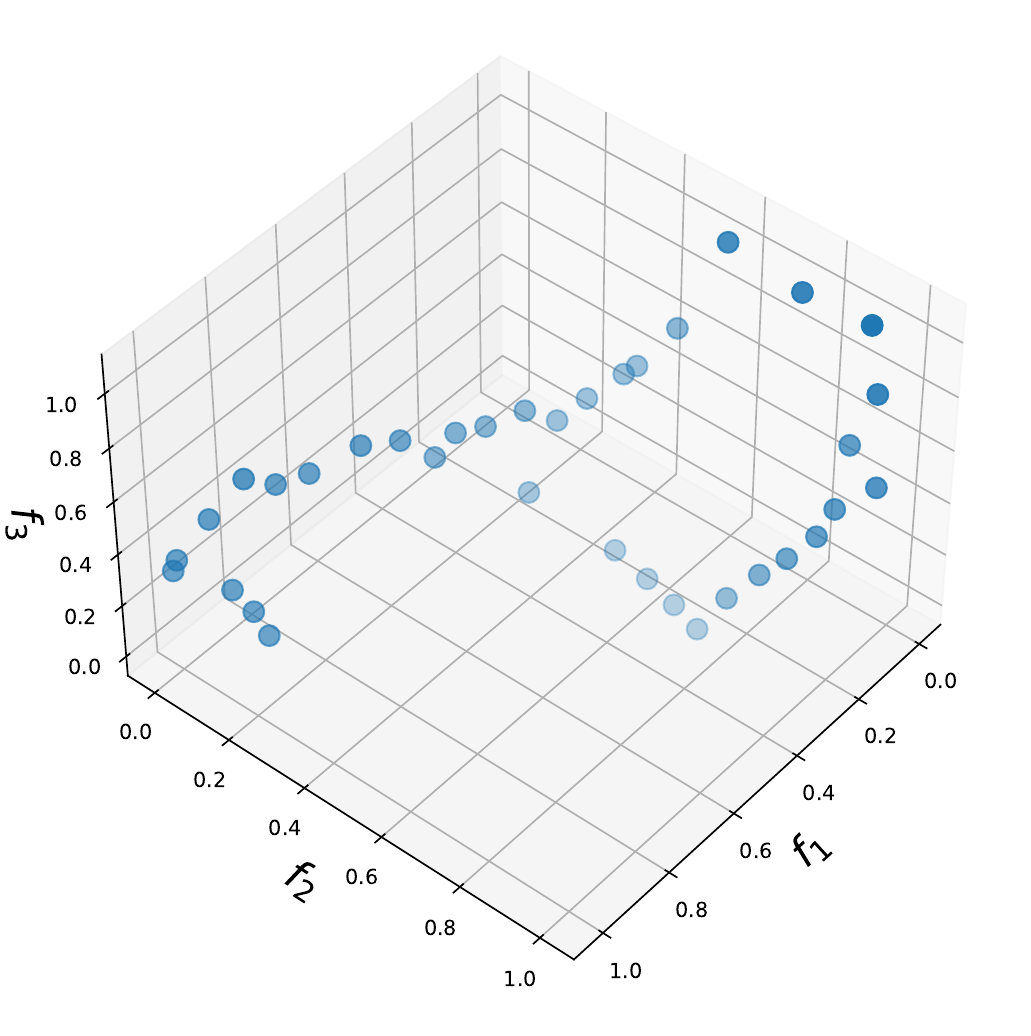}}
    \hfill
    \subfloat[\scriptsize SMS-4 RE42]{\includegraphics[width = \hwidth \textwidth]{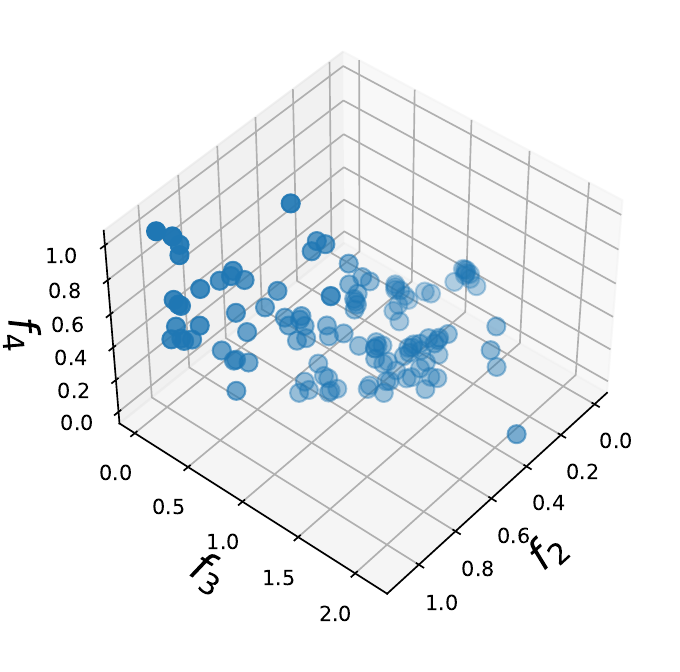}}
    \vspace{-10pt}
    \\
    \subfloat[\scriptsize UMOEA/D-1 RE41]{\includegraphics[width = \hwidth \textwidth]{Figure/main/RE41/adjust/proj_1.pdf}}
    \hfill
    \subfloat[\scriptsize UMOEA/D-2 RE41]{\includegraphics[width = \hwidth \textwidth]{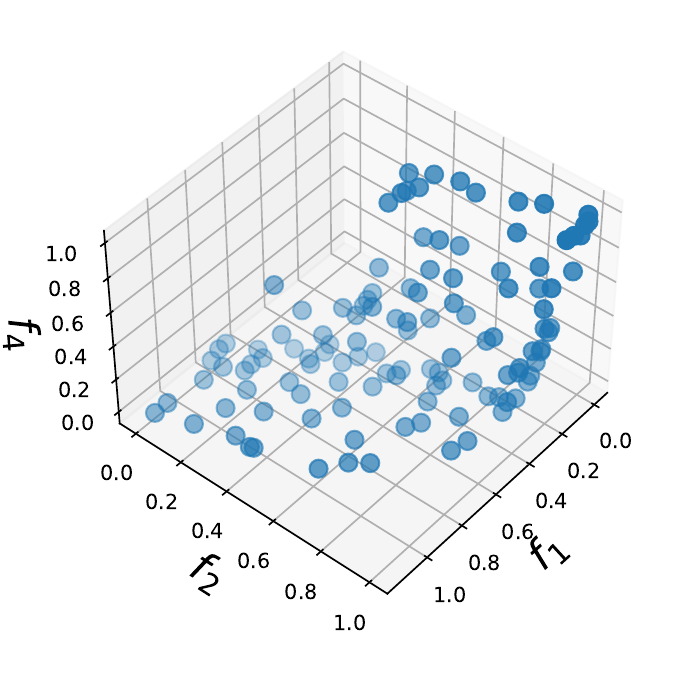}}
    \hfill
    \subfloat[\scriptsize UMOEA/D-3 RE41]{\includegraphics[width = \hwidth \textwidth]{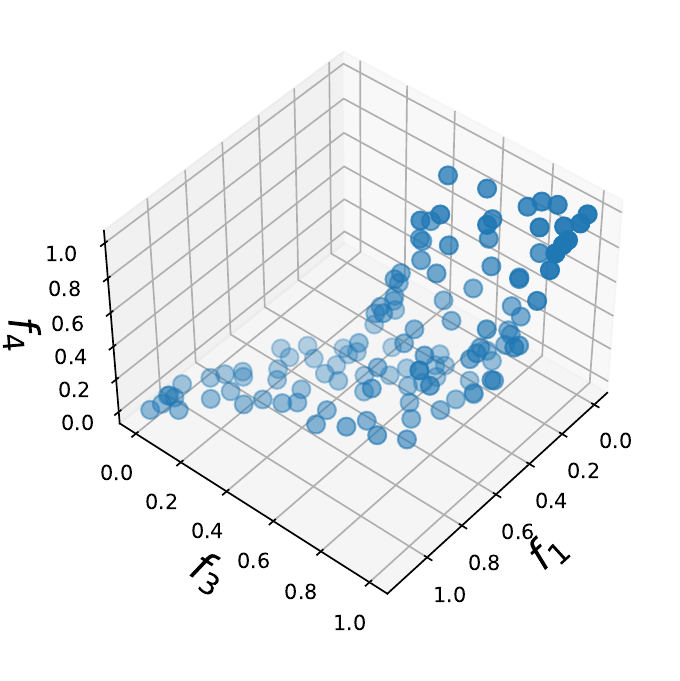}}
    \hfill
    \subfloat[\scriptsize UMOEA/D-4 RE41]{\includegraphics[width = \hwidth \textwidth]{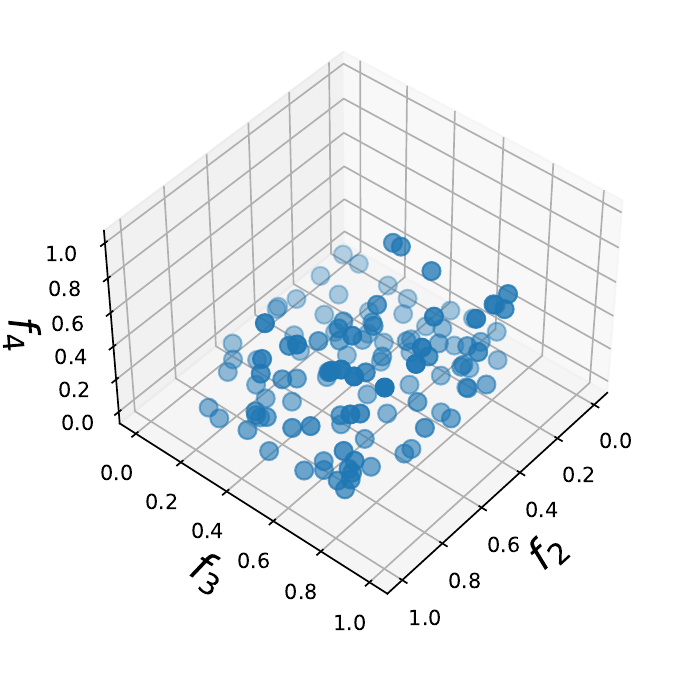}}
    \hfill
    \subfloat[\scriptsize UMOEA/D RE37 ]{\includegraphics[width = \hwidth \textwidth]{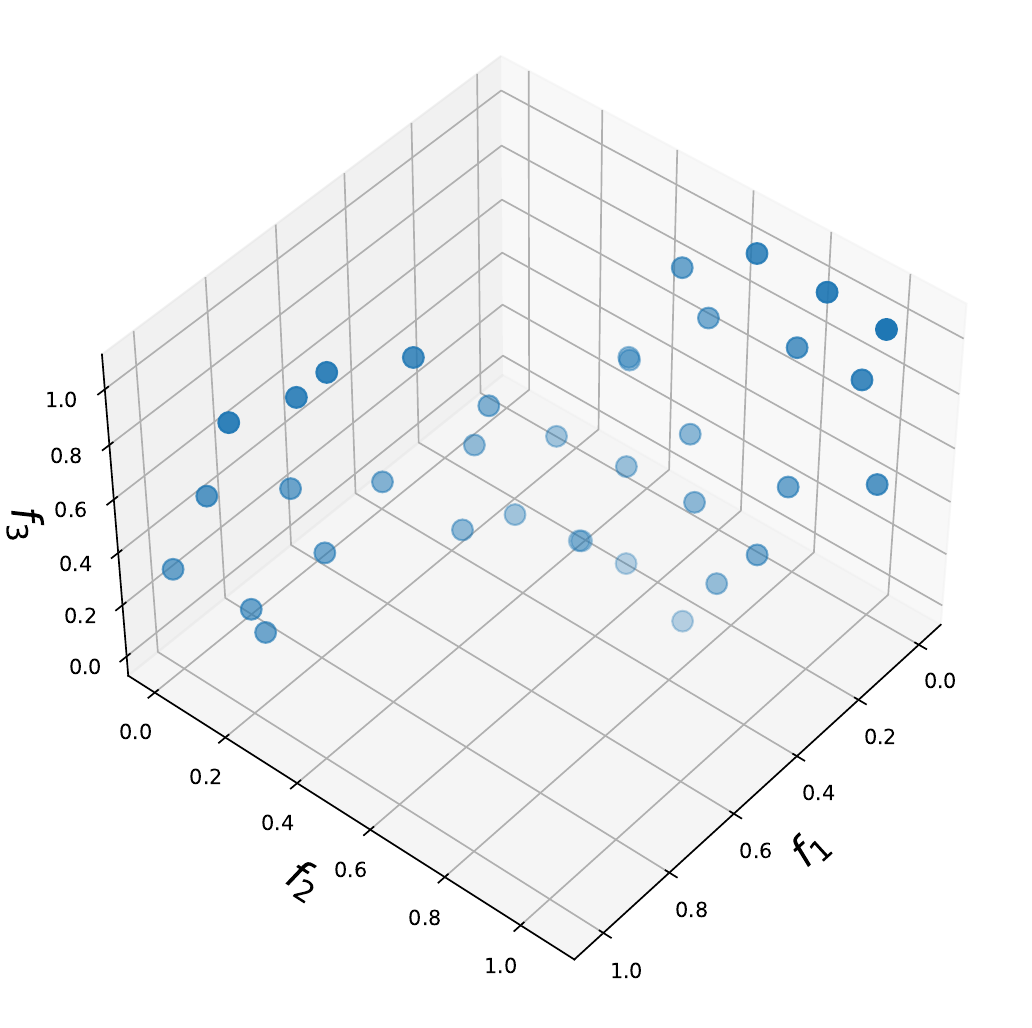}}
    \hfill
    \subfloat[\scriptsize UMOEA/D-4 RE42]{\includegraphics[width = \hwidth \textwidth]{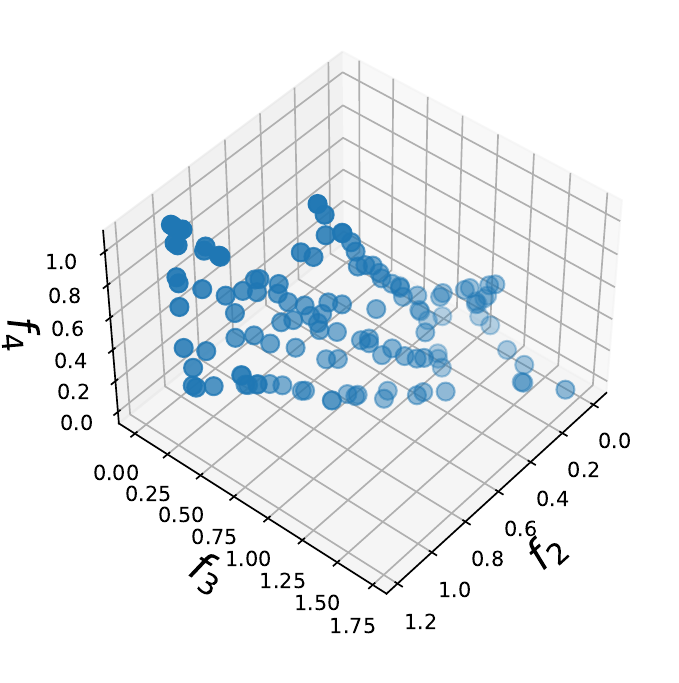}}
    \\
    \vspace{-5pt}
    \caption{Experimental results on the three-objective RE37 and four-objective RE41/RE42 problems. Here, the ```Method-$i, i=1,2,3,4$'' notation refers to the projection of the objectives on the objective space $(f_1, f_2, f_3)$, $(f_1, f_2, f_4)$, $(f_1, f_3, f_4)$, and $(f_2, f_3, f_4)$, respectively. }
    \label{fig:main}
\end{figure*}

\begin{table}
    \centering 
    \footnotesize
    \caption{ The running time (every 1k generations) of  SMS-EMOA, PaLam, and UMOEA/D. }
    \begin{tabular}{lrrrr}
    \toprule
    Running Time (minutes) & DTLZ2 & RE37 & RE41 & RE42 \\
    \midrule
    SMS-EMOA &  1.21     &  3.23 & 24.12 & 18.76 \\
    PaLam  &  0.72     & 1.59 & 7.69 & 6.05 \\
    UMOEA/D &  \tb{0.56}     & \tb{0.71} & \tb{2.48} & \tb{2.55}\\
    \bottomrule
    \end{tabular}%
    \label{tab:running_time}%
\end{table}%

The implementation in this study relies on the pymoo \cite{pymoo} and PyTorch \cite{paszke2019pytorch} libraries. We use the simulated binary crossover operator (SBX) and the polynomial mutation technique \cite{deb2001self} for MOEA/D-based methods. Following the pymoo setting, we do not maintain an external population (EP), as it can be computationally and storage intensive \cite{li2011adaptive}.

We compare our method with \circled{1} vanilla \textbf{ MOEA / D} \cite{zhang2007moea}, \circled{2} MOEA/D with adaptive weight adjustment ( \textbf{MOEA/D-AWA} ) \cite{qi2014moea,de2018moea}, \circled{3} \textbf{PaLam} \cite{siwei2011multiobjective}, \circled{4} \textbf{SMS-EMOA} \cite{beume2007sms}, \circled{5} MOEA/Ds adjustment with linear-segment \cite{dong2020moea} (\textbf{MOEA/D-L}), and \circled{6} \tb{MOEA/D-GP} \cite{wu2017adaptive}. SMS-EMOA and PaLam are hypervolume-based method. Though there are many fast hypervolume approximation algorithms, calculate the hypervolume or its gradient is computation inefficient \cite{guerreiro2020hypervolume}.

To assess performances, we utilize \circled{1} the \texttt{hypervolume} (\texttt{HV}) ($\uparrow$) \cite{guerreiro2020hypervolume}, \circled{2} the \texttt{sparsity} ($\downarrow$) \cite{xu2020prediction}, \circled{3} the \texttt{spacing} ($\downarrow$) \cite{schott1995fault}, \circled{5} the minimal distance on the PF ($\delta$) ($\uparrow$) and its soft version \circled{5} ($\tilde{\delta}$) ($\uparrow$) indicators. Please refer to \Cref{app:sec:comp_mtds} and \ref{app:sec:metrics} for detailed descriptions related methods and indicator details. 

\begin{table}[htbp]
  \centering
  \tiny
  \caption{Results on all problems averaged on five random seeds.} \label{tab:all_main}%
    \begin{tabular}{lrrrrr}
    \toprule
     Method & \multicolumn{1}{r}{\texttt{HV}} & \multicolumn{1}{r}{\texttt{Spacing}} & \multicolumn{1}{r}{\texttt{Sparsity}} & \multicolumn{1}{r}{$\delta$} & \multicolumn{1}{r}{$\tilde{\delta}$} \\
    \midrule
    \multicolumn{6}{c}{RE37} \\
    \midrule
    SMS-EMOA & 1.1143 & 0.0423 & 0.0052 & 0.0294 & -0.1270 \\
    MOEA/D-AWA & 1.0768 & 0.0797 & 0.0101 & 0.0012 & -0.2055 \\
    MOEA/D-GP & 1.0737 & 0.0715 & 0.0083 & 0.0023 & -0.1723 \\
    MOEA/D & 1.0519 & 0.0758 & 0.0122 & 0.0000 & -0.2055 \\
    PaLam & \tb{1.1150} & 0.0712 & 0.0050 & 0.0023 & -0.1473 \\
    UMOEA/D & 1.1114 & \tb{0.0416} & \tb{0.0047} & \tb{0.0483} & \tb{-0.0805} \\
    \midrule
    \multicolumn{6}{c}{RE41} \\
    \midrule
    SMS-EMOA & 1.0976 & 0.0516 & 0.0014 & 0.0037 & -0.2458 \\
    MOEA/D-AWA & 1.1432 & 0.0676 & 0.0026 & 0.0000 & -0.3001 \\
    MOEA/D-GP & 1.1694 & 0.0687 & 0.0013 & 0.0001 & -0.2798 \\
    MOEA/D & 1.1328 & 0.0578 & 0.0034 & 0.0000 & -0.2998 \\
    PaLam & \tb{1.2075} & 0.0551 & 0.0009 & 0.0019 & -0.2352 \\
    UMOEA/D & 1.2072 & \tb{0.0436} & \tb{0.0008} & \tb{0.0060} & \tb{-0.1876} \\
    \bottomrule
    \end{tabular}%
\end{table} %

\subsection{Results Analysis}
The average results on five random seeds for \emph{all} problems are displayed in Table \ref{tab:obj2} and \ref{tab:obj3} in \Cref{app:sec:res_all}, along with the visualization results in \Cref{fig:2obj_zdt1}-\ref{fig:4obj_re42}. Due to page limit, we report on three representative results on RE37, RE41, RE42 in the main paper. We ablate in \Cref{sec:gradmoo} that only using gradient information easily lead to a local optimal solution. 

The proposed method achieves the optimal \texttt{spacing} indicator. For two-objective problems, the \texttt{spacing} indicator is very close to zero, indicating that the distances between adjacent solutions are nearly equal. We observed that HV-based methods such as SMS-EMOA and PaLam generated more diverse solutions compared to the vanilla MOEA/D. The main reason for this is that, in many real-world problems, numerous weight vectors do not intersect with the PF, resulting in duplicate Pareto solutions. This duplication wastes computation resources. UMOEA/D addresses the issue of duplicate Pareto objectives through adaptive weight adjustment. By aiming to maximize the minimal separation, UMOEA/D avoids generating duplicate Pareto solutions, as duplicate solutions would automatically have a minimal distance of zero.

Contrary to the commonly held belief that hypervolume-based methods yield diverse solutions \cite{auger2012hypervolume, guerreiro2020hypervolume}, our findings reveal that hypervolume indicators can be very similiar, while the distribution of solutions can vary significantly. This means that adopting a novel indicator in MOO to accurately measure the diversity level of solutions, as proposed, is necessary. Figure \ref{fig:main}(k) shows that hypervolume-based methods tend to generate solutions on the boundary of the PF and lack guarantees of $\delta$-dominance. In contrast, UMOEA/D generates uniformly distributed Pareto objectives more effectively than hypervolume-based methods across most problems. Another notable advantage of UMOEA/D over hypervolume-based methods is the computational efficiency it offers. Computation of the hypervolume for a 3/4 objective problem can be expensive, while solving MOEA/D subproblems can be parallelized, leading to highly efficient computations. This advantage is supported by the results presented in Table \ref{tab:running_time}, demonstrating that UMOEA/D successfully achieves approximately a great 10$\times$ speed improvement compared to SMS-EMOA and a 3$\times$ speed improvement compared to PaLam.

UMOEA/D offers a notable advantage over other weight adjustment methods, such as MOEA/D-AWA \cite{qi2014moea,de2018moea}, MOEA/D-GP \cite{wu2017adaptive}, and MOEA/D-L \cite{dong2020moea}. It excels in accurately modeling the Pareto Front (PF) while providing theoretical guarantees for the level of uniformity. The figure clearly demonstrates this advantage, where UMOEA/D generates significantly more uniformly distributed Pareto objectives for challenging 3/4 objective problems. In comparison, MOEA/D-AWA produces Pareto objectives that are often closely clustered together, resulting in limited coverage across a broad region. Conversely, UMOEA/D consistently generates solutions that span the entire Pareto front. 

\section{Conclusions and future directions}

This paper attempts to address a long-standing problem in MOEAs: the generation of uniformly distributed Pareto objectives to well represent the characteristics of the whole PF. To the best of our knowledge, this is the first paper that rigorously analyzes the distribution of Pareto objectives, and thereby sheds light on the understanding of solution generation in MOEAs. Building upon the theoretical findings, we introduce a novel algorithm that constructs a uniformly distributed Pareto objective set through NN-assisted weight adjustment. We will extend UMOEA/D for large applications such as drug design in our further work.

\clearpage

\section*{Broader Impact}
As a multiobjective method, our work poses low social risk as its results depend on downstream applications. One use case is building trustworthy AI systems to understand tradeoffs between conflicting objectives. Hence, the negative societal impact of our work is unlikely.

\bibliography{ref}
\bibliographystyle{icml2024}

\newpage
\appendix
\onecolumn
\section{Methodology}

\subsection{MOO theories} \label{app:sec:moo}
As a preliminary, we introduce some basic MOO properties which used in the main paper. The definition of an aggregation function is adopted and modified from \citep[Chapter 2.6]{miettinen1999nonlinear}, where it was originally under the name of value functions.
\begin{definition}[Aggregation function] \label{def:agg}
    An aggregation function $g(\cdot): \mathbb{R}^m \mapsto \mathbb{R}$. $g(\cdot)$ is a decreasing function, i.e., $g(x) < g(x')$, if $x_i < x'_i, \; \forall i \in [m]$.  
\end{definition}
One example of aggregation function is the modified Tchebycheff function (\Cref{eqn:g-mtche}) studied in the main paper. We have the following Lemmas (adopted and modified from \citep[Thereom 2.6.2]{miettinen1999nonlinear}) for an aggregation function,
\begin{lemma}
    \label{lemma:agg1}
    Let $\vy^*$ be one of the optimal solution of $g(\cdot, \vlam)$, then $\vy^*$ is (weakly) Pareto optimal. 
\end{lemma}
\begin{lemma}
    \label{lemma:agg2}
    Let $\vy^*$ be the only optimal solution of $g(\cdot, \vlam)$, then $\vy^*$ is Pareto optimal. 
\end{lemma}
\begin{definition}[Weakly Pareto Optimal Solution]
    $\vx^{(a)}$ is a weakly Pareto solution if there is no other solution $\vx^{(b)} \in \mathscr{X}$ such that $\vf(\vx^{(b)}) \prec \vf(\vx^{(a)})$, where $\vf(\vx^{(b)}) \prec \vf(\vx^{(a)})$ means $f_i(\vx^{(b)}) < f_i(\vx^{(a)})$ for all $i \in [m]$. 
\end{definition}
\begin{corollary} \label{cor:unique}
    According to \Cref{lemma:agg1} and \ref{lemma:agg2}, for a given $\vlam$, if there does not exist weakly Pareto solutions, the optimal solution of $g(\cdot, \vlam)$ is unique and is a Pareto solution. 
\end{corollary}

\section{Missing proofs} \label{sec:missing}
\begin{proof}
    The proof has two steps. The fist step is show under \Cref{ass:full}, $\vh$ is injective. We leave it as \Cref{lma:injective}. 
    \begin{lemma}[Injection] \label{lma:injective}
        We assert that \( \vh(\vlam) = \vy = k \cdot \vlam + \vz \) for each \( \vlam \), where \( k \cdot \vlam + \vz \) represents a Pareto optimal solution. This means that for a given \( \vlam \), the corresponding \( \vy \) is unique. Suppose there exists another \( \vy' = k' \cdot \vlam' + \vz \) such that \( \vh(\vlam) = \vy' \). If \( \vlam' \) differs from \( \vlam \), then by the definition of Pareto optimality, for at least one index \( j \), we have \( \frac{\vy'_j - \vz_j}{\lambda_j} > \frac{\vy_j - \vz_j}{\lambda_j} \). Thus, \( \vy' \) cannot be an optimal solution if \( \vlam' \neq \vlam \), ensuring the injectivity of function $\vh(\vlam)$.        
    \end{lemma}
    With the previous lemma, 
    Consider the function $\rho(\vx^*) = \frac{ f_i(\vx^*) - z_i}{\lambda_i}, \forall i \in [m]$, where $\vx^*$ denotes the optimal solution of the modified Tchebycheff function, defined as $\vx^* = \argmin_{x \in \mathscr{X}} \max_{i \in [m]} \lbr{\frac{f_i(\vx) - z_i}{\lambda_i}} $. Given the strict convexity of the functions $f_i$, the expression $\max_{i \in [m]} \frac{f_i(\vx) - z_i}{\lambda_i}$ is also strictly convex, ensuring the uniqueness of the optimal solution $\vx^*$. The value of $\vlam$ can be derived from the equation $\vlam = \frac{ \vf(\vx^*) - \vz}{\rho(\vx^*)}$. Under the condition of Lemma \ref{lma:injective}, $\rho(\vx^*)$ has an analytical expression, and as a result $\vlam(\vx^*)$ can be expressed as:
    \begin{equation} \label{eqn:lam}
        \vlam(\vx^*) = \frac{ \vf(\vx^*) - \vz}{\sum_{i \in [m]} (f_i(\vx^*)-z_i)} 
    \end{equation}
    This equation implies that $\vlam$ is differentiable from $\vx^*$. Therefore, the inverse function of Equation \ref{eqn:lam} exists. Since $\vlam(\vx^*)$, by the inverse function theorem, $\vx^*(\vlam)$ is differentiable, making the composed function $\vh(\vlam) = \vf \circ \vx^*(\vlam)$ differentiable. The application of Lemma \ref{lma:injective} and uniqueness condition confirms that $\vh(\vlam)$ is a bijection. Combined with its differentiability, this confirms that $\vh(\vlam)$ is indeed a diffeomorphism \footnote{In a recent study by \citet{roy2023optimization}, it was shown that for a smooth linear scalarization function, the Pareto objectives are diffeomorphic to the weight simplex. Our result is a stronger result, as we give the same results for a non-smooth modified Tchebycheff function}.
\end{proof}

\subsection{Set Geometry} \label{app:sec:set}
In this section, we provide several useful definitions for set geometry. These definitions main serve for two purpose: (1) to give the rigid definition of $\delta$-dominance and (2) to prove the generalization bound of a PFL model. 

In Definitions \ref{def:pac} and \ref{def:cov}, we introduce the $\delta$-packing and $\delta$-covering numbers of a set $\mY$, which are the maximal number of $\delta/2$-balls and minimal number of $\delta$-balls one needs, to pack in and cover $\mY$, respectively. They measure the metric entropies of $\mY$; See e.g., \citet{vershynin2018high}. 

Definition \ref{app:def:diam} defines the diameter of $\mY$, which  measures the supremum of the distances between pairs of points in $\mY$. Definition \ref{app:def:vor} gives a partition of $\mY$, given some point set $\mathcal{S}$ in $\mY$. Briefly, any point in $\mY$ will be partitioned to the Voronoi cell that has the minimal distance to it. Intuitively, if points in $\mathcal{S}$ are approximately evenly distributed, then they will induce  Voronoi cells that have similar volumes to each other.

Finally, Hausdorff measure in Definition \ref{app:def:hausdorff} introduces an adaptive and accurate way to measure the volumes of different sets in a multidimensional Euclidean space. For example, a curve  has a trivial zero Borel measure in $\R^2$, yet the $1$-dimensional Hausdorff  measure is non-trivial and could measure its length. Likewise, for a spherical shell in $\R^3$ which has zero Borel measure, the corresponding $2$-dimensional Hausdorff measures its surface area.

\begin{definition}[$\delta$-packing, $\delta$-packing number]\label{def:pac}
    A $\delta$-packing of a set ${\mY}$ is a collection of element $\{\vy^{(1)}, \ldots, \vy^{(N)}\}$ such that, $\rho(\vy^{(j)}, \vy^{(k)})>\delta$ for all $j \neq k$. The $\delta$-packing number $N_{\text{pack}}(\delta, {\mY})$ is the largest cardinality among all  $\delta$-packing. Here $\delta$ is called the packing distance. 
\end{definition}
\begin{definition}[$\delta$-covering, $\delta$-covering number]\label{def:cov}
    A $\delta$-covering of a set ${\mY}$ with respect to a metric $\rho$ is a set $\{ \vtheta^{(1)}, \ldots, \vtheta^{(N)} \} \subset {\mY}$ such that for each $\vtheta \in {\mY}$, there exists some $i \in [N]$ such that $\rho(\vtheta, \vtheta^{(i)}) \leq \delta$. The $\delta$-covering number $N_{\text{cover}}(\delta, {\mY})$ is the minimal cardinality among all $\delta$-coverings.
\end{definition}

\begin{definition}[Set diameter] \label{app:def:diam}
    The \textit{diameter} of any subset $\mY\subseteq \R^m$ is defined by
\begin{equation}
\mathtt{diam}(\mY) = \max\{\rho(\vy^{(1)}, \vy^{(2)} )
\mid \vy^{(1)}, \vy^{(2)} \in \mY \}.
\label{app:eqn:diam}
\end{equation}
\end{definition}

\begin{definition}[Voronoi cells \cite{okabe2009spatial}]
    \label{app:def:vor}
    The Voronoi cells $\lbr{\mathcal{B}_1, \ldots, \mathcal{B}_N}$ of a finite set $\mathcal{S}\subseteq \mY$, where $\mathcal{S} = \lbr{\vy^{(1)}, \ldots, \vy^{(N)}}$, is defined by,
    \begin{equation}
        \mathcal{B}_i = \lbr{ \vy \; | \; \min_{y \in \mY} \rho(\vy, \vy^{(i)})}, \quad \forall i \in [N].
    \end{equation}
\end{definition}

\begin{definition}[Hausdorff measure]
    \label{app:def:hausdorff}
    Consider the metric space $(\R^{m},\rho)$ where $\rho(\cdot,\star)$ is the Euclidean distance. 
    The $d$-dimensional \textit{Hausdorff measure} of any Borel set $\mY\subseteq \R^m$ is 
    $$
    \mathcal{H}_d(\mY) = \lim_{\delta\rightarrow 0}\inf_{\mY\subseteq\cup_{i = 1}^\infty \mathcal{U}_i\atop \mathtt{diam} (\mathcal{U}_i)<\delta}\Bigg[\sum_{i = 1}^\infty \{\mathtt{diam} (\mathcal{U}_i)\}^d\Bigg],\quad d\in(0,m).
    $$ 
\end{definition}
 We use Hausdorff measure to further define the Hausdorff dimension of a set. We first introduce the following Theorem \ref{thm:hd} \cite{folland1999real}, which guarantees that, for any $\mY$, there exists at most one  $d^{\dagger}\in \R$ which makes the $d^\dagger$-dimensional Hausdorff measure of $\mY$ both non-zero and finite. We call this $d^\dagger$ the Hausdorff dimension of $\mY.$
    \begin{theorem}\label{thm:hd}
    For any Borel set $\mY\subseteq \R^m$, suppose that for some $d^\dagger$, $0<\mathcal{H}_{d^\dagger}(\mY) < \infty$. Then we have  $\mathcal{H}_{d'}(\mY) = 0$ for any $d' < d^\dagger$ and $\mathcal{H}_{d'}(\mY) = \infty$ for any $d' > d^\dagger$.
    \end{theorem}
\begin{definition}[Hausdorff dimension]
We say the \textit{Hausdorff dimension} of  $\mY$ is $d^\dagger$ if and only if, $0<\mathcal{H}_{d^\dagger}(\mY)<\infty.$
\end{definition}

\subsection{Notation Table}\label{app:n_table}
For clarity, we present a notation explanation in \Cref{app:tab:notation}.
\begin{table}[h]
    \centering
    \caption{Notations used in this paper.}
    \label{app:tab:notation}
    \begin{tabular}{lll}
    \toprule
    Notation & Meaning                    & Dimension \\ 
    \midrule
    $N$ & Number of solutions. & - \\
    $n$ & Dimension of a solution. & - \\
    $m$ & Number of objectives. & - \\
    $\vx$        & An MOO solution.                 & $n$         \\
    $\mathscr{X}$        & The decision space. $\vx \in \mathscr{X}$ & $\mathbb{R}^n$         \\
    $\vy, \vf(\vx)$  & The objective vector of a solution $\vx$. & $m$ \\
    $\mathscr{Y}$        & The objective space.                 & $\mathbb{R}^m$         \\
    $\mathcal{S}$        & A set of objectives, $\mathcal{S} =\{\vy^{(1)}, \ldots, \vy^{(N)}\}$ .                 & $m$         \\
    $\mathcal{T}$      & The Pareto front. & $\mathbb{R}^{m}$ \\
    $\delta$      & The minimal separation distance of a set belong to $\mathcal{T}$ (c.f. \Cref{eqn:pack_pf}). &  \\
    $\delta_v$      & The maximal diameter of all Voronoi cells. 
    (c.f. \Cref{app:eqn:delta_v}). &  \\
    $\vlam$      & The weight vector. & $m$ \\
    $\vtheta(\vlam), \vtheta$      & The angular parameterization of vector $\vlam$ (c.f. \Cref{app:eqn:pref_angle}).  & $m-1$ \\
    $g^{\text{alg}}(\cdot | \vlam)$ & The multiobjective aggregation function.  & $\mathbb{R}^m \mapsto \mathbb{R}$ \\
    $\vh(\vlam)$ & The function maps a weight to a Pareto solution. $\vh(\vlam): \vOmega \mapsto \mathbb{R}^m$.  &  \\
    $\vh_\vphi(\cdot)$      & The PFL model. $\vh_\vphi(\cdot): \mbr{0, \frac{\pi}{2}}^{m-1} \mapsto \mathbb{R}^m$  &  \\
    $\mathcal{H}_d(\cdot)$ & Hausdorff dimension function. & 
    \\         
    \bottomrule
    \end{tabular}
\end{table}

\subsection{Algorithm} \label{app:sec:alg}
The total algorithms run as Algorithm \ref{alg_1} and \ref{alg_2}. The proposed algorithm mainly adopt the MOEA/D framework. For simplicity, we use the Simulated Binary Crossover (SBX) \cite{deb1995simulated} and poly-nominal mutation mutation operators \cite{deb2014analysing}. It is possible to use more advanced MOEA/D framework, e.g., MOEA/D with differential evolution \cite{tan2012modification}, which is left as future works. 

As summarized by Algorithm 1, the proposed approach dynamically adjusts the weight angles on the estimated PF learned by the current objectives. The updated weights are then set as the new weights for the MOEA/D algorithm. 

\begin{algorithm}[H]
  \SetAlgoLined
  \caption{Training of PFL and weight adjustment. }
  \label{alg_1}
  \textbf{Input:} weight angle set $\mTheta_N$ and objective set $\mathcal{S}$ from MOEA/D.  \\
    \# Training the PFL model. \\
    \For{i=1:$N_{\text{pfl}}$}{
        $\vphi \leftarrow \vphi - \tilde{\eta} \nabla_\vphi l_\text{pfl}$.
    }
    \# Solving the maximal separation problem at the estimated PF (\Probref{eqn:pack_pf_hat}) by gradient ascent. \\
    \For{i=1:$N_{\text{opt}}$}{
        $\mTheta_N \leftarrow \text{Proj}(\mTheta_N + \eta \nabla_{ \mTheta_N } \hat{\delta}_{\hat{\Tau}})$.
    }
    \textbf{Output:} The updated weight angles $\{\vtheta^{(1)}, \ldots, \vtheta^{(N)}\}$.
\end{algorithm}

\begin{algorithm}[H]
  \SetAlgoLined
  \caption{MOEA/D with uniform adaptive weight adjustment (UMOEA/D). } \label{alg_2}
  \textbf{Input: Initial $N$ weight $\lambda_N:\lambda_N=\lbr{\vlam^{(1)}, \ldots, \vlam^{(N)}}$ by \cite{das1998normal}, the initial solution set $\mathcal{X}_N: \mathcal{X}_N=\lbr{\vx^{(1)}, \ldots, \vx^{(N)}}$, the MOO objective function $\vf(\cdot)$.} \\
    \For{k=1:K}{
        \For{i=1:$N_{ \text{inner} }$}{
            \# Step 1. Evolutionary algorithm. \\
            \For{j=1:N}{
                \# Generate a crossover solution from neighbourhoods of $\vx^{(j)}$ using SBX operator. \\
                $\vx^{(j)} \leftarrow \text{SBX}(\vx^{(j_1)}, \vx^{(j_2)})$, where $\vx^{(j_1)}, \vx^{(j_2)}$ are selected randomly from the neighborhood set of $\vx^{(j)}$.\\
                \# Mutation the solution by the polynomial mutation operator. \\
                $\vx^{(j)} \leftarrow \text{Mutate}(\vx^{(j)})$. 
            }
            \# Update the solution each sub-problems by elites. \\
            \For{j=1:N}{
                $\vx^{(j)} \leftarrow \argmin_{i \in B(j) \cup \{\vx^{(j)}\}} g^\text{mtche}( \vf(\vx^{i}), \vlam^{(j)})$. 
                \# $B(j)$ is the neighborhood index set \cite{zhang2007moea} of solution $\vx^{(j)}$. \\
            }
        }
        \# Step 2. weight adjustment. \\
        Calculate the weight angle set $\mTheta_N$ from weights by \Eqref{app:eqn:pref_angle}. \\
        $\mathcal{S} = f \circ \mathcal{X}_N$. \\
        $\mTheta_N \leftarrow$ Algorithm1($\mTheta_N$, $\mathcal{S}$). \\
        Update the weight vector $\vlam^{(1)}, \ldots, \vlam^{(N)}$ by \Eqref{app:eqn:pref_angle}. 
    }
\end{algorithm}

We briefly analyze the running complexity of the proposed method. The main complexity is inherited from MOEA/D \cite{zhang2007moea}. The addition of the PFL model training and the uniformity optimization introduces two additional parts. 

Training the PFL model is a standard supervised learning problem, hence the complexity is proportional to the number of objectives $m$ and sample numbers $N$. The overall complexity is $\mathcal{O}(m N \cdot N_\text{pfl} )$.   

The uniformity optimization involves calculating the lower (or upper) triangular matrix of an adjacency matrix, which has a complexity of $\mathcal{O}(m \cdot \frac{N(N-1)}{2}) = \mathcal{O}(mN^2)$. Therefore, the total complexity of the optimization process is $\mathcal{O}(m N_\text{opt} \cdot \frac{mN(N-1)}{2}) = \mathcal{O}(mN^2 \cdot N_\text{opt})$.

Practically, training time of the PFL and weight adjustment is less than \texttt{1s}, which is neglectable compared with MOEAs. The calculation of the adjacency matrix and the MOEA/D algorithm can be executed in parallel, which can further improve the efficiency of the overall running time.
\paragraph{(The relation between weight angle $\vtheta(\vlam)$ and weight $\vlam$.)} 
Given $\vtheta(\vlam) \in [0, \frac{\pi}{2}]^{m-1}$ as a parameter representation of $\vlam \in \vDelta_{m-1}$, the weight angle and the corresponding weight vector can be converted using the following equations:
\begin{equation} \label{app:eqn:pref_angle}
    \left \{
    \begin{split}
    \sqrt{\lambda_1} &= \cos( \theta_1 )  \\
    \sqrt{\lambda_2} &= \sin(\theta_1) \cos(\theta_2)  \\
    \sqrt{\lambda_3} &= \sin(\theta_1) \sin(\theta_2) \cos(\theta_3) \\
    & \vdots \\
    \sqrt{\lambda_m}  &= \sin(\theta_1) \sin(\theta_2) \ldots \sin(\theta_{m-1}). 
    \end{split}
    \right .  
\end{equation}

For a given weight vector $\vlam$, computing the weight angle can be achieved by solving \Cref{app:eqn:pref_angle}. Similarly, $\vtheta$ can be solved from $\vlam$ by the following equation, 
\begin{equation} \label{app:eqn:pref_angle_inverse}
    \left \{
    \begin{split}
        \theta_1 &= \arg \cos( \sqrt{\lambda_1} ) \\
        \theta_2 &= \arg \cos \sbr{ \frac{\sqrt{\lambda_2}}{\sin \theta_1} } \\
        \theta_3 &= \arg \cos \sbr{ \frac{\sqrt{\lambda_3}}{\sin \theta_1 \sin \theta_2} } \\
        & \vdots \\
        \theta_{m-1} &= \arg \cos \sbr{ \frac{\sqrt{\lambda_{m-1}}}{\sin \theta_1 \sin \theta_2 \ldots \sin \theta_{m-1}} } \\
    \end{split}
    \right .  
\end{equation}

\section{Experiments Details}
\subsection{Comparison Methods} \label{app:sec:comp_mtds}
We give a detailed elaboration of the comparison methods used in the experiments as follows. The code for the proposed method will be made publicly available after publication.

\begin{enumerate}
\item The vanilla \textbf{MOEA/D} method \cite{zhang2007moea} employs diverse distributed weight vectors to explore a diverse Pareto solution set. However, the uniformity observed in the weight space may not lead to uniformity in the objective space, resulting in a coarse level of solution diversity sought by MOEA/D.

\item The MOEA/D with adaptive weight adjustment ( \textbf{MOEA/D-AWA} ) \cite{qi2014moea,de2018moea}. MOEA/D-AWA is an improvement over the vanilla MOEA/D, which aims to improve the 
replaces the most crowded solution with the most sparse solution. A detailed comparison of MOEA/D-AWA and the proposed method can be found in \Cref{app:sec:moead_awa_illus}.

Since the source code for their implementation was not publicly available, we implemented it by ourselves. In the original implementation of MOEA/D-AWA, they maintain an external population (EP). However, as the modern MOEA frameworks (e.g., pymoo, platemo) are no longer dependent on EP, we employ a neural network surrogate model to predict the most sparse solution \cite{qi2014moea}. 

\item The Pareto adaptive weight (\textbf
{PaLam}) method \cite{siwei2011multiobjective} approximates the PF using a simple math mode $y_1^p + y_2^p=1$ and generates uniform Pareto objectives by utilizing the hypervolume indicator \cite{guerreiro2020hypervolume}. Since real-world problems often exhibit complex PFs, to ensure fairness, we employ a neural model for training to predict the true PF instead of relying on a simple mathematical model. We use the code in \url{https://github.com/timodeist/multi_objective_learning} to develop a new gradient-based algorithm for {pa}$\mathbf{\vlam}$ to achieve fast optimization for the \texttt{HV} indicator predicted by neural networks. Our improved version of the vanilla pa$\vlam$ significantly outperforms its original implementation.

\item The \textbf{SMS-EMOA} \cite{beume2007sms} method, which uses the hypervolume indicator as the guidance for the multiobjective evolutionary optimizations. The code for SMS-EMOA directly follows the pymoo library.

\item The \tb{MOEA/D-GP} \cite{wu2017adaptive} method. This method employs Gaussian process (GP) modeling to capture the shape of the PF and utilizes the objective function $\sum_{1\leq i \neq j \leq N} \frac{1}{d_{ij}}$, where $d_{ij}$ represents the distance from objective $y^{(i)}$ to $y^{(j)}$, to promote the generation of uniformly distributed Pareto objectives. In comparison to MOEA/D-GP, our method offers two significant advantages. Firstly, our method employs more accurate neural models to characterize the preference-to-objective mapping function. This allows for a more precise representation of the relationship between preferences and objectives, leading to enhanced modeling capabilities. Additionally, the adapted uniform indicator in our method is optimized more efficiently. Instead of calculating the gradients of all objectives, we only need to compute the gradients of two objectives. This optimization technique improves computational efficiency without sacrificing the quality of the results. Furthermore, our method provides superior theoretical guarantees than MOEA/D-GP.

\item The \tb{MOEA/D-L} \cite{dong2020moea} method. This method suggests utilizing linear chains as a means to model the shape of the PF. However, due to their limited representational capacity, linear models may not be well-suited for capturing the complexities of complex and high-dimensional PFs. Consequently, the performance of MOEA/D-L falls short compared to our proposed UMOEA/D method.
\end{enumerate}

\subsection{Metrics}
\label{app:sec:metrics}
In order to measure the performance of our proposed method, we employ the following metrics to measure uniformity and the solution quality. The up-arrow ($\uparrow$)/down-arrow($\downarrow$) signifies that a higher/lower value of this indicator is preferable.
\begin{enumerate}
    \item The \texttt{hypervolume} ($\uparrow$) (\texttt{HV}) indicator \cite{guerreiro2020hypervolume}, which serves as a measure of the convergence (the distance to the PF) and only a \textit{coarse} measure of the sparsity/uniformity level of a solution set. 
    
    \item The \texttt{sparsity} $(\downarrow)$ indicator \cite{xu2020prediction}, which measures the sum of distance of a set of solutions in the non-dominated sorting order \cite{deb2002fast}.
    
    \item The \texttt{spacing} ($\downarrow$) indicator \cite{schott1995fault}, which measures the uniformity of a set of solutions. It is quantified as the standard deviation of the distance set $\{d_1, \ldots, d_N\}$.
    \begin{equation}
        \texttt{spacing}({\mathcal{S}}) = \text{std} (d_1, \ldots, d_N), 
    \end{equation}
    where $d_i = \min_{j \in [m], j \neq i} \rho(\vy^{(i)}, \vy^{(j)})$, serving as the minimal distance from solution $\vy^{(i)}$ to the rest of objectives.
    \item The $\delta$ ($\uparrow$) and $\tilde{\delta}$ ($\uparrow$) indicators represent the (soft) minimal distances among different solutions within a solution set.
    \begin{equation}
        \tilde{\delta} = - \frac{1}{K} \log \sum_{1 \leq i \neq j \leq N} \exp{(-K \cdot \rho(\vy^{(i)}, \vy^{(j)}))}. 
    \end{equation}
    A large $\delta$/$\tilde{\delta}$ indicator means that, any different solution pairs are far away from each other. Noted that, it is possible that $\tilde{\delta} < 0$. $K$ is a positive constant.  
\end{enumerate}

\subsection{Results on all problems} \label{app:sec:res_all}
\begin{figure*}[h!]
    \centering
    \subfloat[AWA]{\includegraphics[width = \qwidth \linewidth]{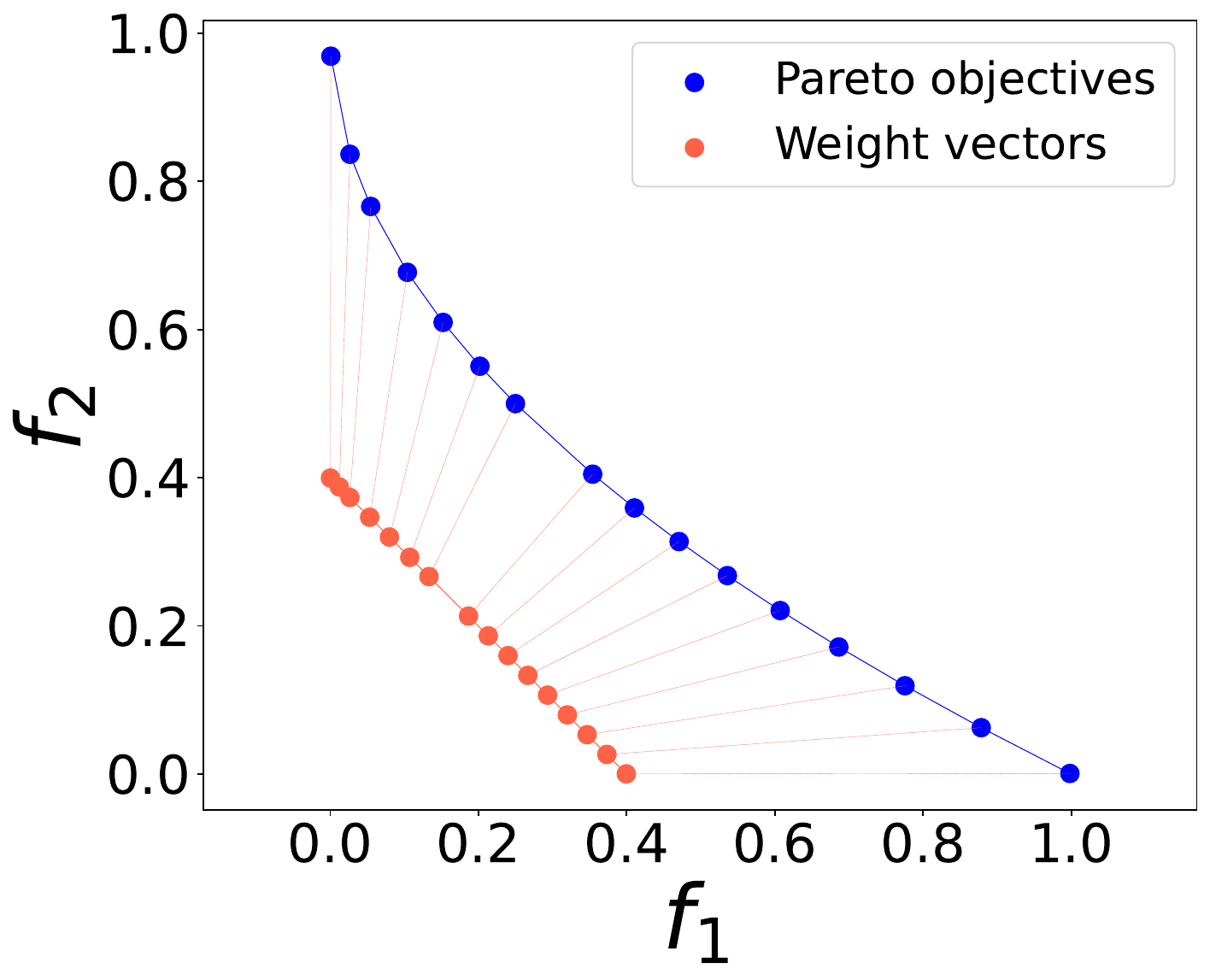}}
    \hfill
    \subfloat[GP]{\includegraphics[width = \qwidth \linewidth]{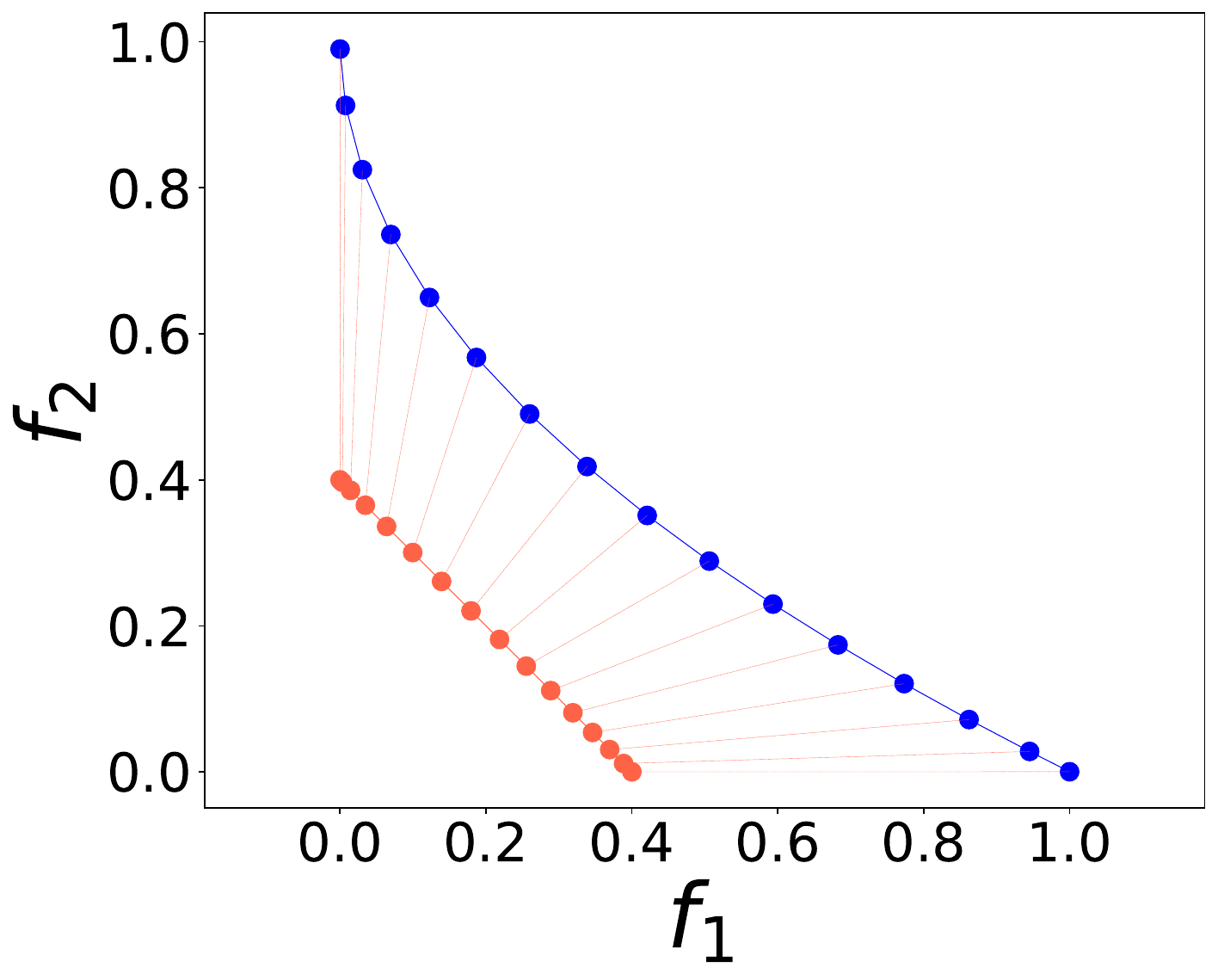}}
    \hfill
    \subfloat[MOEA/D-L]{\includegraphics[width = \qwidth \linewidth]{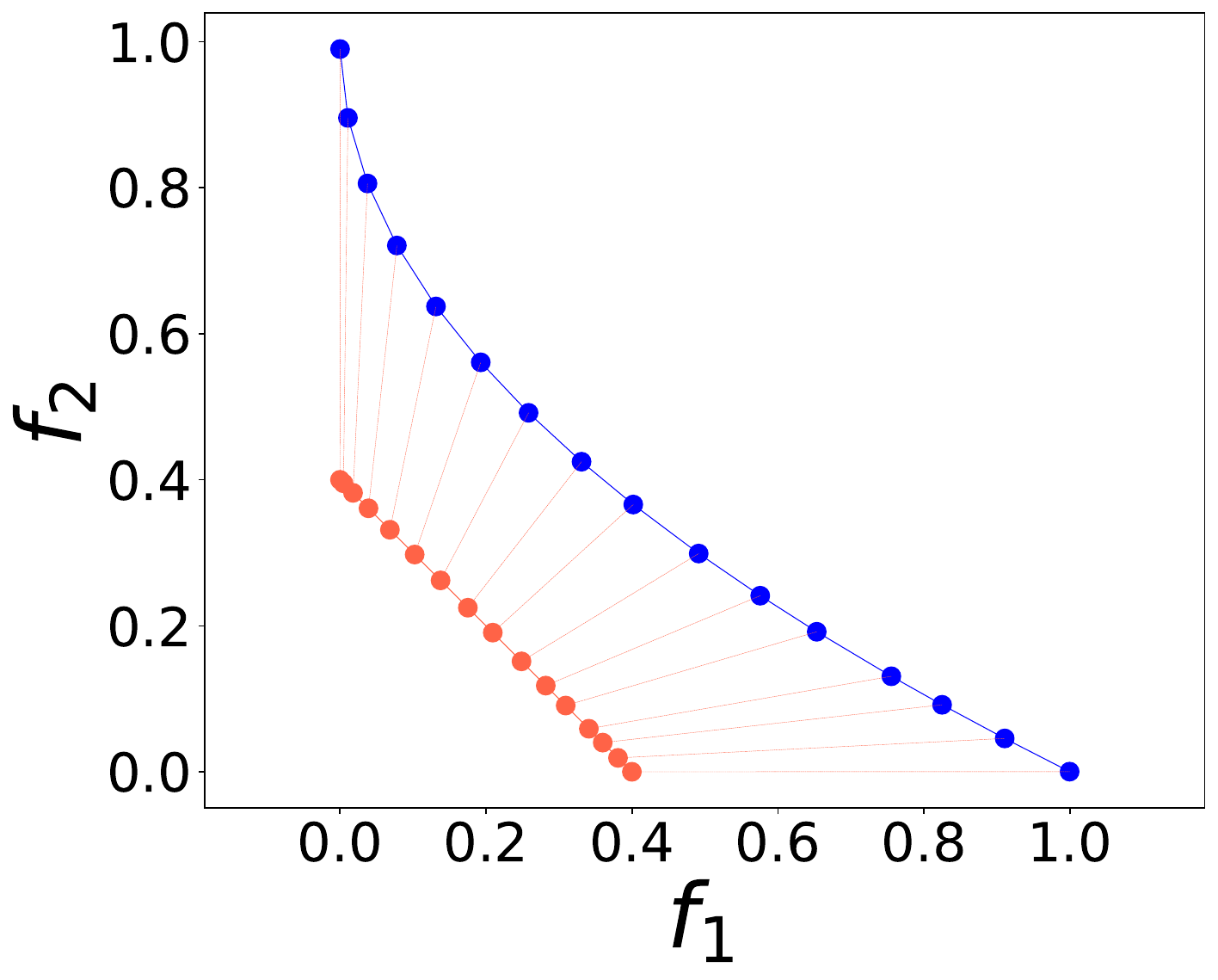}}
    \hfill
    \subfloat[MOEA/D]{\includegraphics[width = \qwidth \linewidth]{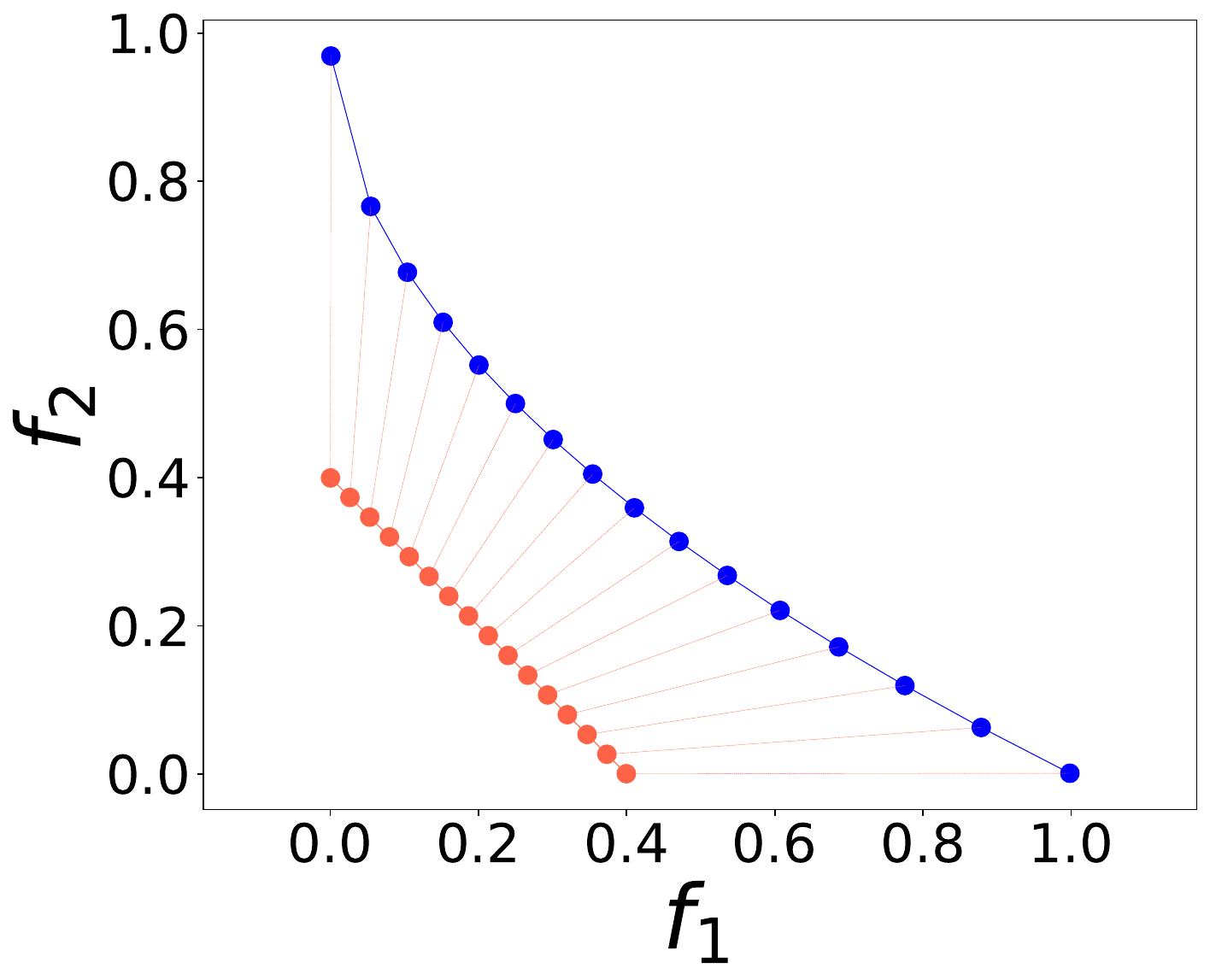}}
    \hfill
    \subfloat[PaLam]{\includegraphics[width = \qwidth \linewidth]{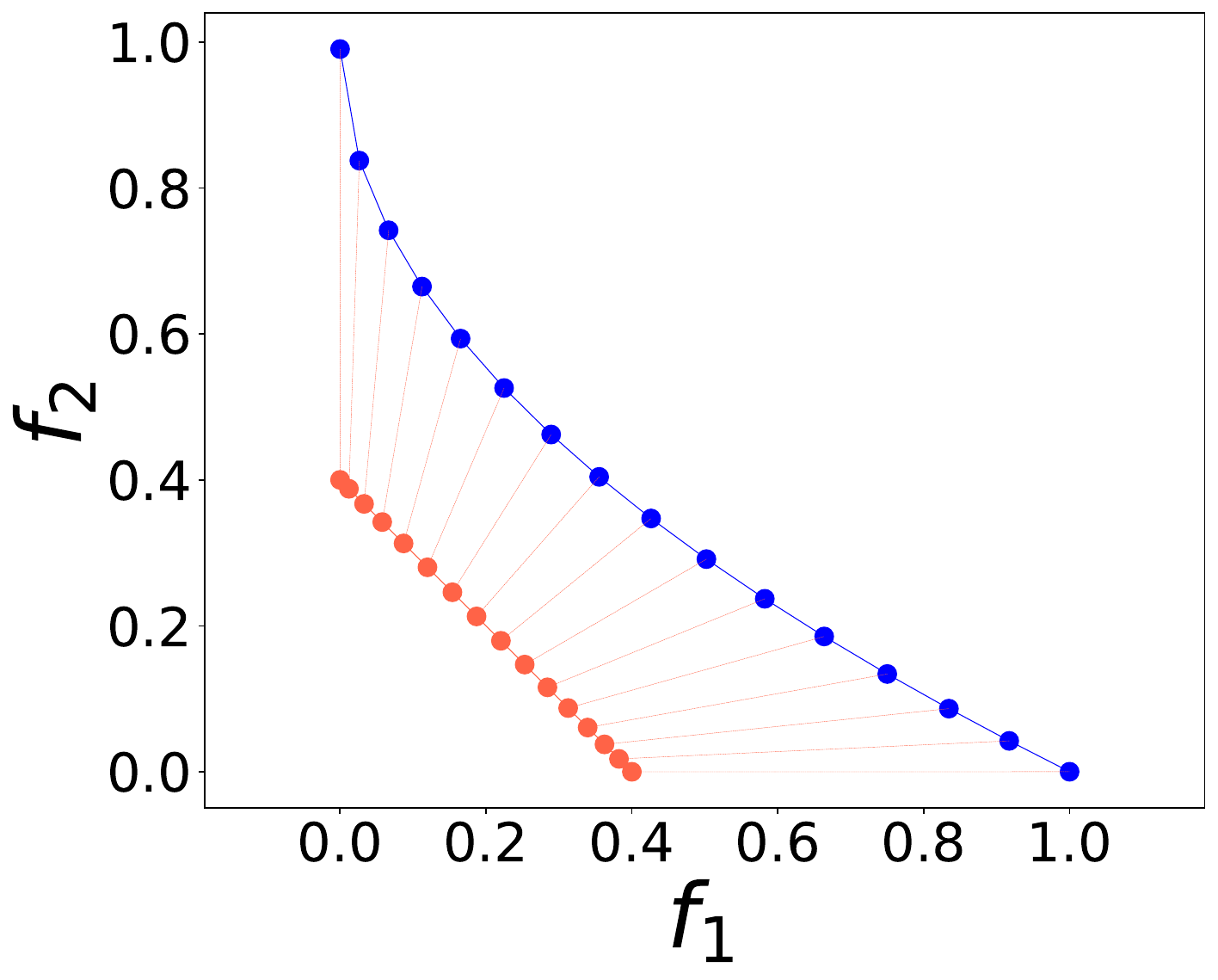}}
    \hfill
    \subfloat[SMS]{\includegraphics[width = \qwidth \linewidth]{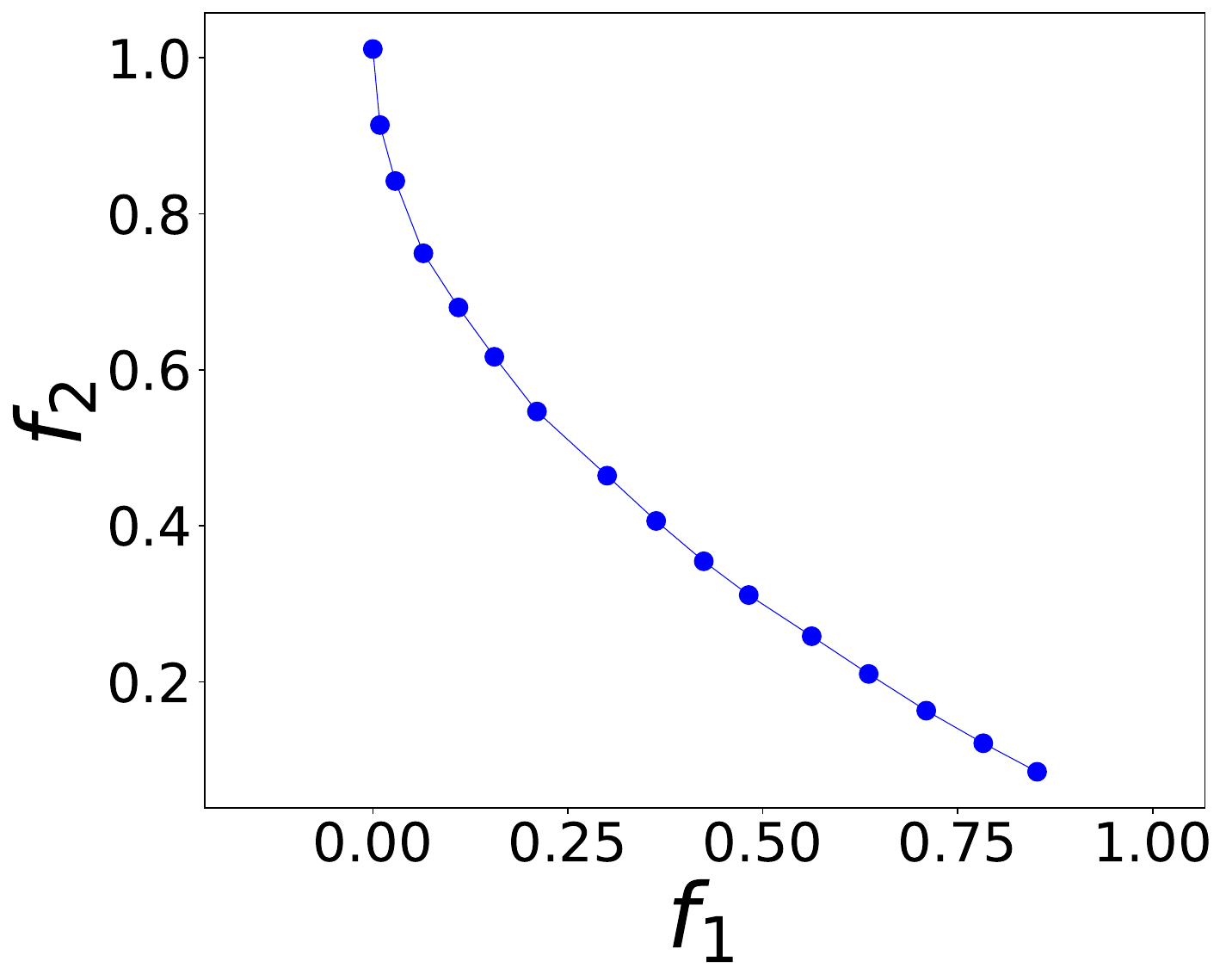}}
    \hfill
    \subfloat[UMOEA/D]{\includegraphics[width = \qwidth \linewidth]{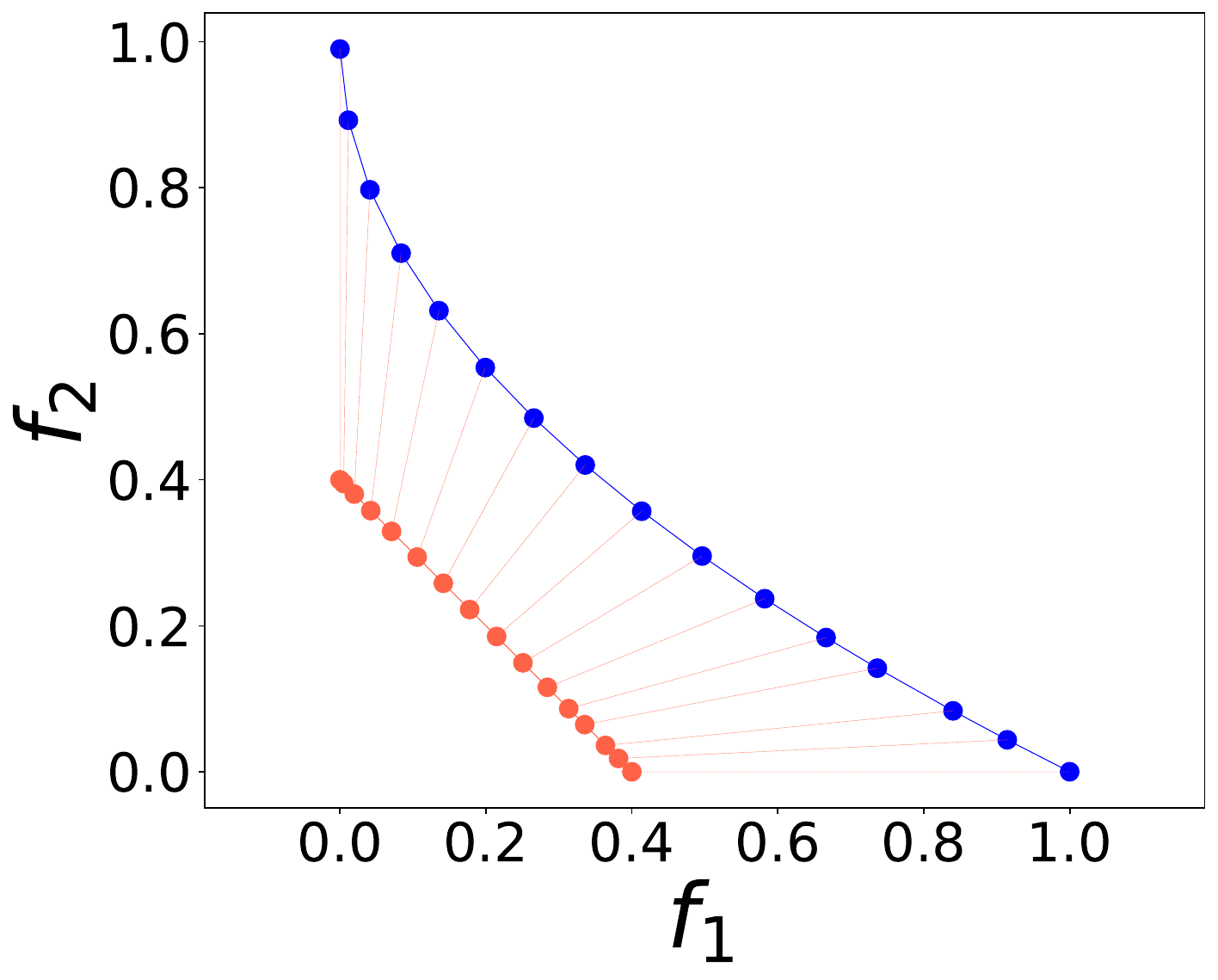}}
    \hfill
    \caption{Results on ZDT1.} \label{fig:2obj_zdt1}
    \subfloat[AWA]{\includegraphics[width = \qwidth \linewidth]{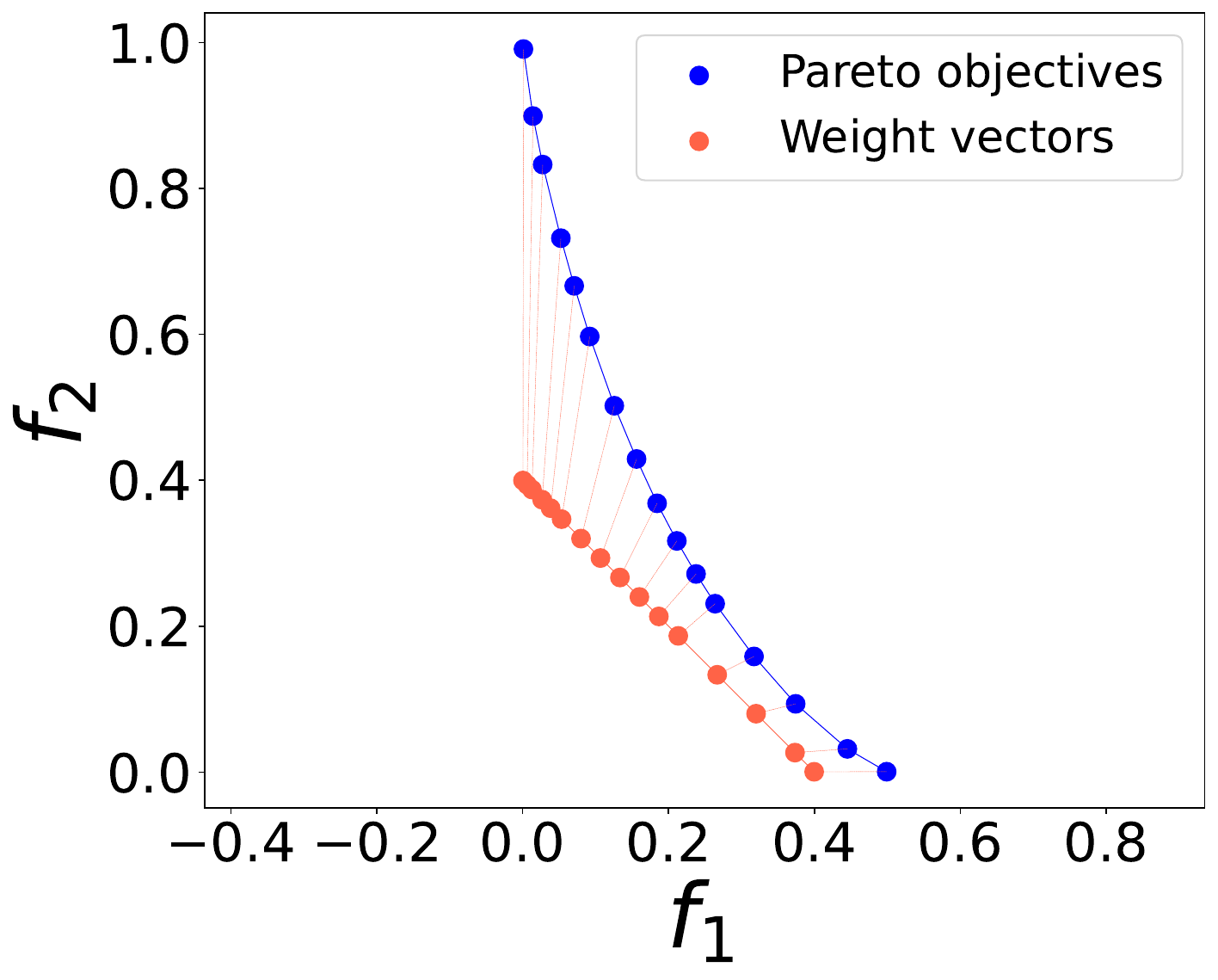}}
    \hfill
    \subfloat[GP]{\includegraphics[width = \qwidth \linewidth]{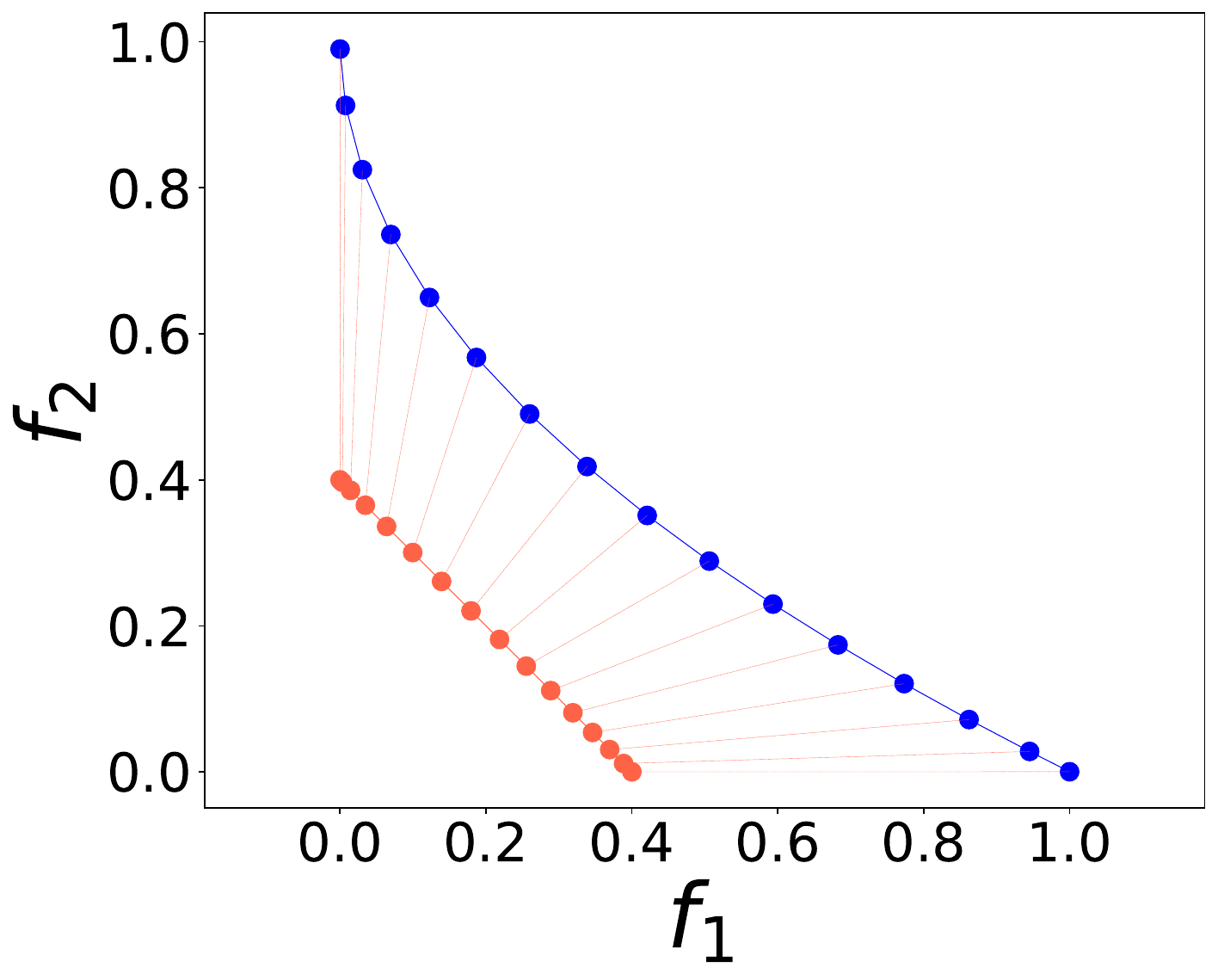}}
    \hfill
    \subfloat[MOEA/D-L]{\includegraphics[width = \qwidth \linewidth]{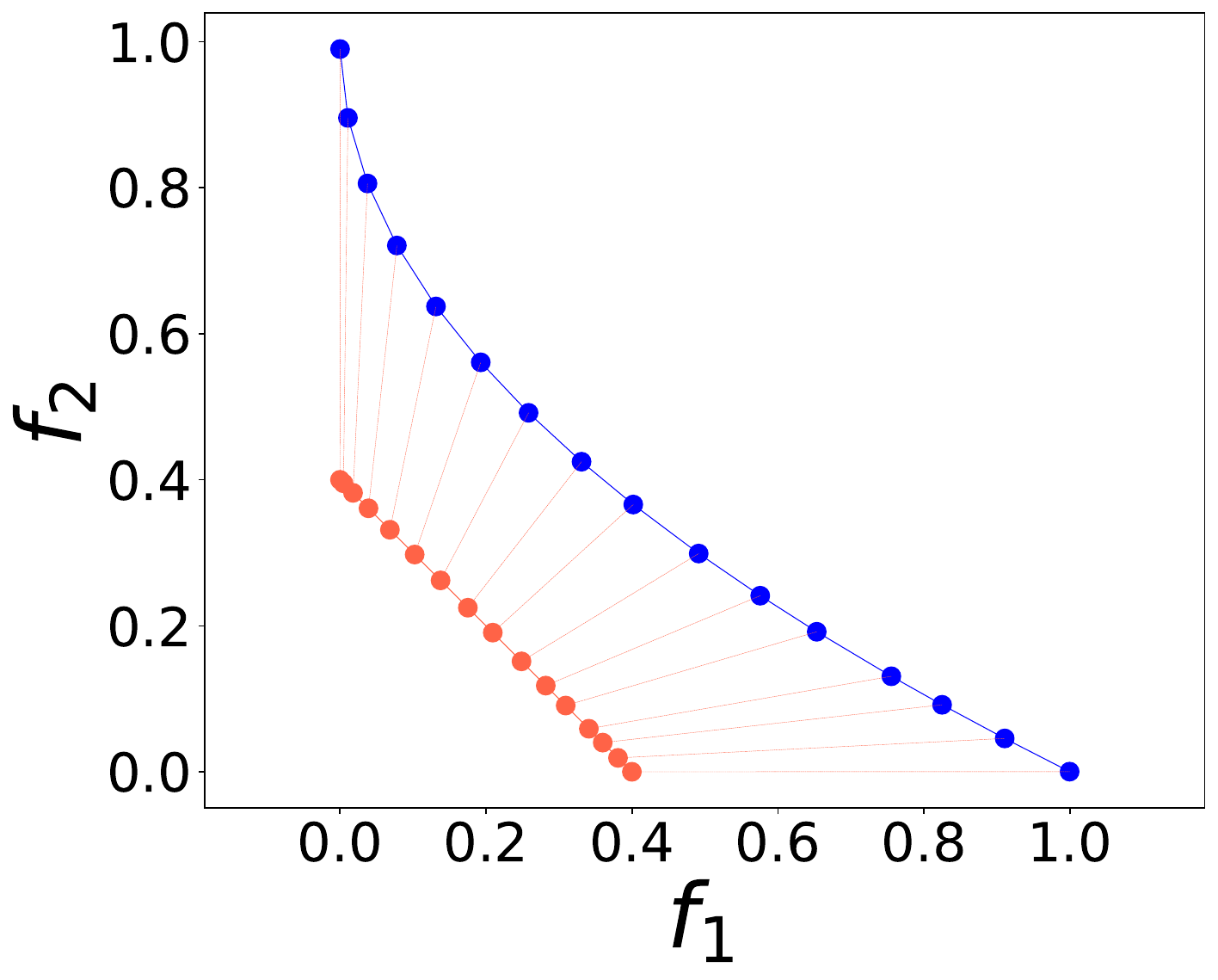}}
    \hfill
    \subfloat[MOEA/D]{\includegraphics[width = \qwidth \linewidth]{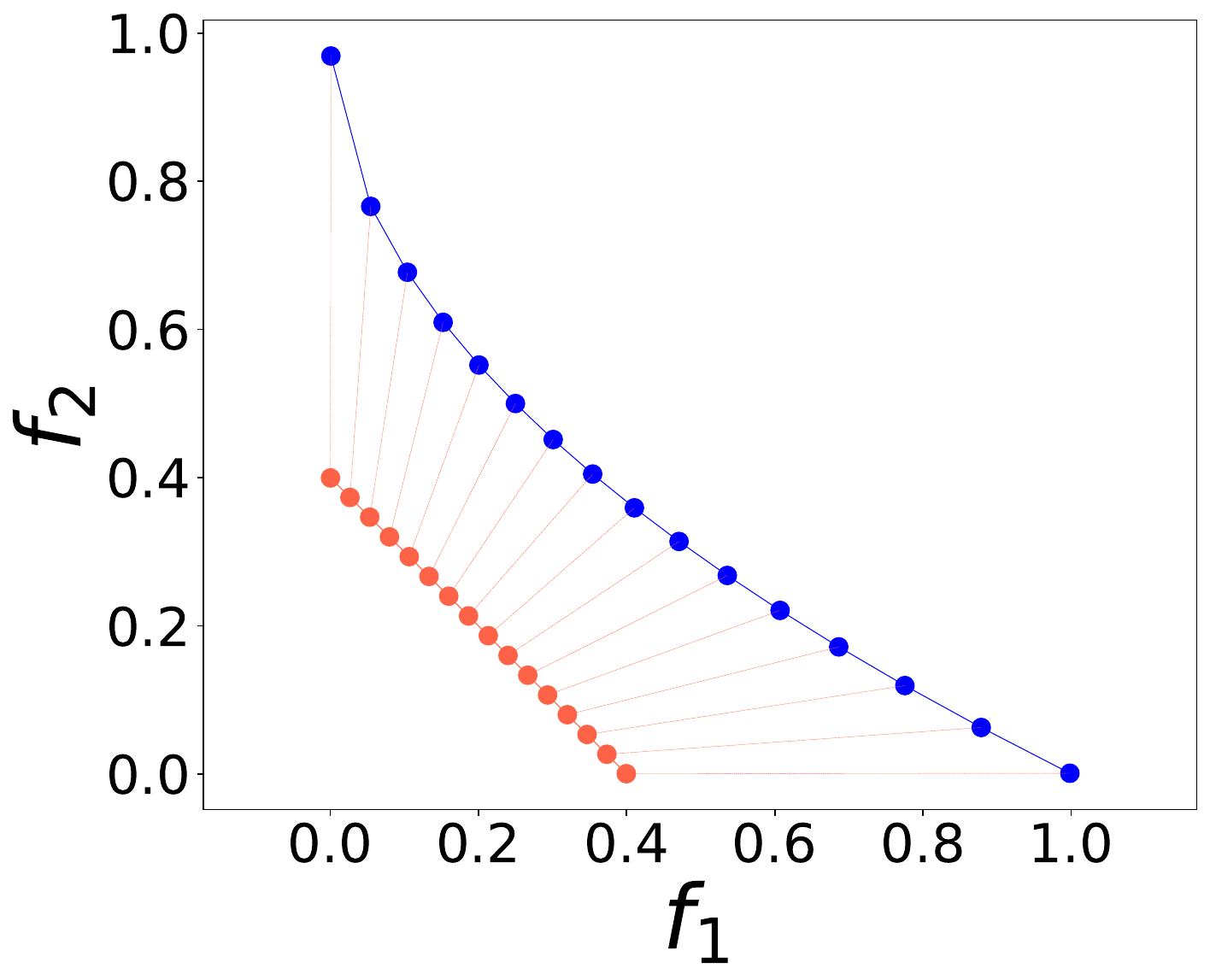}}
    \hfill
    \subfloat[PaLam]{\includegraphics[width = \qwidth \linewidth]{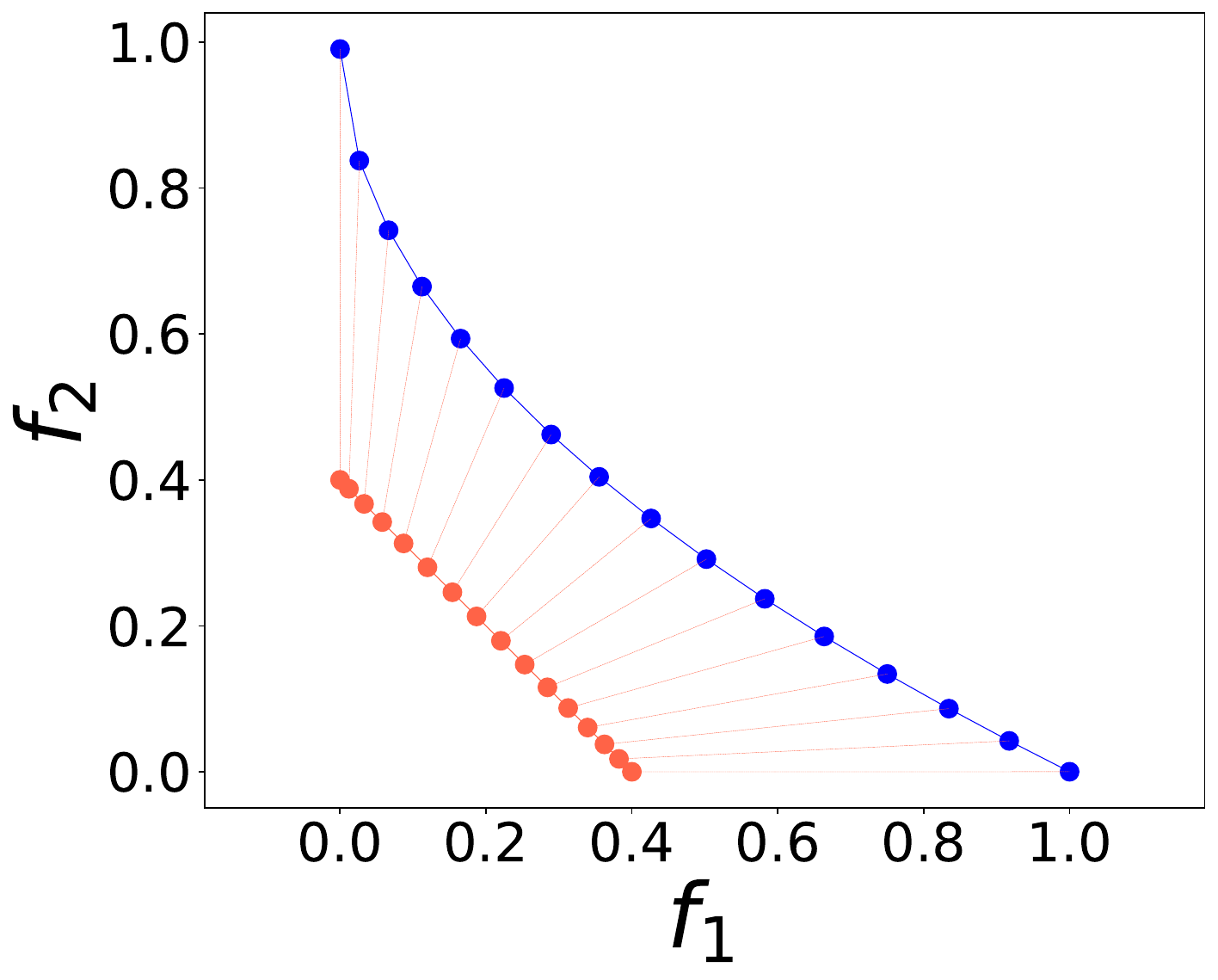}}
    \hfill
    \subfloat[SMS]{\includegraphics[width = \qwidth \linewidth]{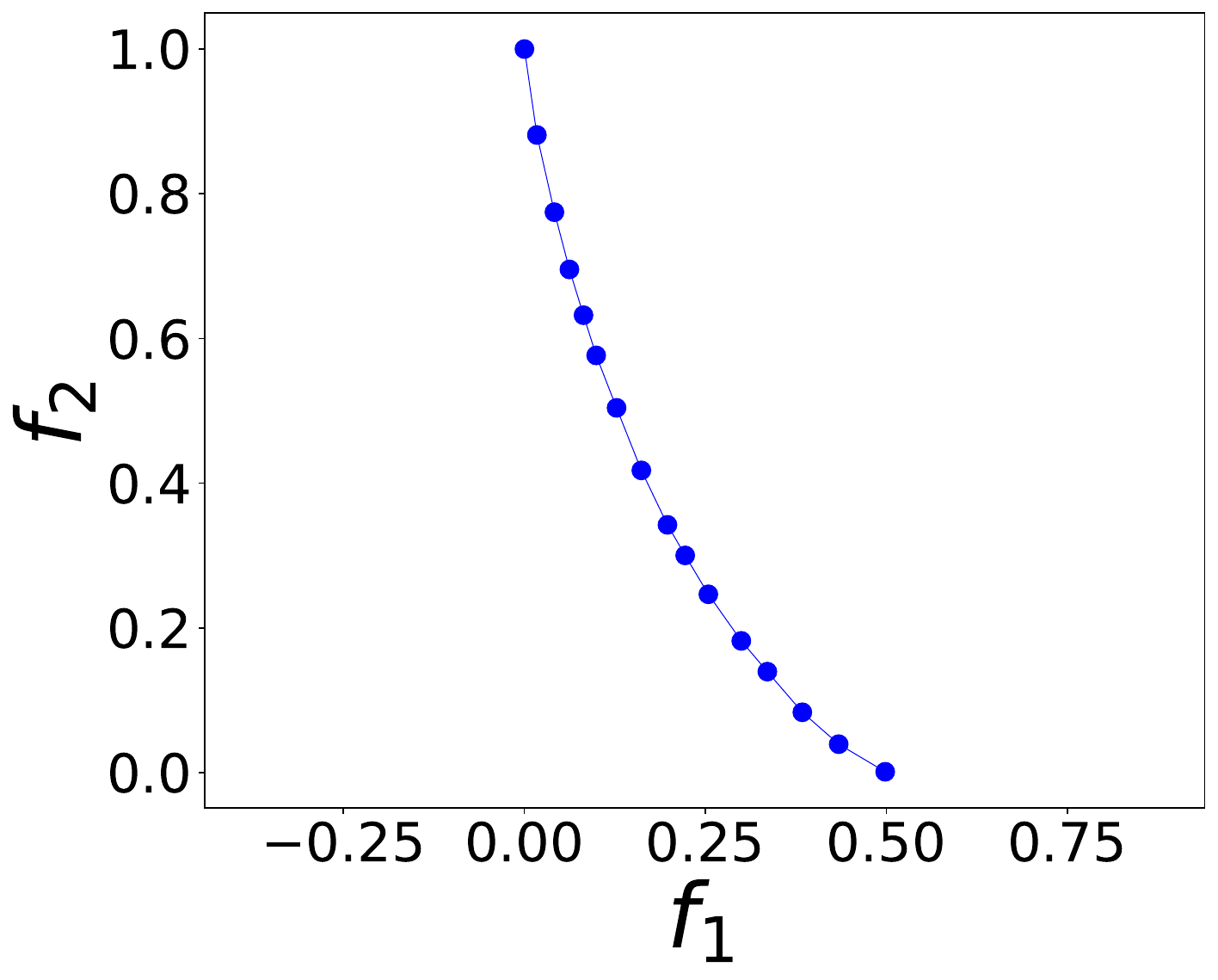}}
    \hfill
    \subfloat[UMOEA/D]{\includegraphics[width = \qwidth \linewidth]{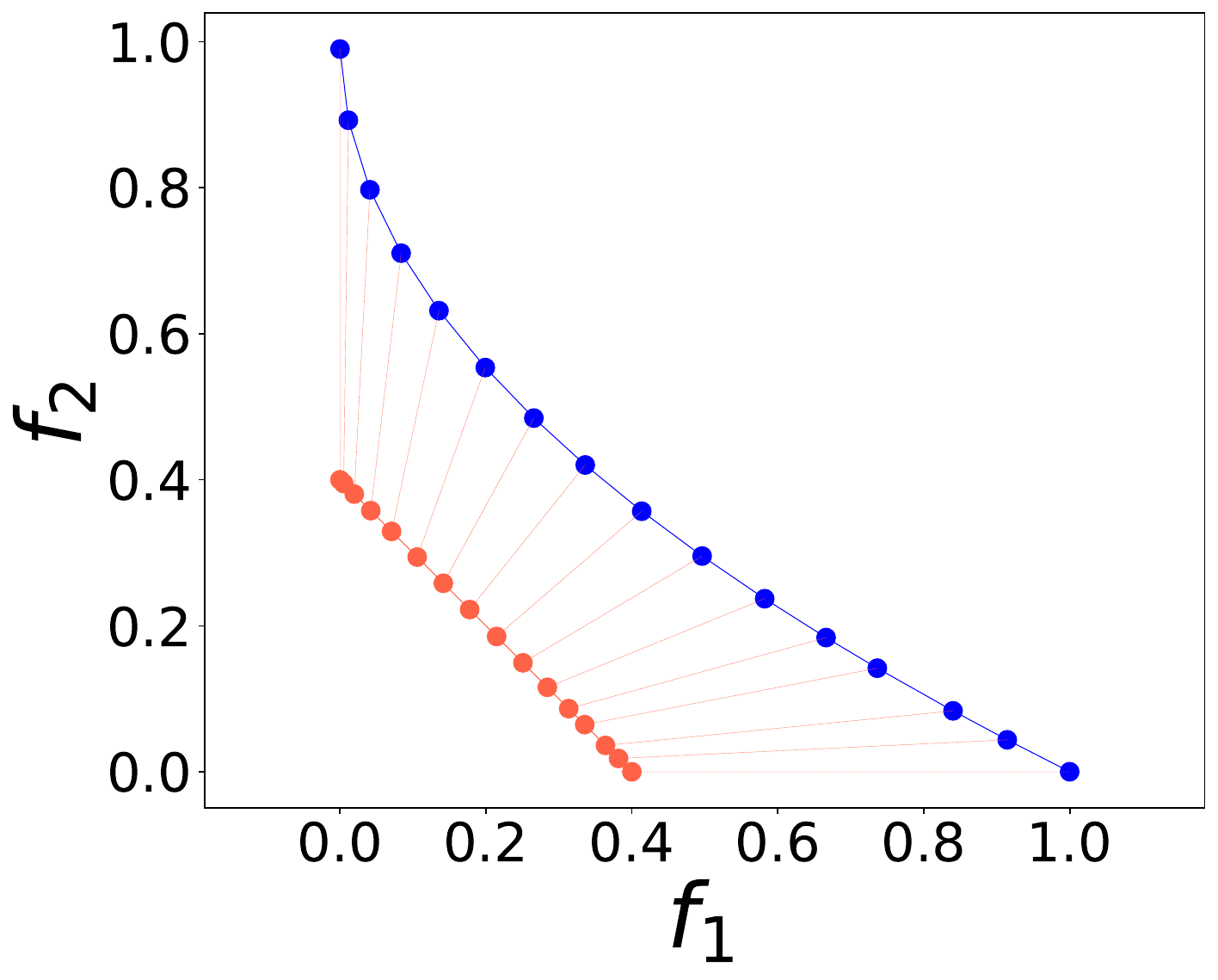}}
    \hfill
    \caption{Results on RE21. } 
    \label{fig:2obj_re21}
\end{figure*}

\begin{figure*}[h!]
    \centering
    \subfloat[AWA]{\includegraphics[width = \hwidth \linewidth]{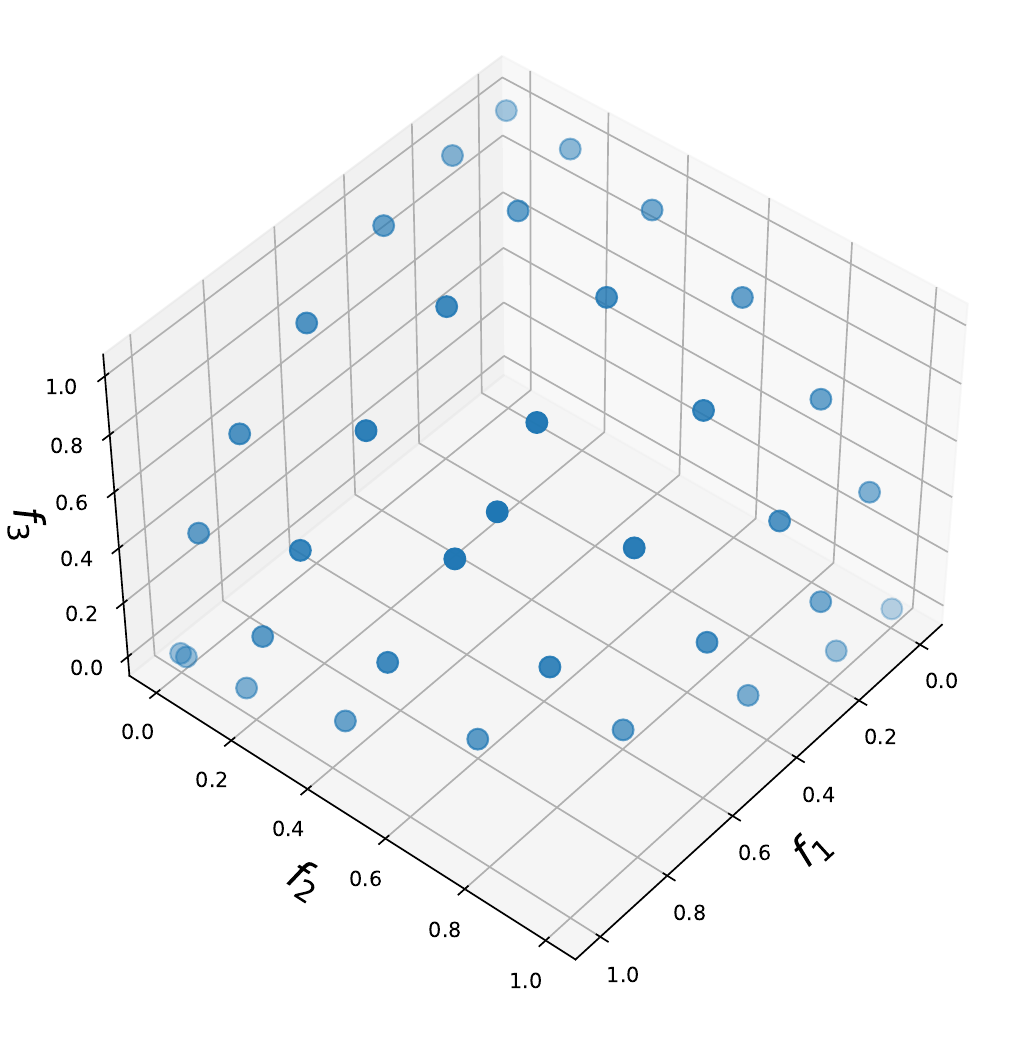}}
    \hfill
    \subfloat[GP]{\includegraphics[width = \hwidth \linewidth]{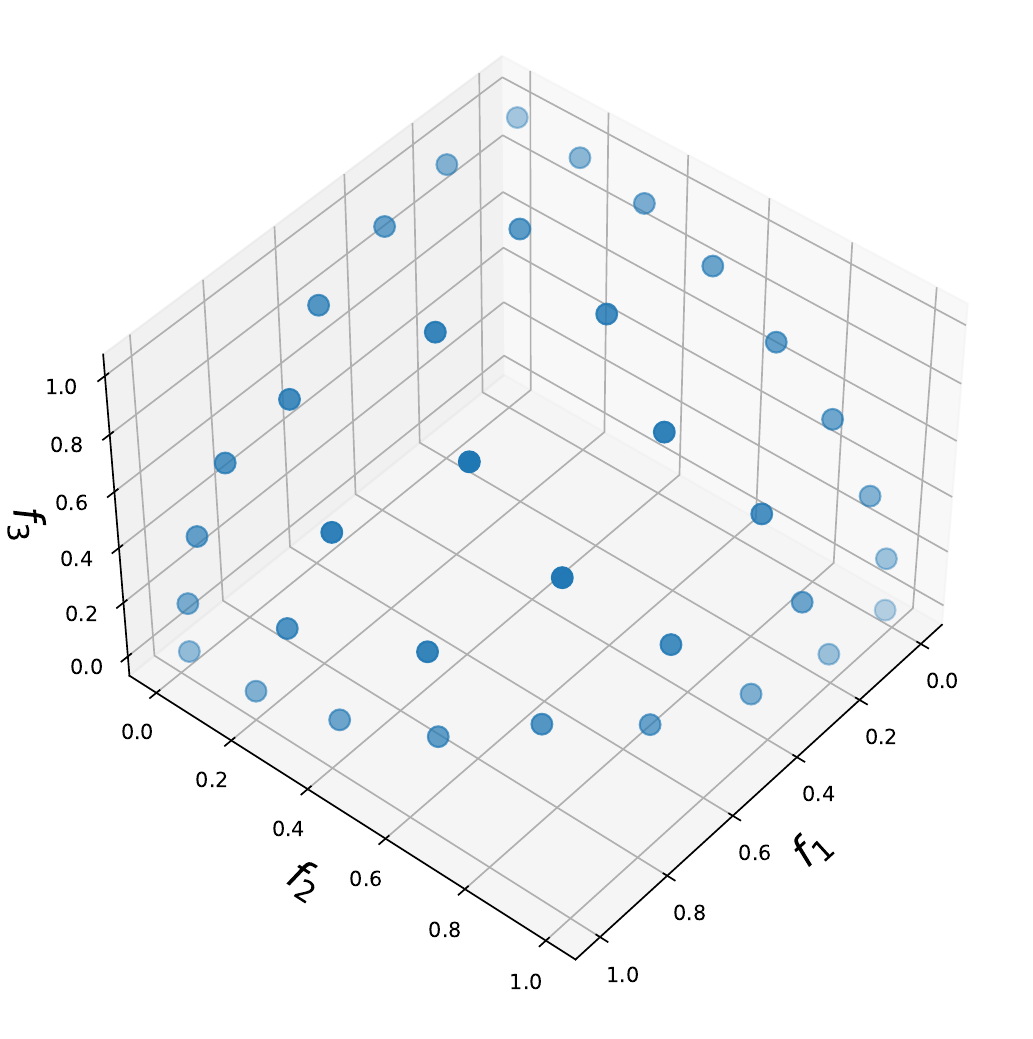}}
    \hfill
    \subfloat[MOEA/D]{\includegraphics[width = \hwidth \linewidth]{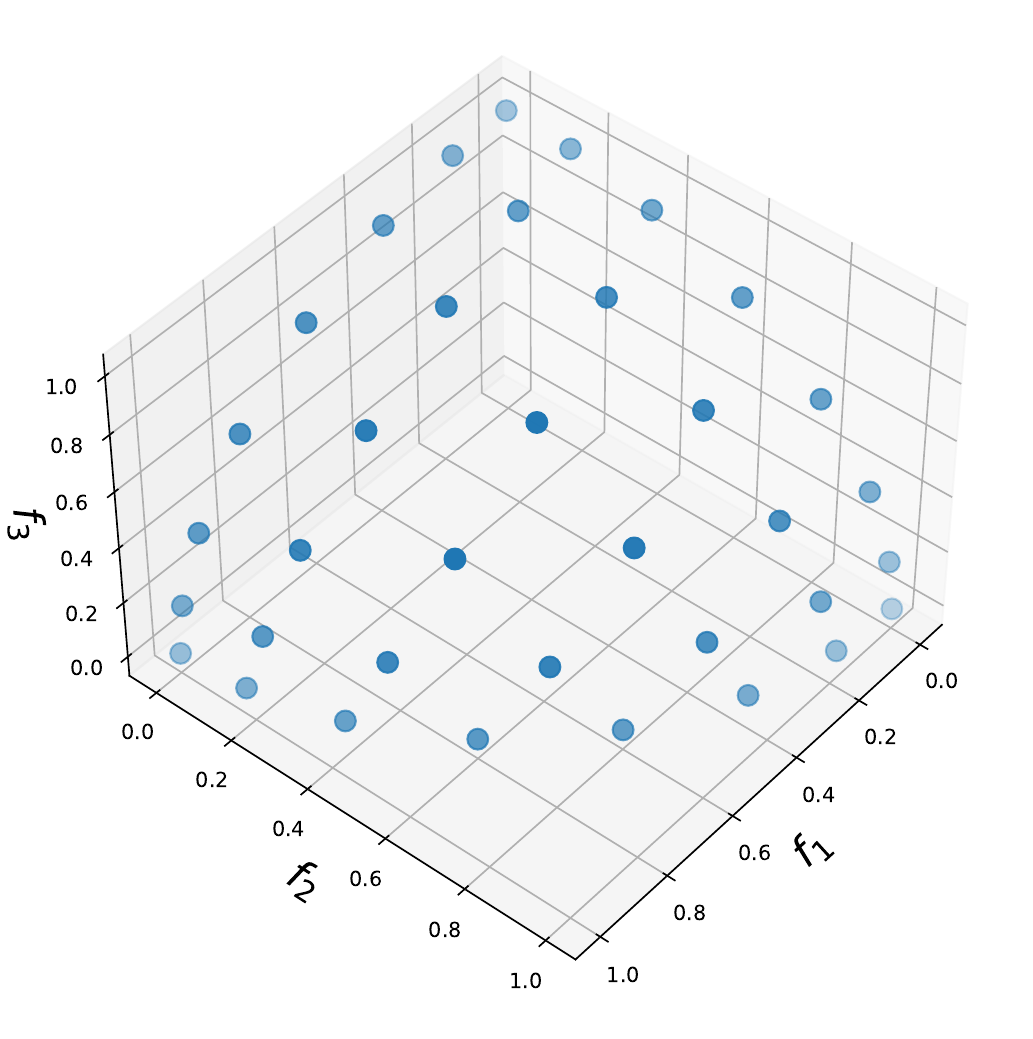}}
    \hfill
    \subfloat[PaLam]{\includegraphics[width = \hwidth \linewidth]{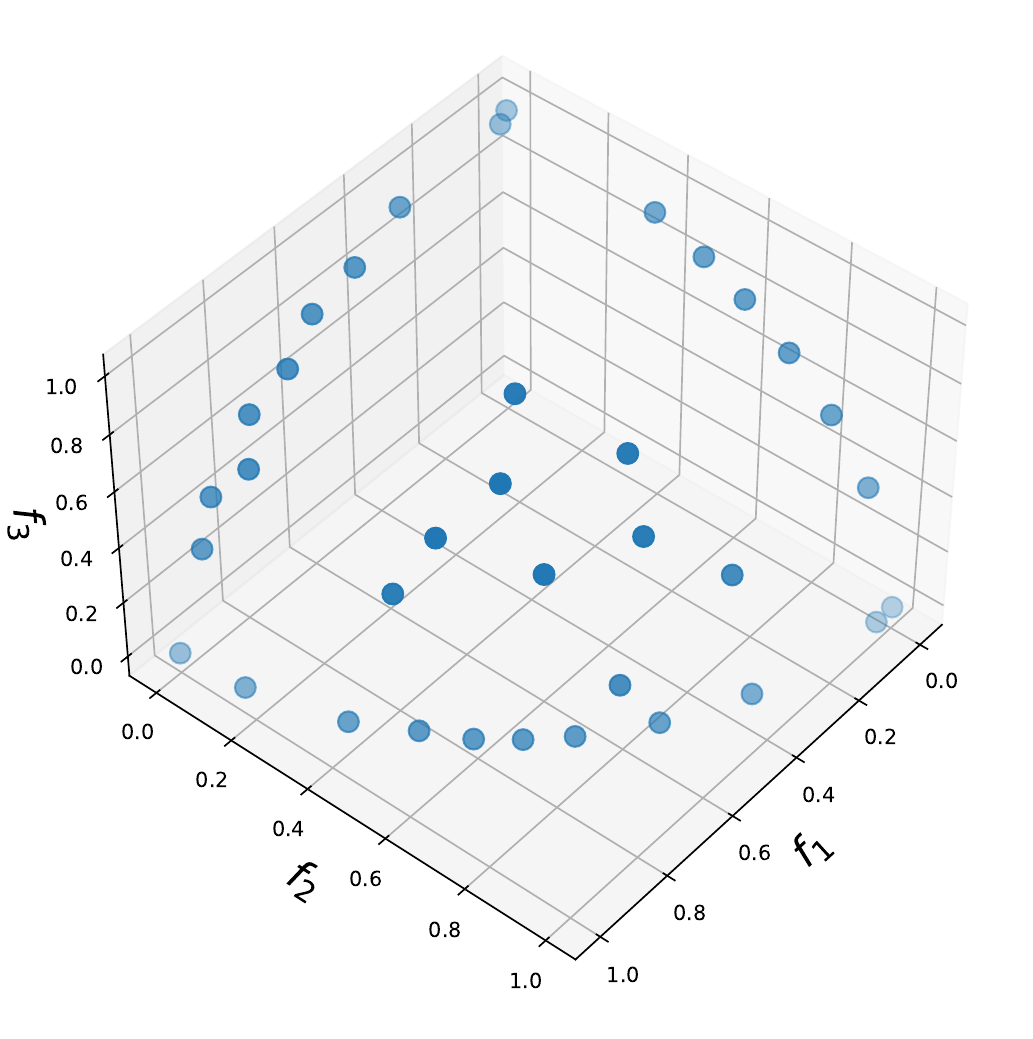}}
    \hfill
    \subfloat[SMS]{\includegraphics[width = \hwidth \linewidth]{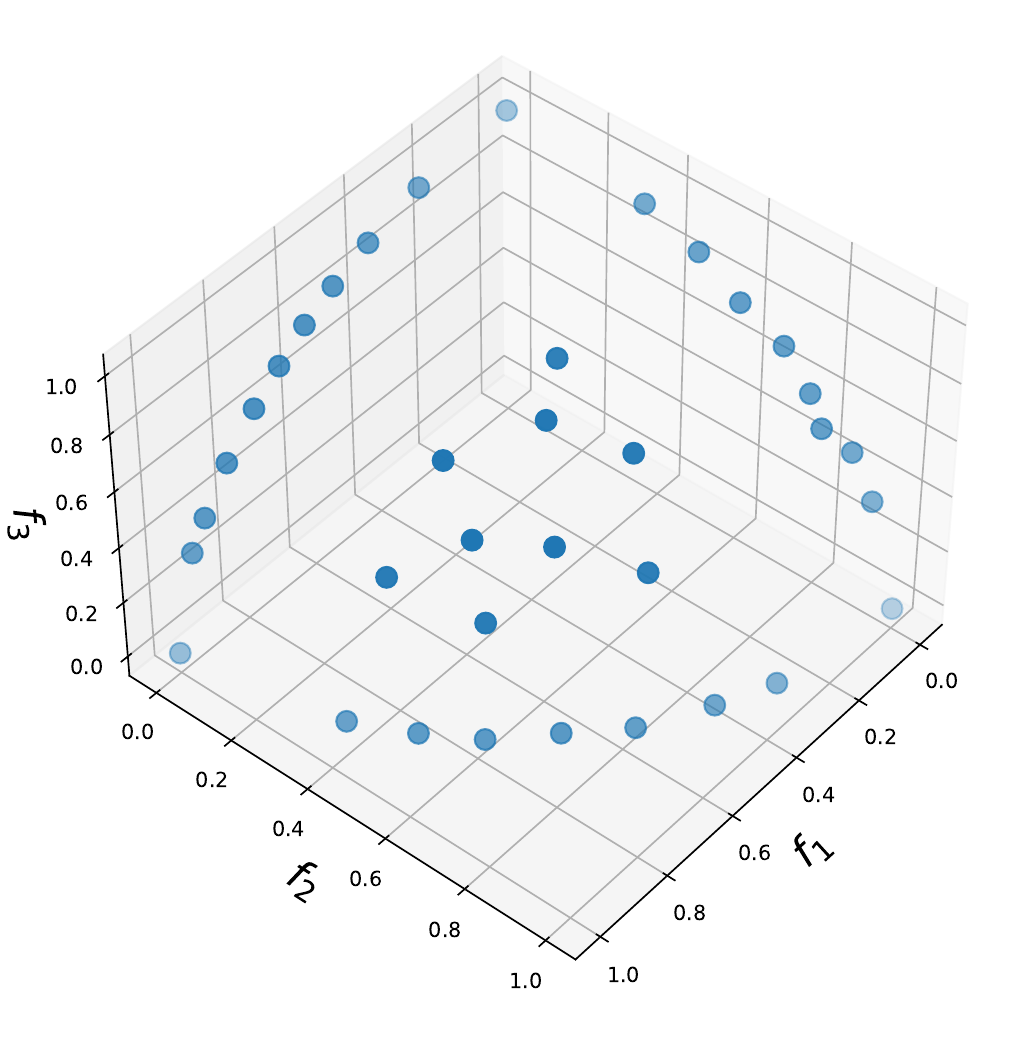}}
    \hfill
    \subfloat[UMOEA/D]{\includegraphics[width = \hwidth \linewidth]{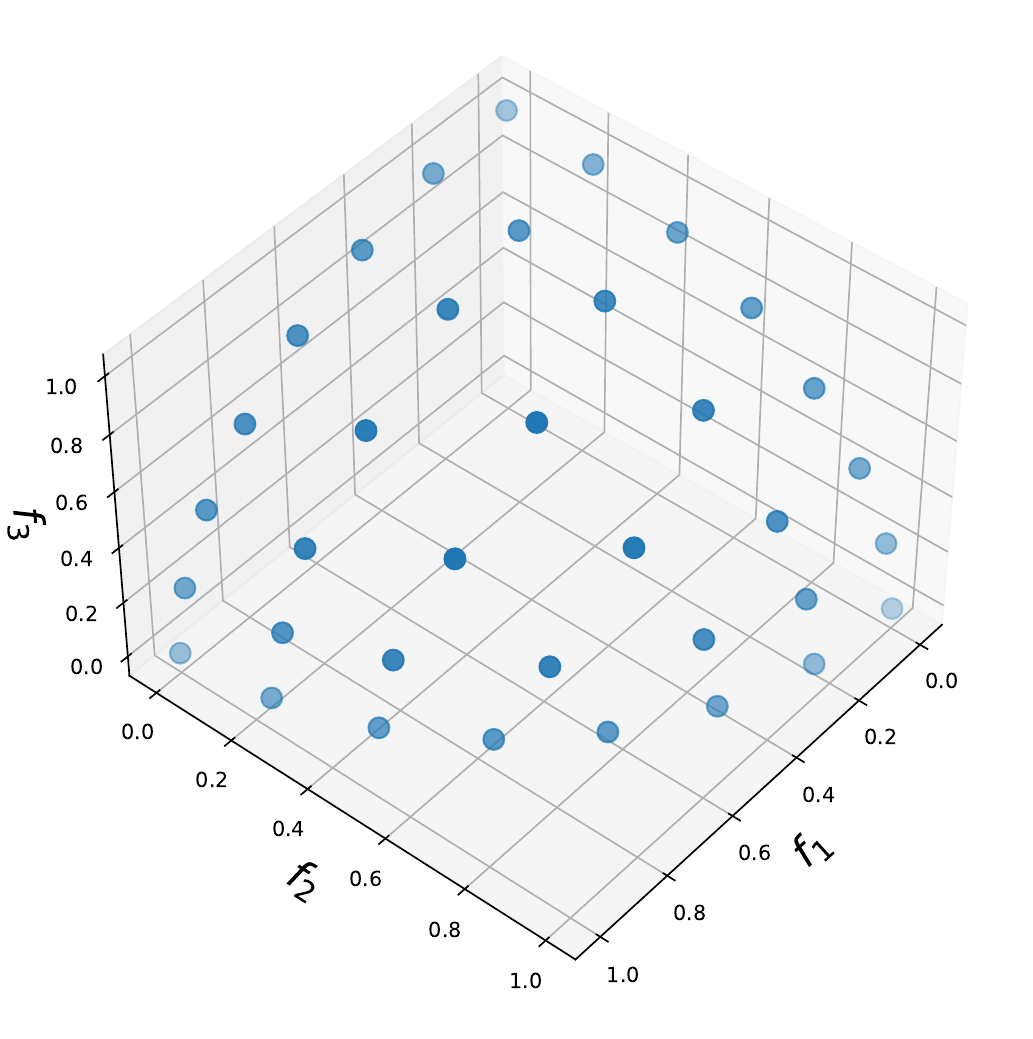}}
    \hfill
    \caption{Results on DTLZ2.}
    \label{fig:3obj_dtlz2}
    \subfloat[AWA]{\includegraphics[width = \hwidth \linewidth]{Figure/main/RE37/awa/final.pdf}}
    \hfill
    \subfloat[GP]{\includegraphics[width = \hwidth \linewidth]{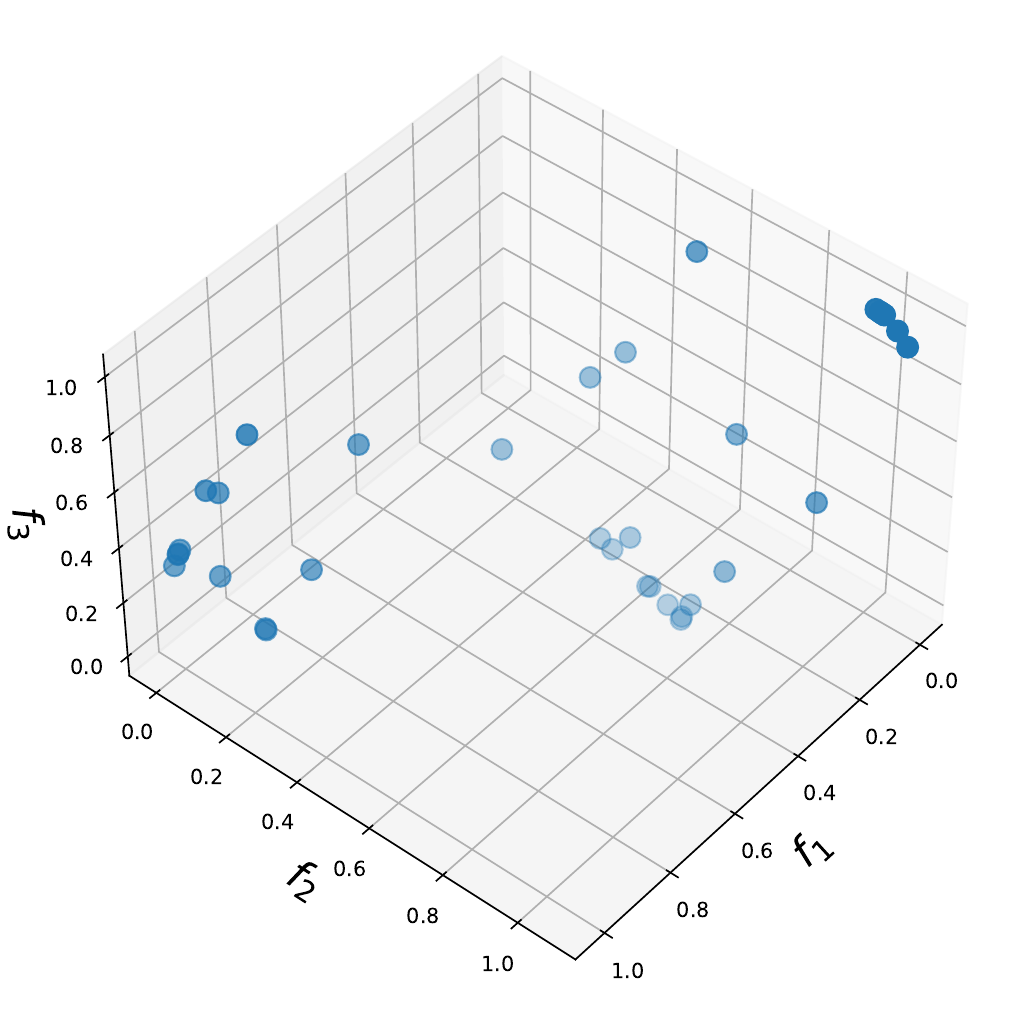}}
    \hfill
    \subfloat[MOEA/D]{\includegraphics[width = \hwidth \linewidth]{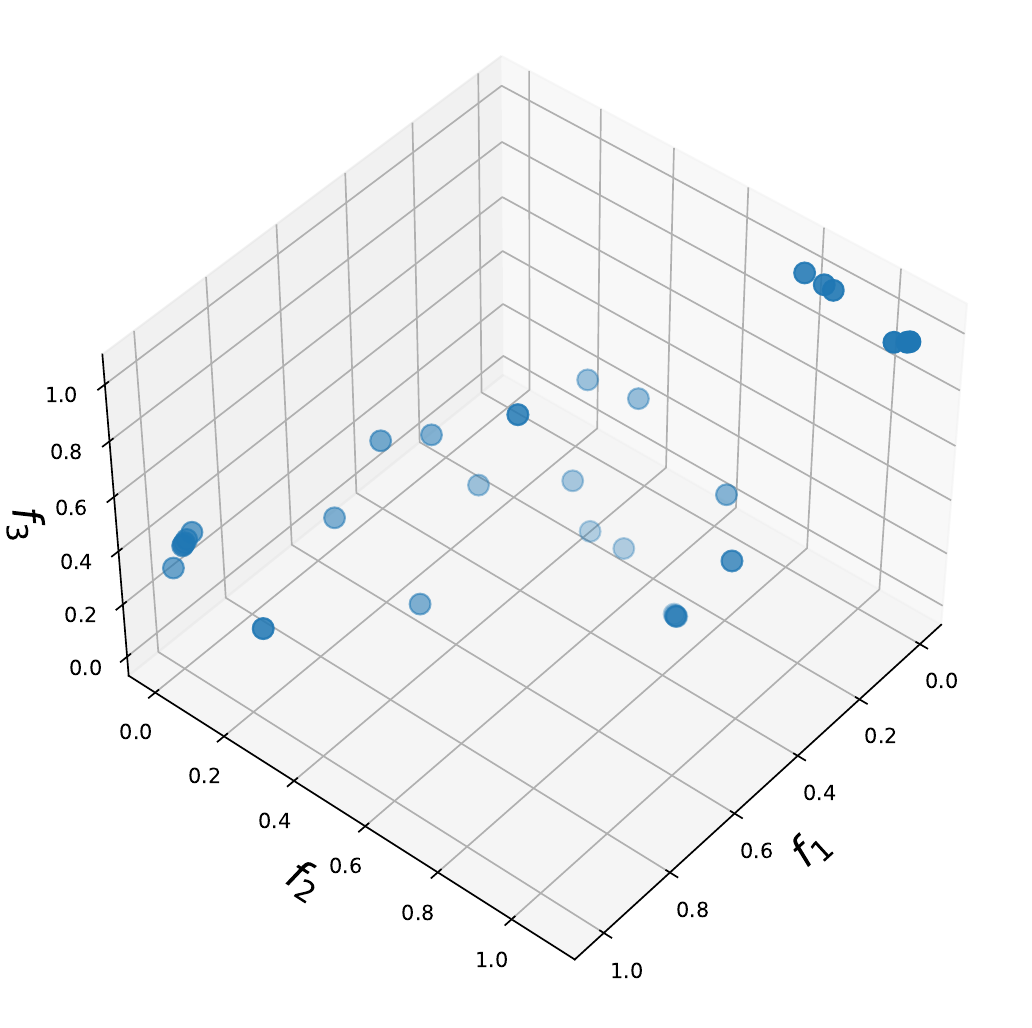}}
    \hfill
    \subfloat[PaLam]{\includegraphics[width = \hwidth \linewidth]{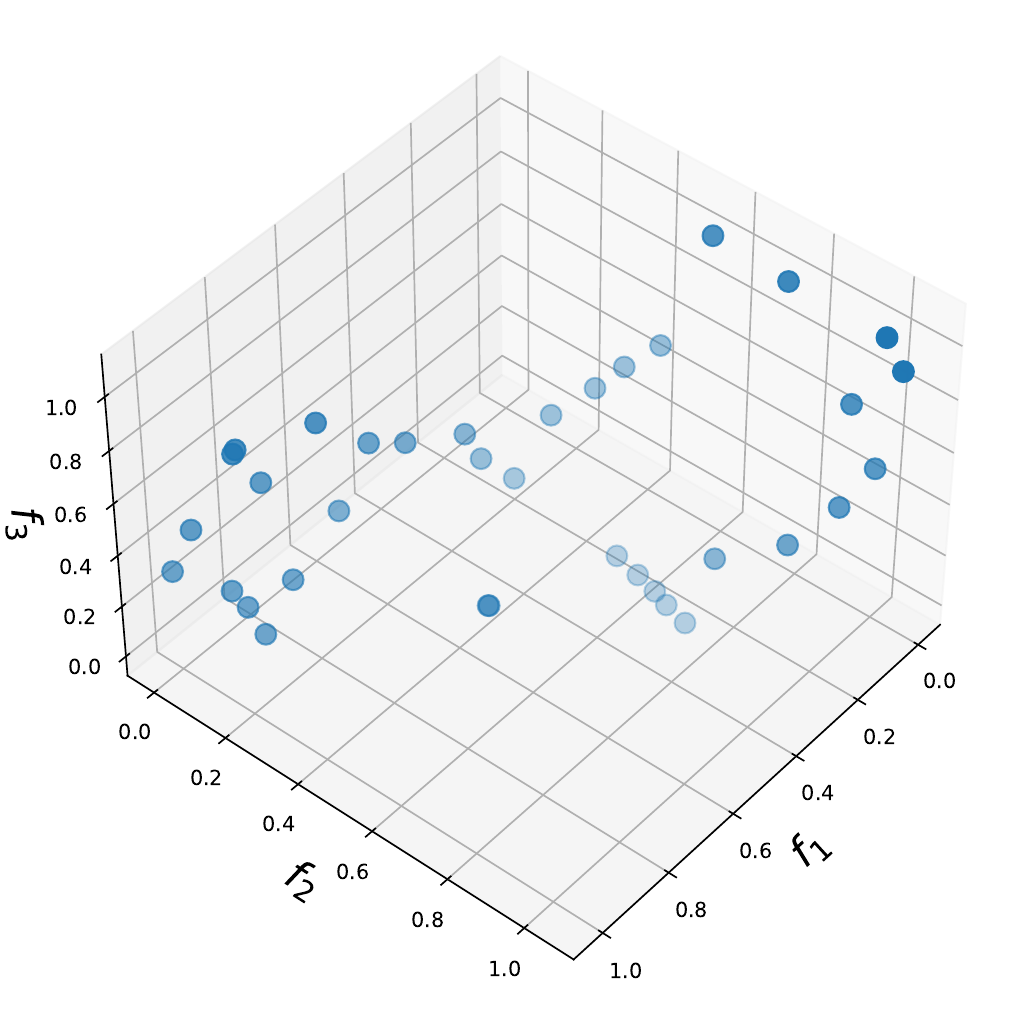}}
    \hfill
    \subfloat[SMS]{\includegraphics[width = \hwidth \linewidth]{Figure/main/RE37/sms/final.pdf}}
    \hfill
    \subfloat[UMOEA/D]{\includegraphics[width = \hwidth \linewidth]{Figure/main/RE37/adjust/final.pdf}}
    \hfill
    \caption{Results on RE37. } 
    \label{fig:3obj_re37}
\end{figure*}

\begin{figure*}[h!]
    \subfloat[AWA-1]{\includegraphics[width = \hwidth \linewidth]{Figure/main/RE41/awa/proj_1.pdf}}
    \hfill
    \subfloat[GP-1]{\includegraphics[width = \hwidth \linewidth]{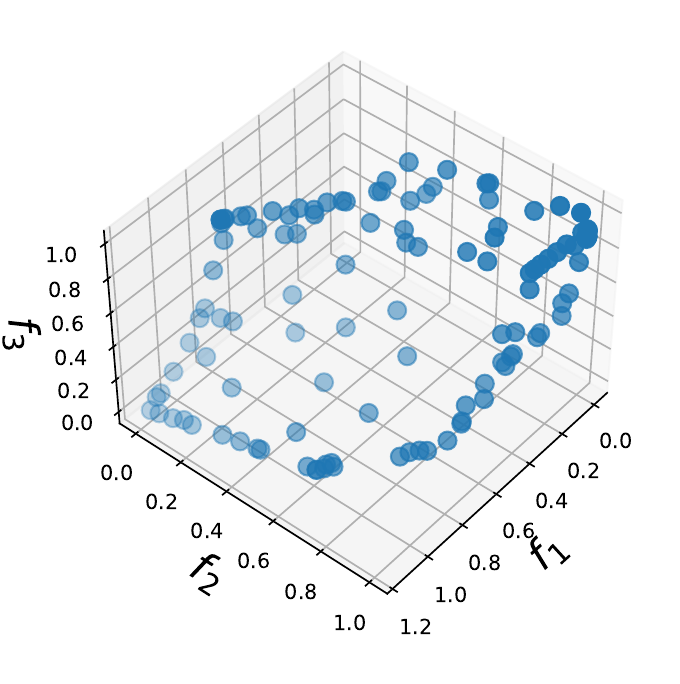}}
    \hfill
    \subfloat[MOEA/D-1]{\includegraphics[width = \hwidth \linewidth]{Figure/main/RE41/moead/proj_1.pdf}}
    \hfill
    \subfloat[PaLam-1]{\includegraphics[width = \hwidth \linewidth]{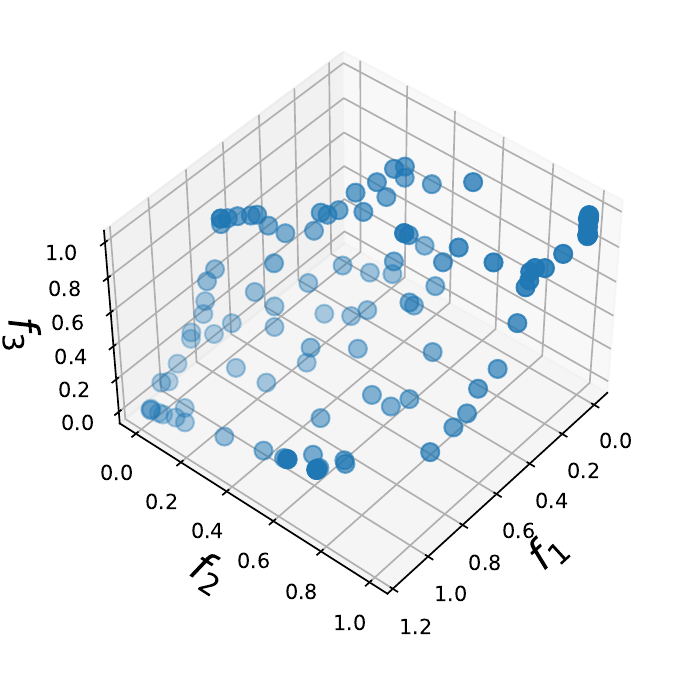}}
    \hfill
    \subfloat[SMS-1]{\includegraphics[width = \hwidth \linewidth]{Figure/main/RE41/sms/proj_1.pdf}}
    \hfill
    \subfloat[UMOEA/D-1]{\includegraphics[width = \hwidth \linewidth]{Figure/main/RE41/adjust/proj_1.pdf}} \\
    \subfloat[AWA-2]{\includegraphics[width = \hwidth \linewidth]{Figure/main/RE41/awa/proj_2.pdf}}
    \hfill
    \subfloat[GP-2]{\includegraphics[width = \hwidth \linewidth]{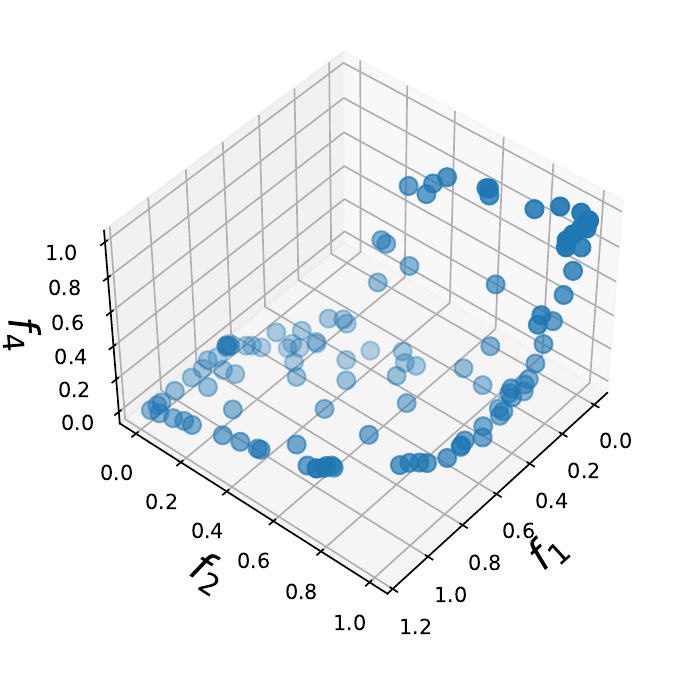}}
    \hfill
    \subfloat[MOEA/D-2]{\includegraphics[width = \hwidth \linewidth]{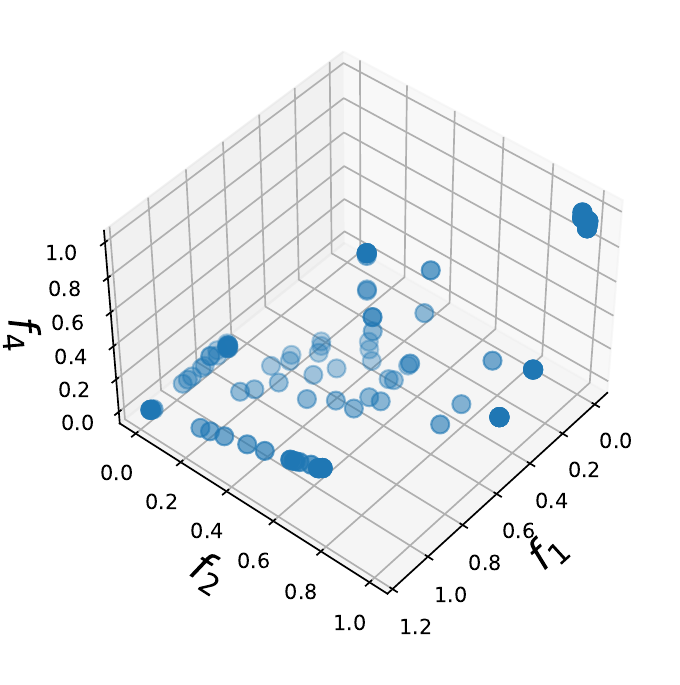}}
    \hfill
    \subfloat[PaLam-2]{\includegraphics[width = \hwidth \linewidth]{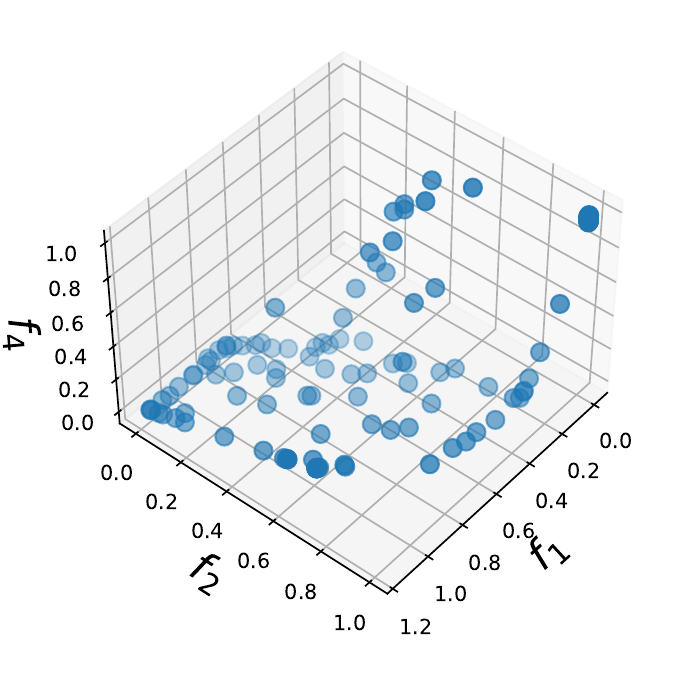}}
    \hfill
    \subfloat[SMS-2]{\includegraphics[width = \hwidth \linewidth]{Figure/main/RE41/sms/proj_2.pdf}}
    \hfill
    \subfloat[UMOEA/D-2]{\includegraphics[width = \hwidth \linewidth]{Figure/main/RE41/adjust/proj_2.pdf}} \\
    \subfloat[AWA-3]{\includegraphics[width = \hwidth \linewidth]{Figure/main/RE41/awa/proj_3.pdf}}
    \hfill
    \subfloat[GP-3]{\includegraphics[width = \hwidth \linewidth]{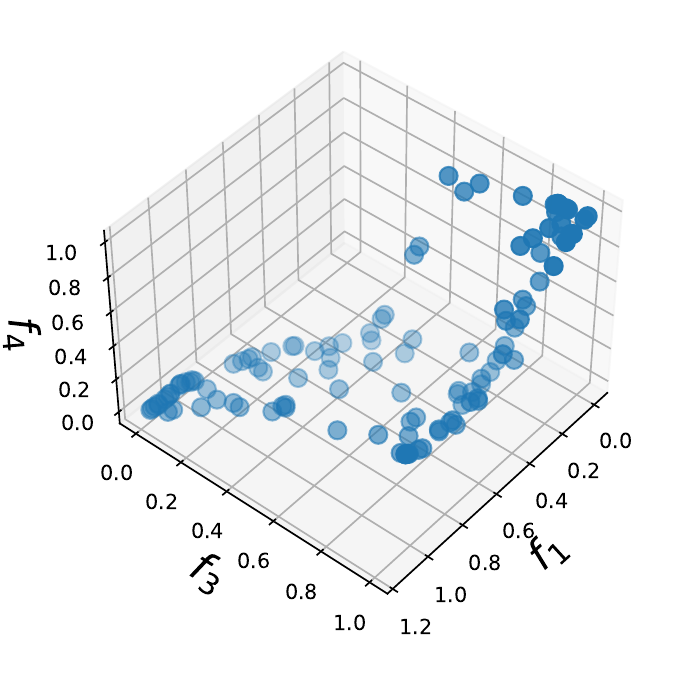}}
    \hfill
    \subfloat[MOEA/D-3]{\includegraphics[width = \hwidth \linewidth]{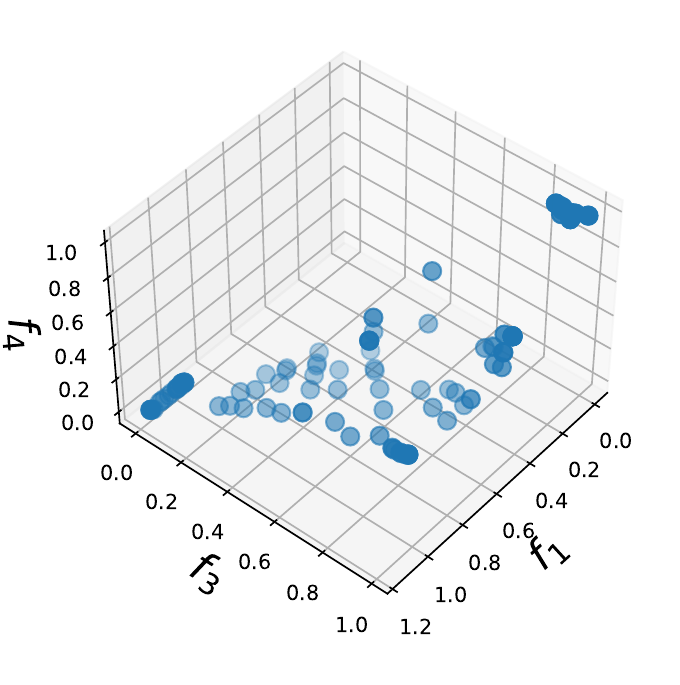}}
    \hfill
    \subfloat[PaLam-3]{\includegraphics[width = \hwidth \linewidth]{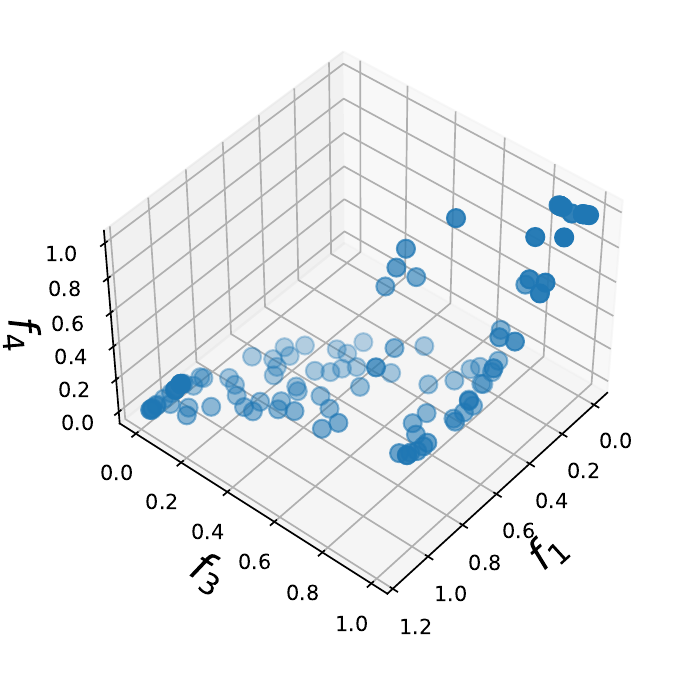}}
    \hfill
    \subfloat[SMS-3]{\includegraphics[width = \hwidth \linewidth]{Figure/main/RE41/sms/proj_3.pdf}}
    \hfill
    \subfloat[UMOEA/D-3]{\includegraphics[width = \hwidth \linewidth]{Figure/main/RE41/adjust/proj_3.pdf}} \\
    \subfloat[AWA-4]{\includegraphics[width = \hwidth \linewidth]{Figure/main/RE41/awa/proj_4.pdf}}
    \hfill
    \subfloat[GP-4]{\includegraphics[width = \hwidth \linewidth]{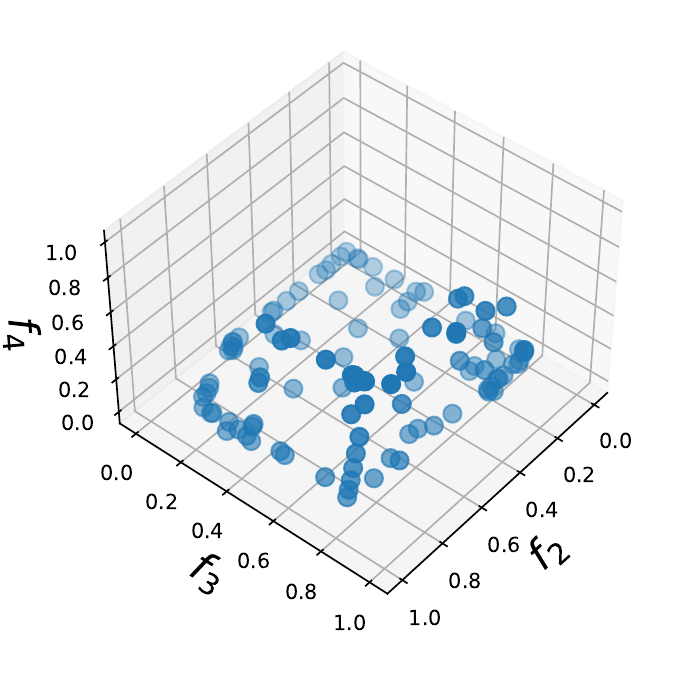}}
    \hfill
    \subfloat[MOEA/D-4]{\includegraphics[width = \hwidth \linewidth]{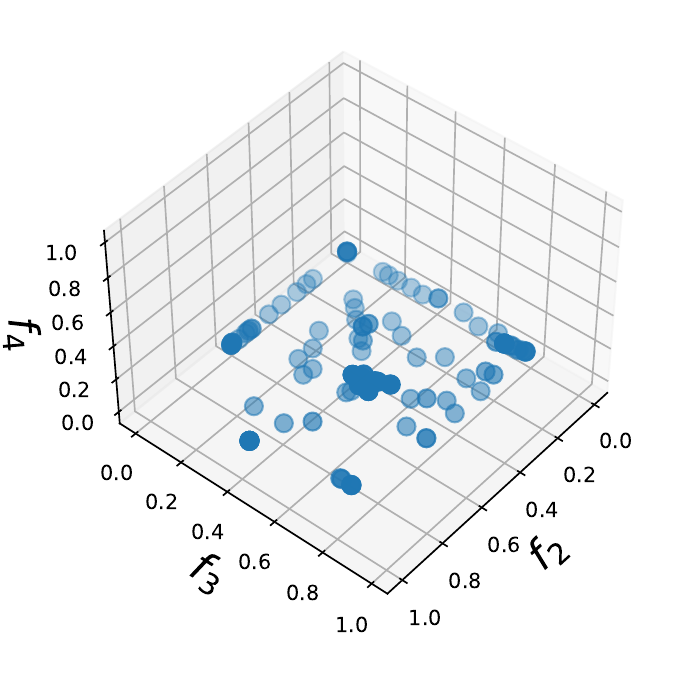}}
    \hfill
    \subfloat[PaLam-4]{\includegraphics[width = \hwidth \linewidth]{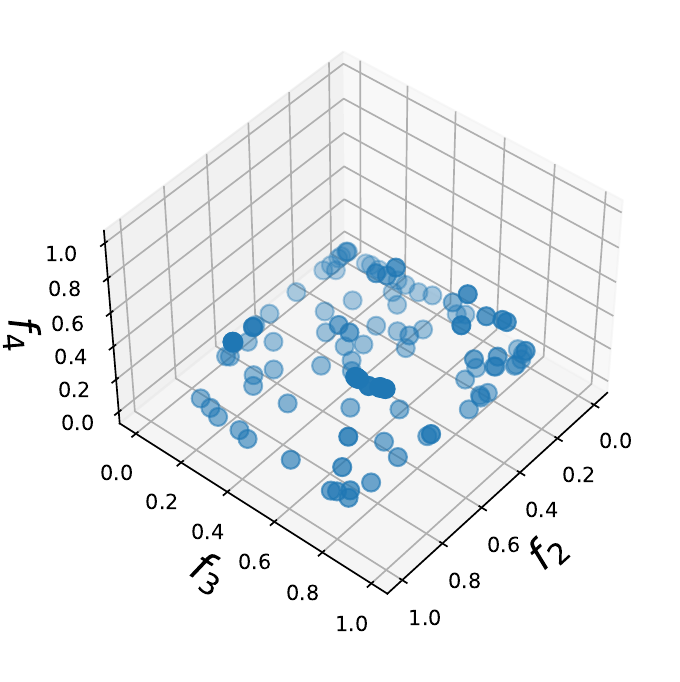}}
    \hfill
    \subfloat[SMS-4]{\includegraphics[width = \hwidth \linewidth]{Figure/main/RE41/sms/proj_4.pdf}}
    \hfill
    \subfloat[UMOEA/D-4]{\includegraphics[width = \hwidth \linewidth]{Figure/main/RE41/adjust/proj_4.pdf}}
    \hfill
    \caption{Results on RE41.}
    \label{fig:4obj_re41}
\end{figure*}

\begin{figure*}[h!]
    \subfloat[AWA-1]{\includegraphics[width = \hwidth \linewidth]{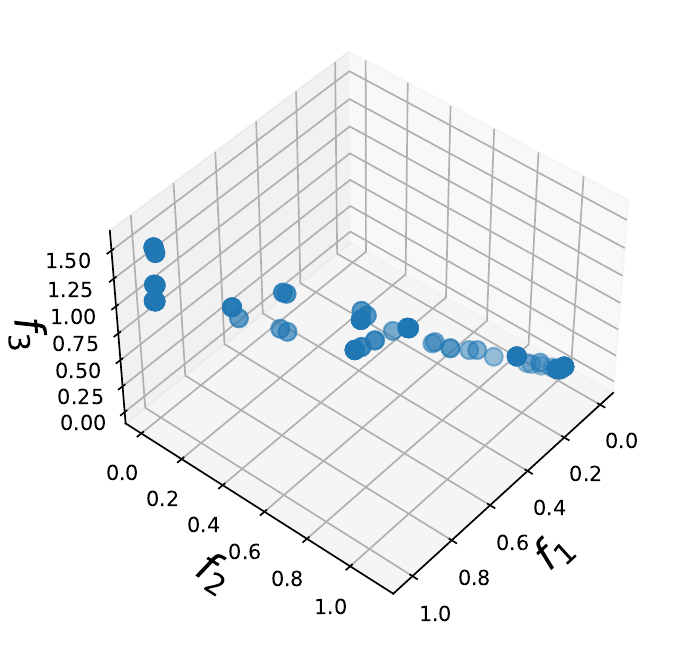}}
    \hfill
    \subfloat[GP-1]{\includegraphics[width = \hwidth \linewidth]{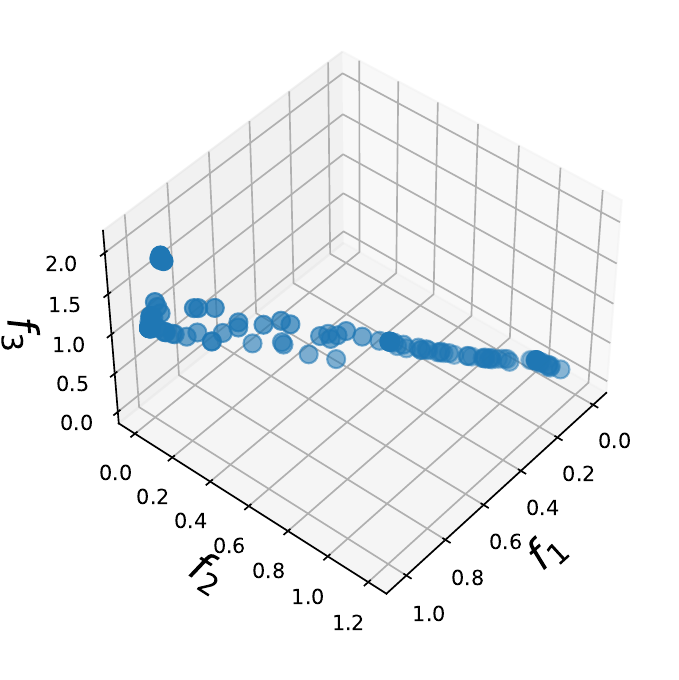}}
    \hfill
    \subfloat[MOEA/D-1]{\includegraphics[width = \hwidth \linewidth]{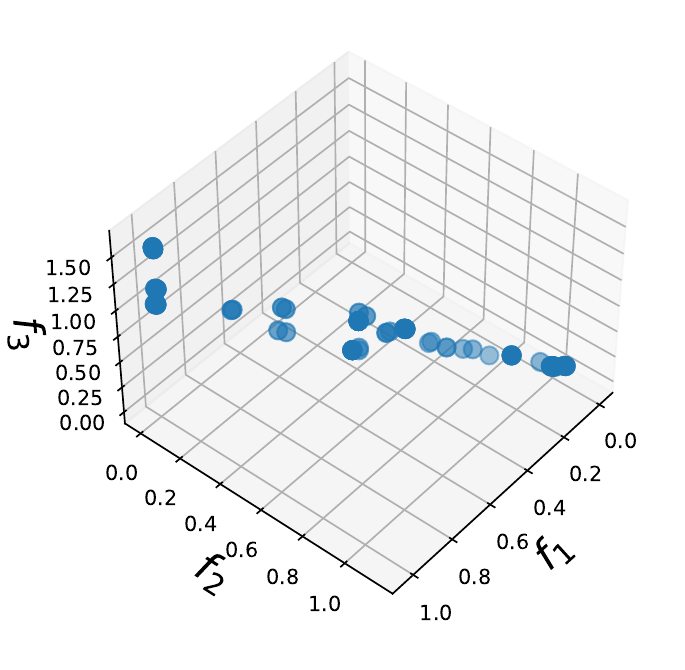}}
    \hfill
    \subfloat[PaLam-1]{\includegraphics[width = \hwidth \linewidth]{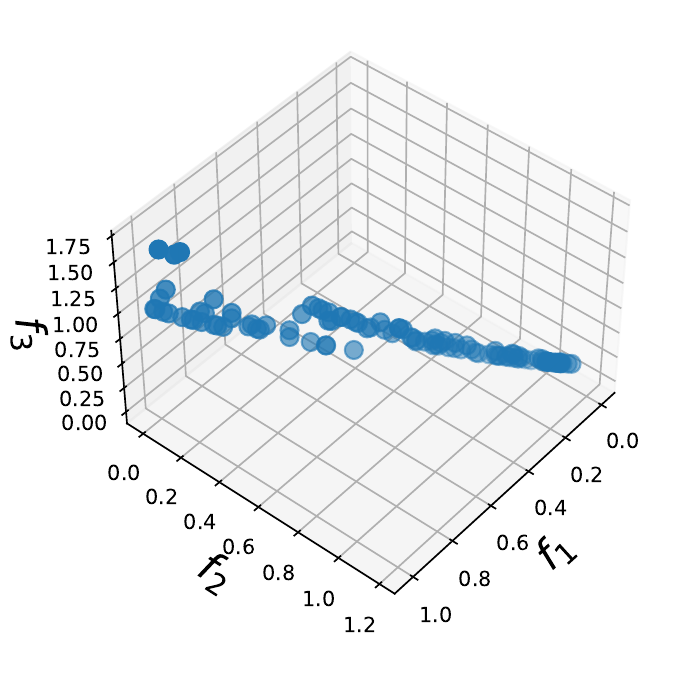}}
    \hfill
    \subfloat[SMS-1]{\includegraphics[width = \hwidth \linewidth]{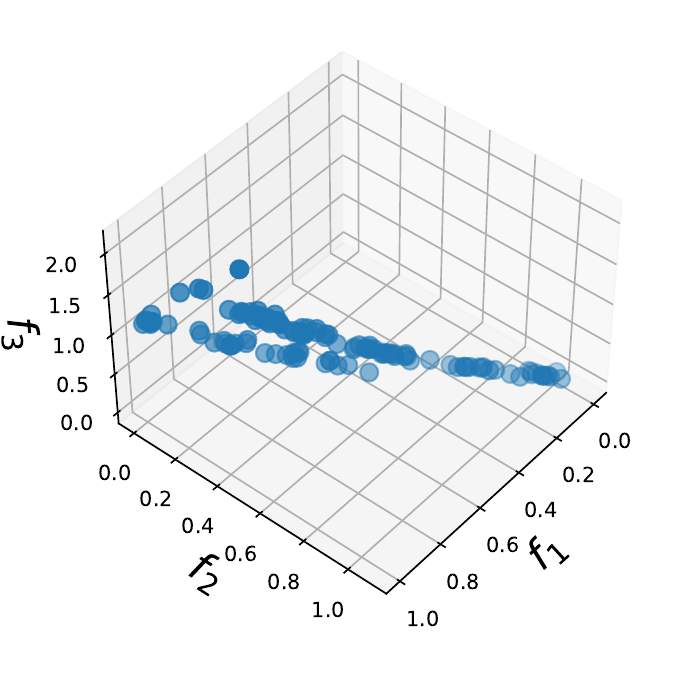}}
    \hfill
    \subfloat[UMOEA/D-1]{\includegraphics[width = \hwidth \linewidth]{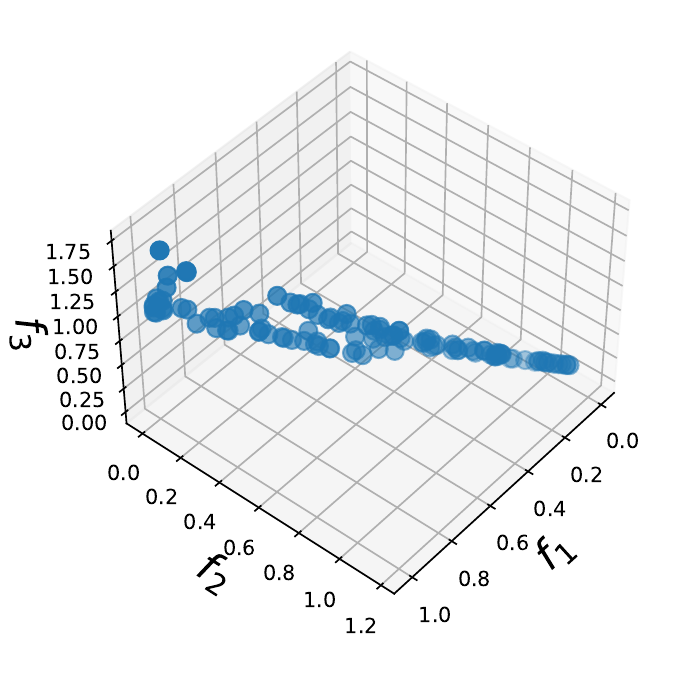}} \\
    \subfloat[AWA-2]{\includegraphics[width = \hwidth \linewidth]{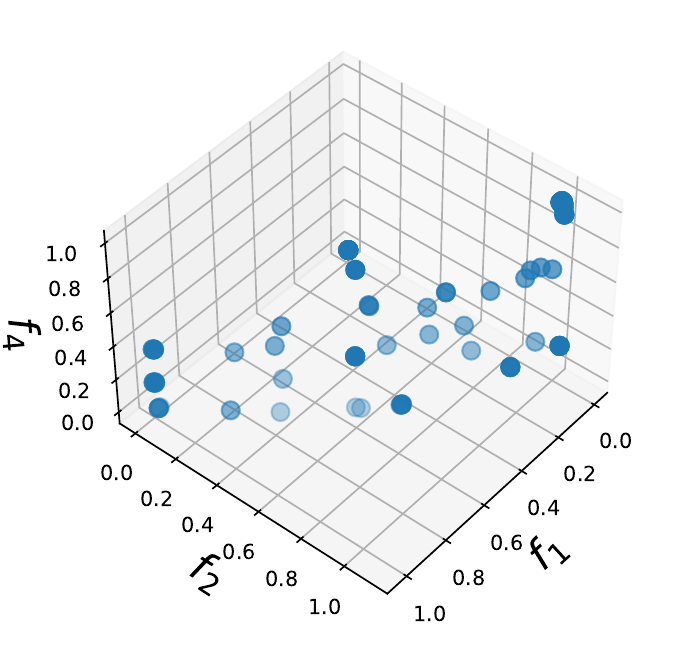}}
    \hfill
    \subfloat[GP-2]{\includegraphics[width = \hwidth \linewidth]{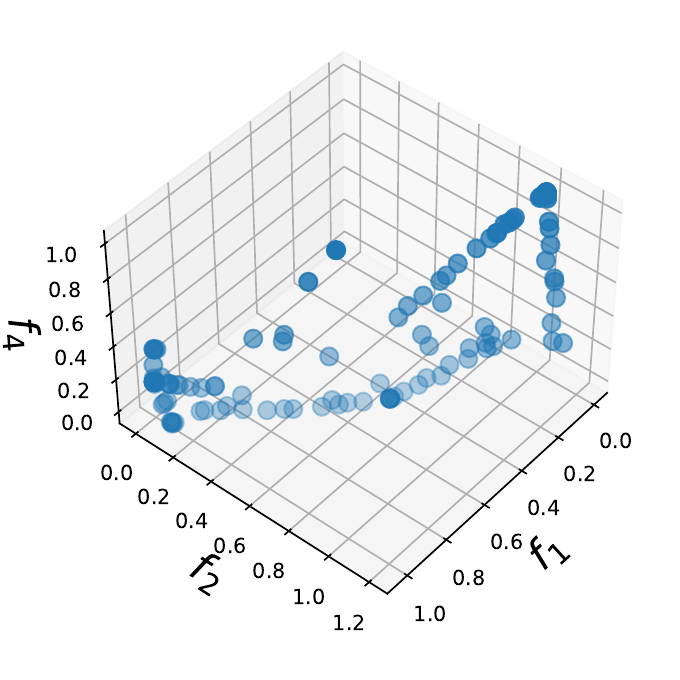}}
    \hfill
    \subfloat[MOEA/D-2]{\includegraphics[width = \hwidth \linewidth]{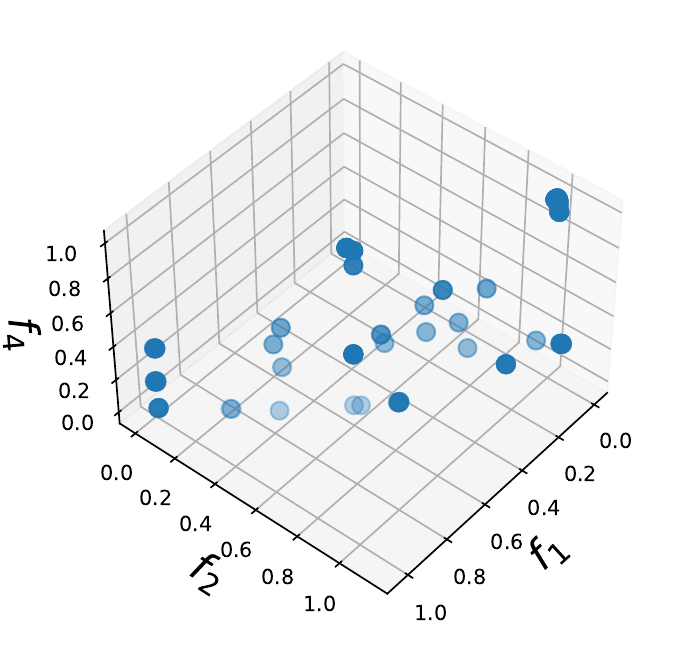}}
    \hfill
    \subfloat[PaLam-2]{\includegraphics[width = \hwidth \linewidth]{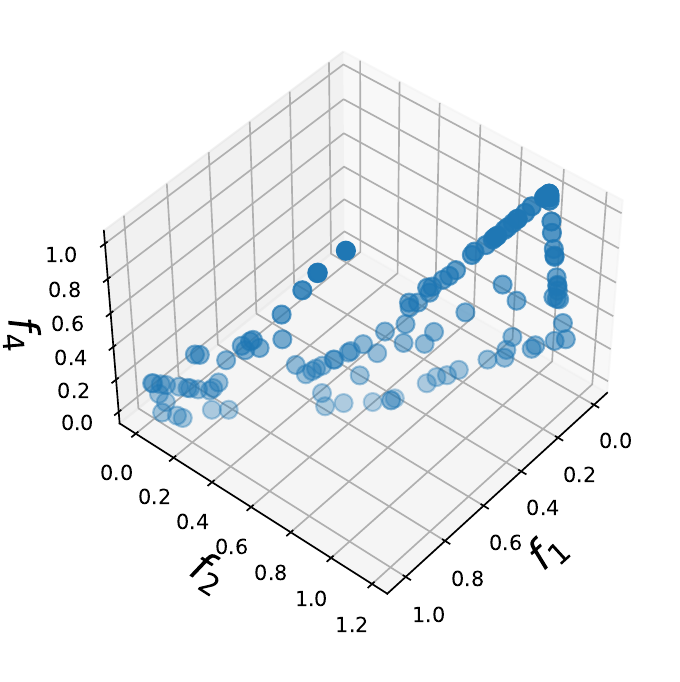}}
    \hfill
    \subfloat[SMS-2]{\includegraphics[width = \hwidth \linewidth]{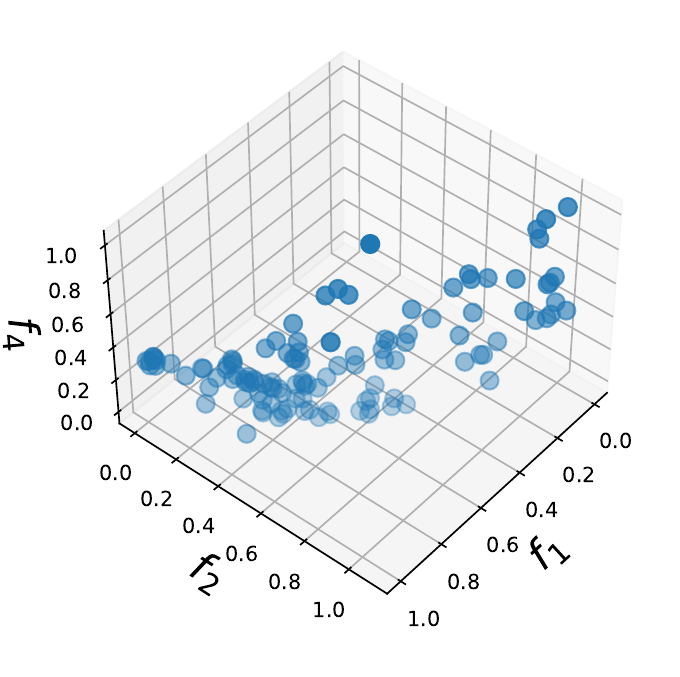}}
    \hfill
    \subfloat[UMOEA/D-2]{\includegraphics[width = \hwidth \linewidth]{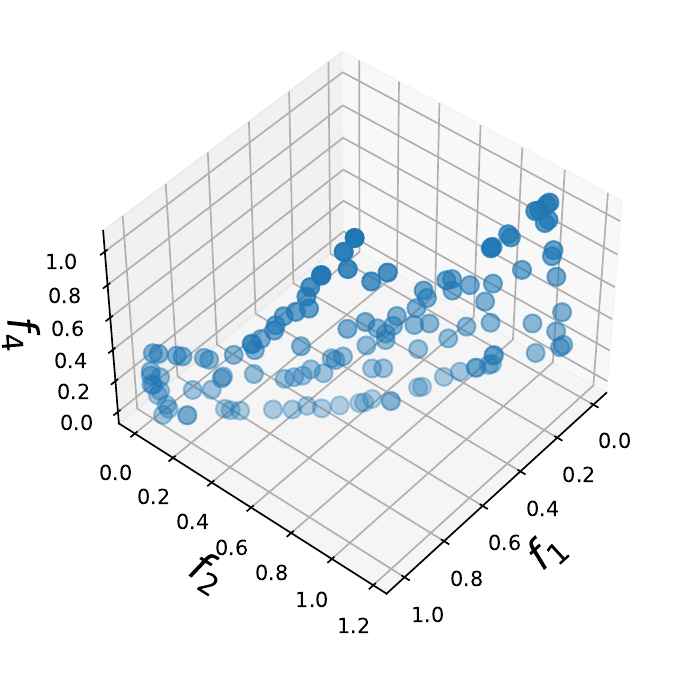}} \\
    \subfloat[AWA-3]{\includegraphics[width = \hwidth \linewidth]{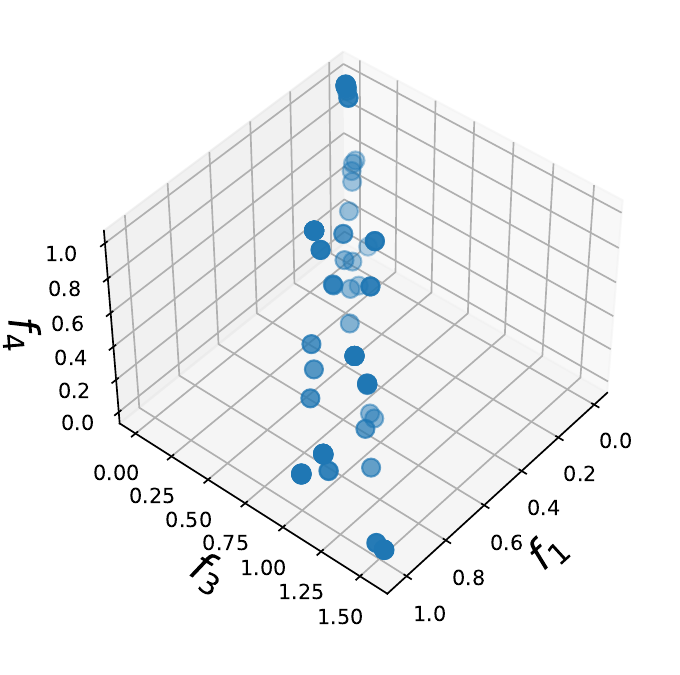}}
    \hfill
    \subfloat[GP-3]{\includegraphics[width = \hwidth \linewidth]{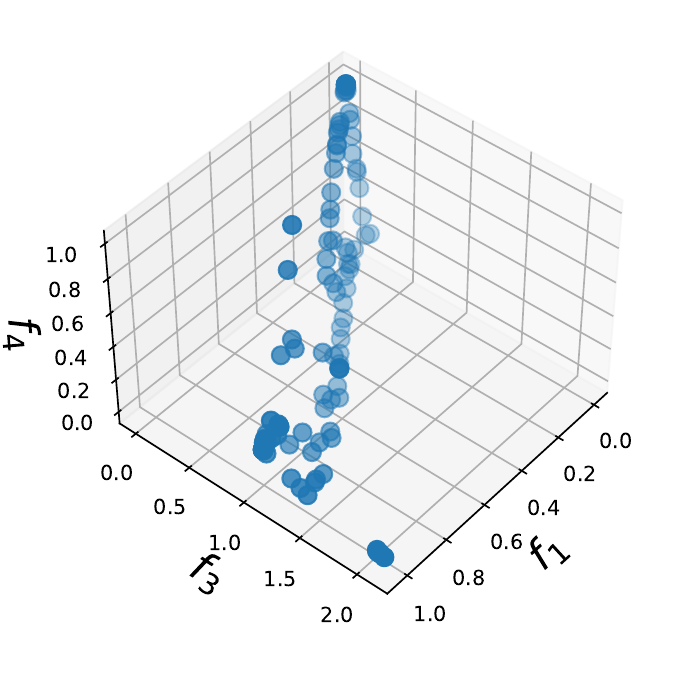}}
    \hfill
    \subfloat[MOEA/D-3]{\includegraphics[width = \hwidth \linewidth]{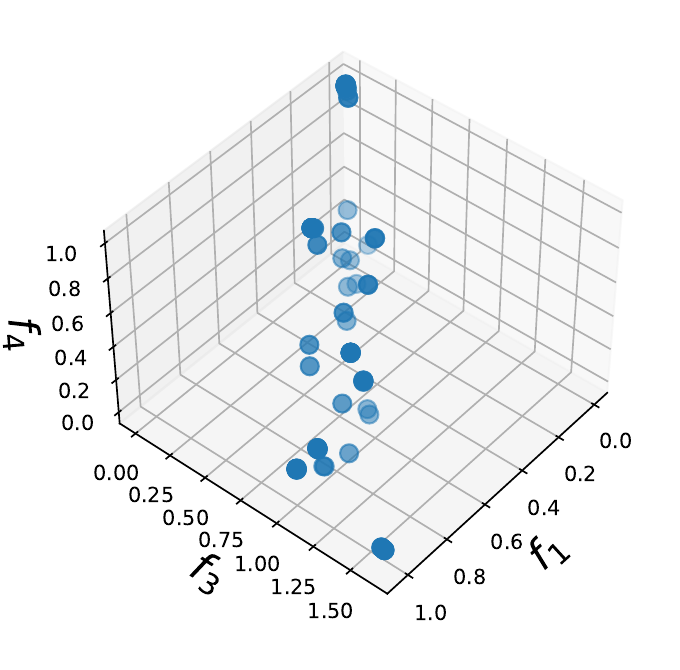}}
    \hfill
    \subfloat[PaLam-3]{\includegraphics[width = \hwidth \linewidth]{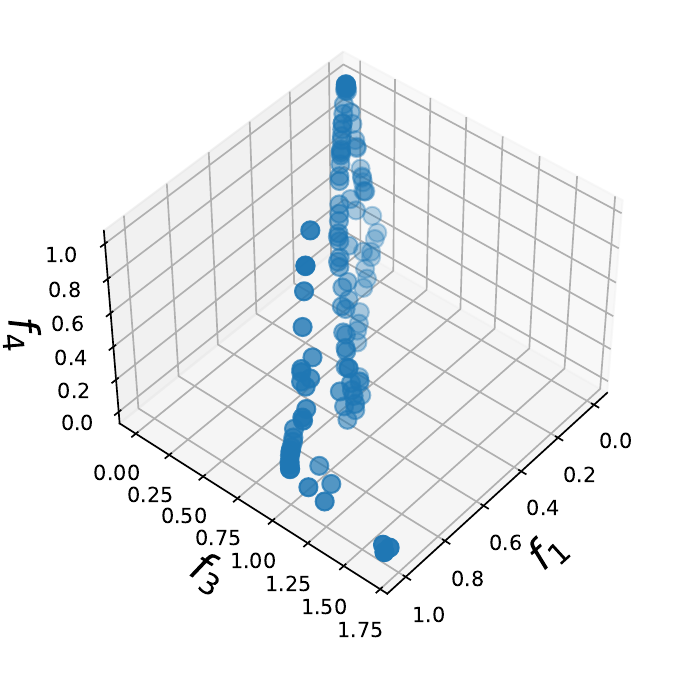}}
    \hfill
    \subfloat[SMS-3]{\includegraphics[width = \hwidth \linewidth]{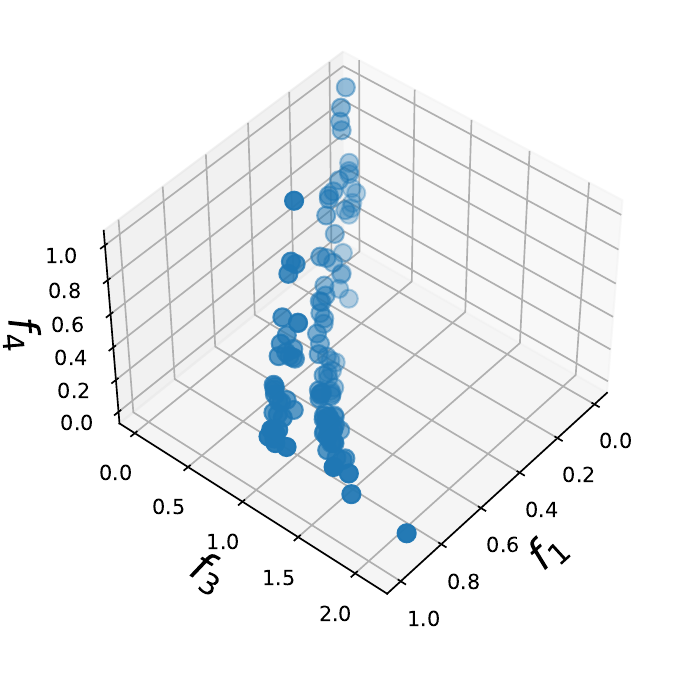}}
    \hfill
    \subfloat[UMOEA/D-3]{\includegraphics[width = \hwidth \linewidth]{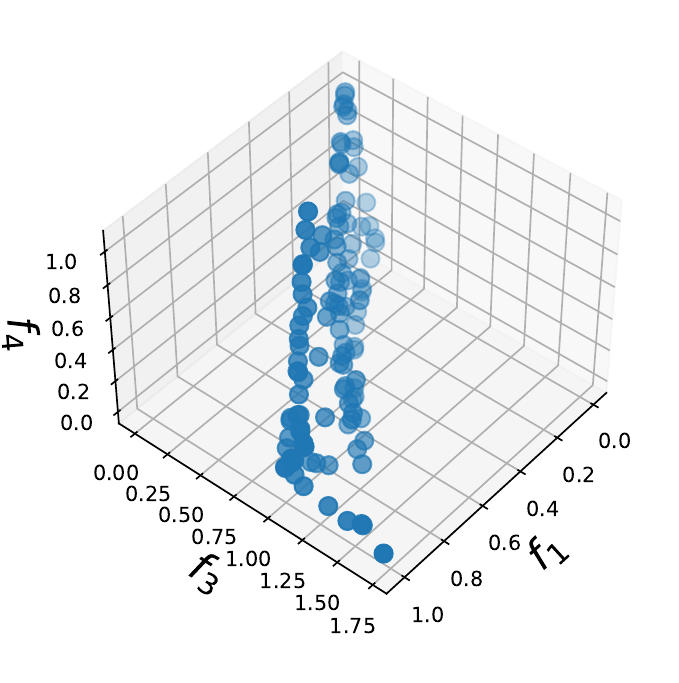}} \\
    \subfloat[AWA-4]{\includegraphics[width = \hwidth \linewidth]{Figure/main/RE42/awa/proj_4.pdf}}
    \hfill
    \subfloat[GP-4]{\includegraphics[width = \hwidth \linewidth]{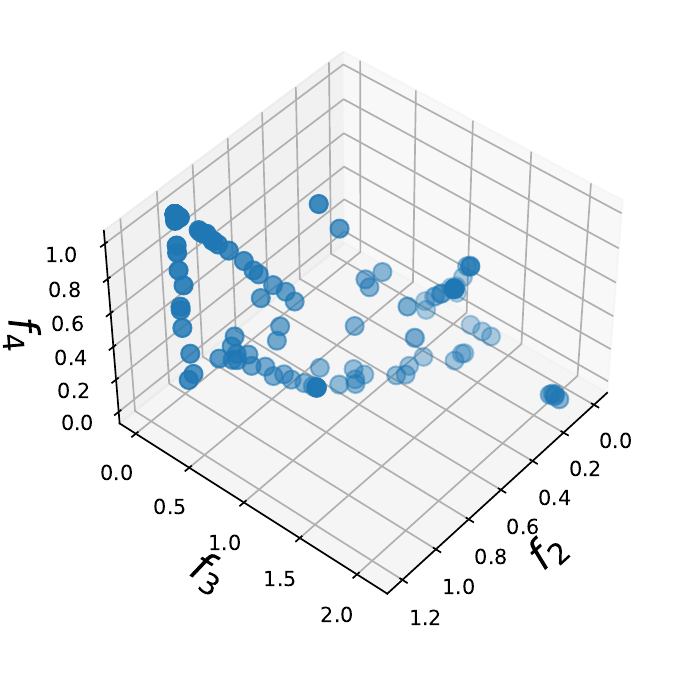}}
    \hfill
    \subfloat[MOEA/D-4]{\includegraphics[width = \hwidth \linewidth]{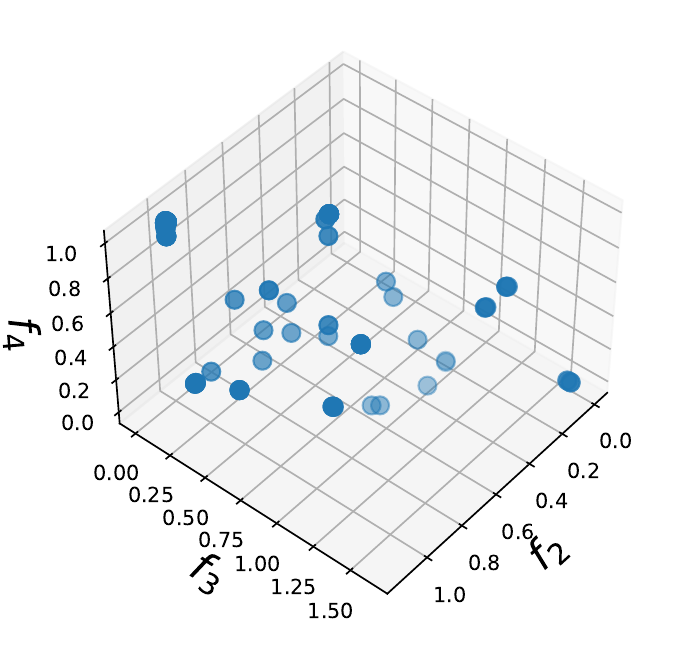}}
    \hfill
    \subfloat[PaLam-4]{\includegraphics[width = \hwidth \linewidth]{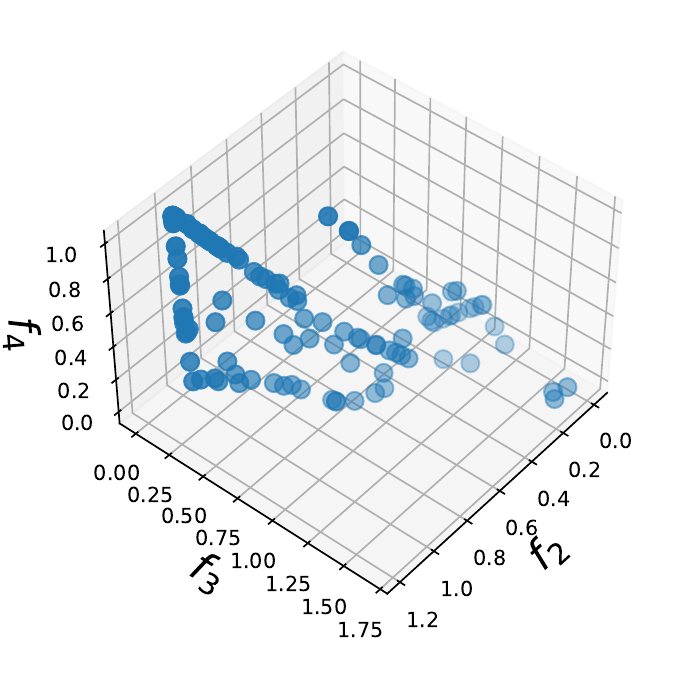}}
    \hfill
    \subfloat[SMS-4]{\includegraphics[width = \hwidth \linewidth]{Figure/main/RE42/sms/proj_4.pdf}}
    \hfill
    \subfloat[UMOEA/D-4]{\includegraphics[width = \hwidth \linewidth]{Figure/main/RE42/adjust/proj_4.pdf}}
    \hfill
    \caption{Results on RE42.}
    \label{fig:4obj_re42}
\end{figure*}

In this subsection, we visualize the results on representative problems through Figure \ref{fig:2obj_zdt1} to \ref{fig:4obj_re42}. Those figures depict the outcomes of MOEA/D, SMS-EMOA, MOEA/D-GP, MOEA/D-L, PaLam, MOEA/D-AWA, and the proposed UMOEA/D. Numerical results are shown in \Cref{tab:obj2} and \Cref{tab:obj3}. (Soft) Optimal distances between minimal Pareto objectives are marked in bold. Except for RE42, our method achieve the optimal pairwise distances among all problems. For RE42, SMS-EMOA has the optimal distance value, however SMS-EMOA has not converged on this paper. 

As we have mentioned in the main paper, the hypervolume indicator is only a coarse indicator to measure the uniformity level. The introduction of it is mainly to determine whether the algorithm has converged or not. So we do not mark the highest hypervolume indicator in bold in Appendix. 

We summarize the key experimental findings as follows. 

\textbf{(Neighbour distances are equal in two-objective problems).} For the standard ZDT1 and ZDT2 problems, which have a PF ranging from zero to one, the Pareto objectives optimized by MOEA/D are not uniformly distributed. However, our method ensures that the distance between adjacent solutions is equal, indicating a more uniform distribution.
For MOEA/D, when considering a convex-shaped PF like ZDT1, solutions tend to be denser in the middle. Conversely, for a concave-shaped PF like ZDT2, solutions are denser towards the margins. However, the proposed method's Pareto objective distribution remains unaffected by the shape of the true PF.

\paragraph{(The proposed method is robust to objectives with different scales).} When function ranges differ in scale (see \Cref{fig:2obj_re21}), the uniformity of pure MOEA/D worsens. In this scenario, achieving weight uniformity does not guarantee objective uniformity. Solutions become even sparser in the upper-left region of the objective space. In contrast, the proposed method consistently generates uniform Pareto objectives. Hypervolume-based methods remain unaffected when function objectives have different scales. However, objectives produced by hypervolume-based methods are not strictly uniform, and hypervolume-based methods are typically slower.

\paragraph{(HV is not a accurate uniformity indicator).}
The hypervolume indicator only provides an approximate measure of solution diversity. \Tableref{app:tab:all} illustrates that similar hypervolume indicators can correspond to significantly different solution distributions. In the case of hypervolume-based methods (PaLam or SMS-EMOA), the largest hypervolume indicator does not necessarily lead to the most uniform objectives for most problems.   

\paragraph{(Weight uniformity induces solution uniformity in DTLZ1).} As mentioned in examples in the main paper, the PF of DTLZ1 can be represented as an affine transformation of the 2-simplex $\vDelta_2$, since $\vh(\vx) = \frac12 \vx$. For problems considered in this paper, DTLZ1 is the only scenario where uniform weights result in uniformly distributed Pareto objectives. Hence, MOEA/D with uniform weights performs exceptionally well on the DTLZ1 task. The maximal manifold separation indicator, $\delta$, outperforms all other methods, and the proposed method achieves a value of $\delta = 0.099$, which is very close to that of MOEA/D and significantly outperforms other methods. 

\paragraph{(Results on the difficult RE37/RE41/RE42 problem).} Finally, we present the results for the challenging three-objective real-world problem RE37/RE41/RE42. One of the difficulties of this problem is the presence of many weights within the weight simplex $\vDelta_2$ that do not intersect with the PF. The pure MOEA/D algorithm produces numerous duplicate solutions when weights do not intersect with the PF, resulting in wasted resources and poor solution diversity. The hypervolume-based method SMS-EMOA exhibits an interesting phenomenon: it mainly focuses on the marginal region of the PF, which may not always meet user demands. In contrast, the proposed method is the only method that generates uniform objectives that cover the entire PF.

For three-objective problems, since DTLZ 2-4 share the same PF. We take DTLZ2 as a representative, as shown in \Cref{fig:3obj_dtlz2}. We also show the projection on the difficult four-objective RE41 and RE42 problem as shown in \Cref{fig:4obj_re41} and \Cref{fig:4obj_re42}.

\clearpage
\begin{table}[htbp]
  \centering
  \footnotesize
  \caption{Results on two-objective problems averaged on five random seeds.} \label{tab:obj2} %
    \begin{tabular}{lrrrrr}
    \toprule
     Method & \multicolumn{1}{r}{\texttt{HV}} & \multicolumn{1}{r}{\texttt{Spacing}} & \multicolumn{1}{r}{\texttt{Sparsity}} & \multicolumn{1}{r}{$\delta$} & \multicolumn{1}{r}{$\tilde{\delta}_\Tau$} \\
    \midrule
    \multicolumn{6}{c}{ZDT1} \\
    \midrule
    SMS-EMOA & 1.2620 & 0.0143 & 0.0064 & 0.0489 & -0.1101 \\
    MOEA/D-AWA & 1.2609 & 0.0308 & 0.0073 & 0.0000 & -0.1195 \\
    MOEA/D-GP & 1.2626 & 0.0175 & 0.0061 & 0.0331 & -0.1059 \\
    MOEAD-L & 1.2628 & 0.0090 & 0.0060 & 0.0479 & -0.1052 \\
    MOEA/D & 1.2589 & 0.0553 & 0.0090 & 0.0432 & -0.1304 \\
    PaLam & 1.2634 & 0.0122 & 0.0061 & 0.0621 & -0.1070 \\
    UMOEA/D & 1.2634 & \tb{0.0026} & \tb{0.0059} & \tb{0.0720} & \tb{-0.1046} \\
    \midrule
    \multicolumn{6}{c}{ZDT2} \\
    \midrule
    SMS-EMOA & 0.4077 & 0.0153 & \tb{0.0020} & 0.0144 & -0.1780 \\
    MOEA/D-AWA & 0.7418 & 0.0155 & 0.0099 & 0.0744 & -0.0813 \\
    MOEA/D-GP & 0.7421 & 0.0124 & 0.0098 & 0.0741 & -0.0799 \\
    MOEAD-L & 0.7423 & 0.0086 & 0.0100 & 0.0797 & -0.0810 \\
    MOEA/D & 0.7418 & 0.0160 & 0.0099 & 0.0702 & -0.0813 \\
    PaLam & 0.7435 & 0.0256 & 0.0106 & 0.0388 & -0.0867 \\
    UMOEA/D & 0.7429 & \tb{0.0082} & 0.0099 & \tb{0.0814} & \tb{-0.0800} \\
    \midrule
    \multicolumn{6}{c}{ZDT4} \\
    \midrule
    SMS-EMOA & 0.0785 & 1.1220 & 8.2860 & 0.0212 & -0.0883 \\
    MOEA/D-AWA & 1.0708 & 0.0259 & 0.0109 & 0.0580 & -0.0832 \\
    MOEA/D-GP & 1.0731 & 0.0142 & 0.0097 & 0.0646 & -0.0805 \\
    MOEAD-L & 1.0731 & 0.0042 & 0.0097 & 0.0865 & -0.0797 \\
    MOEA/D & 1.0718 & 0.0349 & 0.0105 & 0.0704 & -0.0927 \\
    PaLam & 1.0748 & 0.0120 & 0.0098 & 0.0872 & -0.0811 \\
    UMOEA/D & 1.0738 & \tb{0.0036} & \tb{0.0096} & \tb{0.0879} & \tb{-0.0794} \\
    \midrule
    \multicolumn{6}{c}{ZDT6} \\
    \midrule
    SMS-EMOA & 0.5808 & 0.0382 & 0.0077 & 0.0204 & -0.1296 \\
    MOEA/D-AWA & 0.6867 & 0.0152 & 0.0064 & 0.0425 & -0.1040 \\
    MOEA/D-GP & 0.6873 & 0.0093 & 0.0063 & 0.0623 & -0.1025 \\
    MOEAD-L & 0.6876 & 0.0019 & 0.0063 & 0.0723 & -0.1021 \\
    MOEA/D & 0.6867 & 0.0125 & 0.0064 & 0.0548 & -0.1040 \\
    PaLam & 0.6878 & 0.0072 & 0.0063 & 0.0605 & -0.1021 \\
    UMOEA/D & 0.6877 & \tb{0.0019} & \tb{0.0062} & \tb{0.0730} & \tb{-0.1018} \\
    \midrule
    \multicolumn{6}{c}{RE21} \\
    \midrule
    SMS-EMOA & 1.2621 & 0.0179 & 0.0062 & 0.0519 & -0.1100 \\
    MOEA/D-AWA & 1.2616 & 0.0197 & 0.0065 & 0.0286 & -0.1092 \\
    MOEA/D-GP & 1.2627 & 0.0165 & 0.0061 & 0.0369 & -0.1058 \\
    MOEAD-L & 1.2631 & 0.0049 & 0.0060 & \tb{0.0656} & -0.1053 \\
    MOEA/D & 1.2591 & 0.0553 & 0.0090 & 0.0432 & -0.1304 \\
    PaLam & 1.2637 & 0.0121 & 0.0061 & 0.0627 & -0.1070 \\
    UMOEA/D & 1.2632 & \tb{0.0044} & \tb{0.0059} & \tb{0.0656} & \tb{-0.1048} \\
    \midrule
    \multicolumn{6}{c}{RE22} \\
    \midrule
    SMS-EMOA & 1.2007 & 0.0167 & 0.0065 & 0.0418 & -0.1062 \\
    MOEA/D-AWA & 1.2005 & 0.0152 & 0.0063 & 0.0485 & -0.1084 \\
    MOEA/D-GP & 1.2006 & 0.0235 & 0.0065 & 0.0227 & -0.1092 \\
    MOEAD-L & 1.2004 & 0.0071 & 0.0062 & \tb{0.0633} & \tb{-0.1041} \\
    MOEA/D & 1.1984 & 0.0317 & 0.0074 & 0.0329 & -0.1242 \\
    PaLam & 1.2013 & 0.0128 & 0.0063 & 0.0475 & -0.1049 \\
    UMOEA/D & 1.2008 & \tb{0.0044} & \tb{0.0061} & \tb{0.0663} & \tb{-0.1041} \\
    \bottomrule
    \end{tabular}   
\end{table}

\begin{table}[htbp]
  \centering
  \footnotesize
  \caption{Results on three-objective and four-objective problems averaged on five random seeds.} \label{tab:obj3} %
    \begin{tabular}{lrrrrr}
    \toprule
     Method & \multicolumn{1}{r}{\texttt{HV}} & \multicolumn{1}{r}{\texttt{Spacing}} & \multicolumn{1}{r}{\texttt{Sparsity}} & \multicolumn{1}{r}{$\delta$} & \multicolumn{1}{r}{$\hat{\delta}$} \\
    \midrule
    \multicolumn{6}{c}{DTLZ1} \\
    \midrule
    SMS-EMOA & 1.6971 & 0.0098 & 0.0014 & 0.0619 & -0.1723 \\
    MOEA/D-AWA & 1.6955 & 0.0054 & 0.0030 & 0.0775 & -0.1682 \\
    MOEA/D-GP & 1.6963 & 0.0193 & \tb{0.0012} & 0.0410 & -0.1715 \\
    MOEA/D & 1.6957 & \tb{0.0002} & 0.0030 & \tb{0.1002} & -0.1696 \\
    PaLam & 1.6973 & 0.0122 & \tb{0.0012} & 0.0516 & -0.1712 \\
    UMOEA/D & 1.6967 & 0.0007 & 0.0030 & 0.0988 & \tb{-0.1694} \\  
    \midrule
    \multicolumn{6}{c}{DTLZ2} \\
    \midrule
    SMS-EMOA & 1.1166 & 0.0566 & 0.0103 & 0.0819 & -0.0781 \\
    MOEA/D-AWA & 1.1010 & 0.0475 & 0.0079 & 0.1155 & -0.0266 \\
    MOEA/D-GP & 1.0902 & 0.0432 & 0.0062 & 0.1517 & -0.0258 \\
    MOEA/D & 1.1000 & 0.0446 & 0.0080 & 0.1640 & -0.0194 \\
    PaLam & 1.1116 & 0.0555 & \tb{0.0067} & 0.0380 & -0.0652 \\
    UMOEA/D & 1.1041 & \tb{0.0236} & 0.0110 & \tb{0.2011} & \tb{-0.0069} \\
    \midrule
    \multicolumn{6}{c}{DTLZ3} \\
    \midrule
    SMS-EMOA & 1.1105 & 0.0524 & 0.0103 & 0.0952 & -0.0727 \\
    MOEA/D-AWA & 1.1012 & 0.0460 & 0.0076 & 0.1112 & -0.0267 \\
    MOEA/D-GP & 1.0850 & 0.0471 & \tb{0.0062} & 0.1093 & -0.0363 \\
    MOEA/D & 1.0980 & 0.0447 & 0.0080 & 0.1636 & -0.0191 \\
    PaLam & 1.1108 & 0.0534 & 0.0069 & 0.0443 & -0.0722 \\
    UMOEA/D & 1.1022 & \tb{0.0221} & 0.0111 & \tb{0.2171} & \tb{-0.0061} \\
    \midrule
    \multicolumn{6}{c}{DTLZ4} \\
    \midrule
    SMS-EMOA & 1.1168 & 0.0532 & 0.0102 & 0.1002 & -0.0779 \\
    MOEA/D-AWA & 1.1020 & 0.0421 & 0.0079 & 0.1392 & -0.0193 \\
    MOEA/D-GP & 1.0921 & 0.0420 & \tb{0.0054} & 0.1604 & -0.0236 \\
    MOEA/D & 1.0999 & 0.0446 & 0.0080 & 0.1640 & -0.0194 \\
    PaLam & 1.1155 & 0.0491 & 0.0076 & 0.1067 & -0.0616 \\
    UMOEA/D & 1.1037 & \tb{0.0225} & 0.0110 & \tb{0.2197} & \tb{-0.0066} \\
    \midrule
    \multicolumn{6}{c}{RE37} \\
    \midrule
    SMS-EMOA & 1.1143 & 0.0423 & 0.0052 & 0.0294 & -0.1270 \\
    MOEA/D-AWA & 1.0768 & 0.0797 & 0.0101 & 0.0012 & -0.2055 \\
    MOEA/D-GP & 1.0737 & 0.0715 & 0.0083 & 0.0023 & -0.1723 \\
    MOEA/D & 1.0519 & 0.0758 & 0.0122 & 0.0000 & -0.2055 \\
    PaLam & 1.1150 & 0.0712 & 0.0050 & 0.0023 & -0.1473 \\
    UMOEA/D & 1.1114 & \tb{0.0416} & \tb{0.0047} & \tb{0.0483} & \tb{-0.0805} \\
    \midrule
    \multicolumn{6}{c}{RE41} \\
    \midrule
    SMS-EMOA & 1.0976 & 0.0516 & 0.0014 & 0.0037 & -0.2458 \\
    MOEA/D-AWA & 1.1432 & 0.0676 & 0.0026 & 0.0000 & -0.3001 \\
    MOEA/D-GP & 1.1694 & 0.0687 & 0.0013 & 0.0001 & -0.2798 \\
    MOEA/D & 1.1328 & 0.0578 & 0.0034 & 0.0000 & -0.2998 \\
    PaLam & 1.2075 & 0.0551 & 0.0009 & 0.0019 & -0.2352 \\
    UMOEA/D & 1.2072 & \tb{0.0436} & \tb{0.0008} & \tb{0.0060} & \tb{-0.1876} \\
    \midrule
    \multicolumn{6}{c}{RE42} \\
    \midrule
    SMS-EMOA & 0.6080 & 0.0505 & 0.0018 & \tb{0.0009} & -0.2749 \\
    MOEA/D-AWA & 0.5927 & 0.0463 & 0.0047 & 0.0000 & -0.3455 \\
    MOEA/D-GP & 0.6639 & 0.0512 & 0.0028 & 0.0000 & -0.3058 \\
    MOEA/D & 0.5872 & 0.0466 & 0.0052 & 0.0000 & -0.3469 \\
    PaLam & 0.6920 & 0.0440 & \tb{0.0013} & 0.0002 & -0.2735 \\
    UMOEA/D & 0.6922 & \tb{0.0386} & 0.0019 & \underline{0.0007} & \tb{-0.2535} \\
    \bottomrule
    \end{tabular} 
\end{table} 

\clearpage
\subsection{Difference of the proposed UMOEA/D and MOEA/D-AWA}
\label{app:sec:moead_awa_illus}
\begin{figure*}[h!]
    \centering
    \subfloat[ZDT1. \label{fig_moea_epsl_a}]{\includegraphics[width = 0.22 \linewidth]{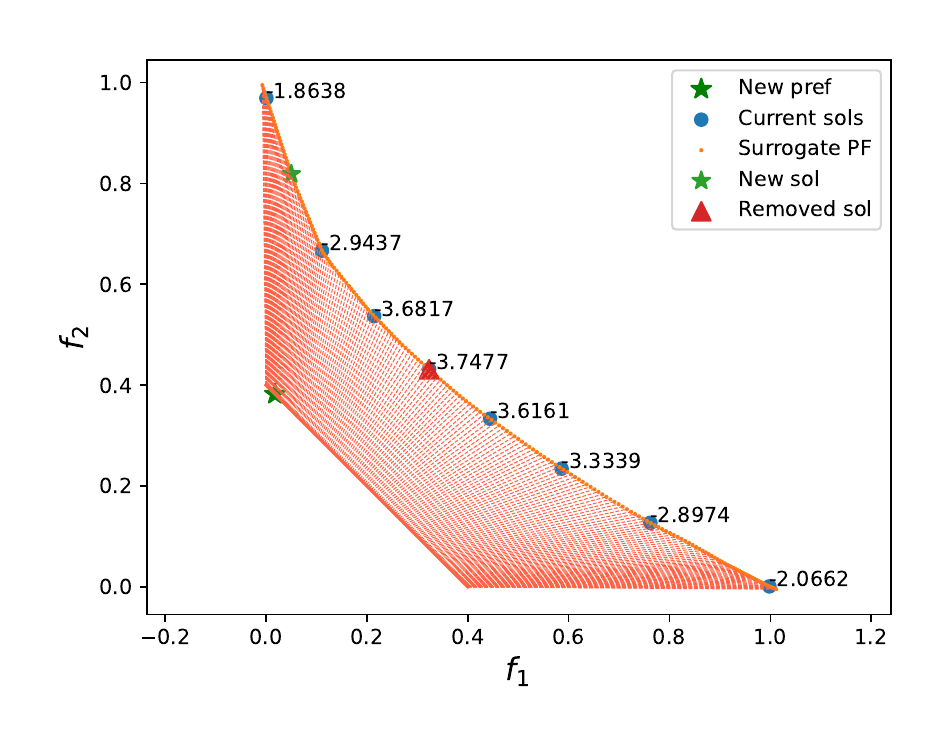}}
    \hfill
    \subfloat[DTLZ2. \label{fig_moea_epsl_b}]{\includegraphics[width = 0.22\linewidth]{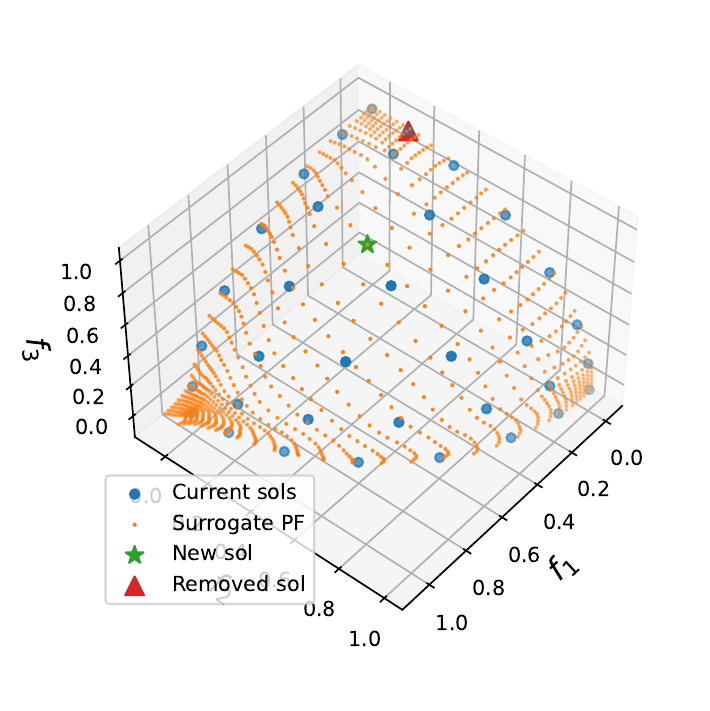}}
    \hfill
    \subfloat[RE37. \label{fig_moea_epsl_b}]{\includegraphics[width = 0.22\linewidth]{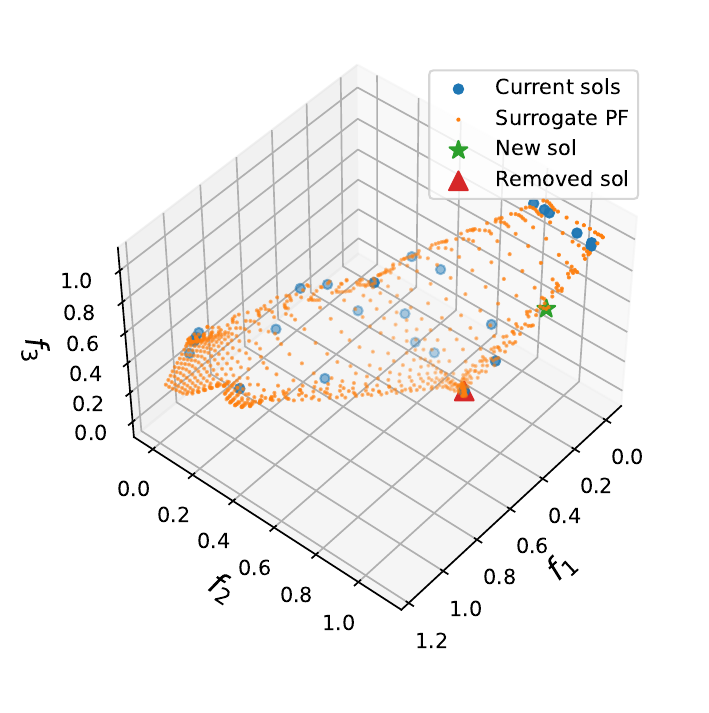 }}
    \hfill
    \caption{weight adjustment in MOEA/D-AWA.}
    \label{app:fig:awa_adjust}
\end{figure*}

The weight adjustment processes of MOEA/D-AWA \cite{qi2014moea} on two-objective ZDT1, three-objective DTLZ2/RE37 are demonstrated in \Cref{app:fig:awa_adjust}. During each update, the algorithm eliminates the most crowded objective (depicted as a red dot in \Cref{app:fig:awa_adjust}) and adds the most sparse objective (represented by a green dot). The resulting weight is indicated by a green star.
The sparsity measure is determined using Equation (4) as outlined in the original publication \cite{qi2014moea}. From Figure \ref{app:fig:awa_adjust}, it is evident that the initial objectives obtained by MOEA/D-mtche (MOEA/D with modified Tchebycheff aggregation function) are unevenly distributed on the (surrogate) PF. By contrast, MOEA/D-AWA successfully eliminates the most crowded objective in the upper-right region of the DTLZ2 problem, adding a new objective at the center of the PF. 

However, the aforementioned strategy is heuristic in nature, lacking a guarantee of achieving optimal solutions during the final adjustment phase. Another difference is MOEA/D-AWA compared with the proposed method is that it only remove and add one solution for each weight adjustment, which make it less efficient compared with the proposed method.

\subsection{Comparison with Gradient-based MOOs} \label{sec:gradmoo}
\begin{figure*}[h!]
    \centering
    \subfloat[Linear aggregation, $g(\vf(\vx), \vlam) = \sum_{i=1}^m \lambda_i f_i(\vx)$. \label{fig_moea_epsl_a}]{\includegraphics[width=0.35 \linewidth]{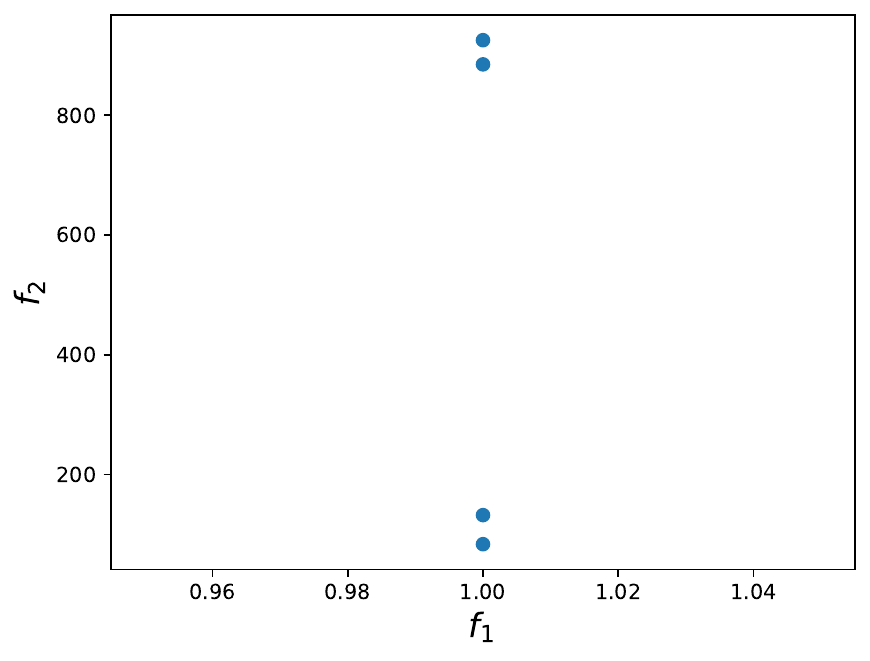}}
    \hspace{.6in}
    \subfloat[mtche aggregation, $g(\vf(\vx), \vlam) = \max_{i \in [m]} \lbr{\frac{f_i(x) - z_i}{\lambda_i}}$. \label{fig_moea_epsl_b}]{\includegraphics[width = 0.35\linewidth]{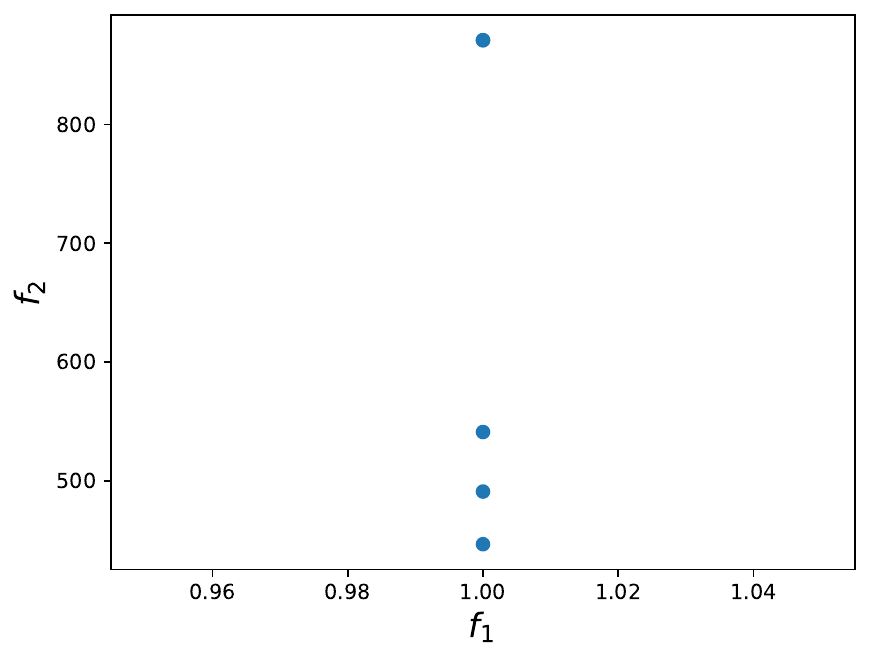}}
    \hfill
    \caption{Results of gradient-based MOOs on ZDT4.}
    \label{app:fig:gradmoo}
\end{figure*}

In this section, we show that, since a large number of MOO problems have plenty of local optimal solutions, pure gradient-based methods easily fail on these problems. We take the ZDT4 problem as an example, which has the form of,
\begin{equation}
    \left \{
    \begin{aligned}
        & f_1(\vx) = x_1 \\
        & g(\vx) = 1 + 10(n-1) + \sum_{i=2}^n ( x_i^2 - 10 \cos(4 \pi x_i)) \\
        & h(f_1(\vx), g(\vx)) = 1 - \sqrt{f_1 / g(\vx) } \\
        & f_2(\vx) = g(\vx) \cdot h(f_1(\vx), g(\vx)) \\
        & 0 \leq x_1 \leq 1, \\
        & -10 \leq x_i \leq 10, \quad i = 2,\ldots,n.
    \end{aligned}
    \right .
\end{equation}
It is clear that due to the term $\sum_{i=2}^n (x_i^2 - 10 \cos(4 \pi x_i))$, the objective function has plenty of locally optimas. Simply using gradient methods fails on this problems. The left figure in \Cref{app:fig:gradmoo} is using the simple linear aggregation function (weighted sum), where $g(\vf(\vx), \vlam) = \sum_{i=1}^m \lambda_i f_i(\vx)$. And the left figure use the modified Tchebycheff aggregation function as defined by \Cref{eqn:g-mtche}. By directly optimizing this aggregation functions by gradient descent methods, the second objective $f_2$ keeps a very large value which is far from the global optimal. 

The gradient of an aggregation function can be decomposed by the chain rule as: $\frac{\partial g}{\partial \vx} = \sum_{i=1}^m \frac{\partial g}{\partial f_i} \cdot \frac{\partial f_i}{\partial \vx}$. However, when $\|\frac{\partial f_i}{\partial \vx}\| = 0$ for a specific index $i$, the gradient information for objective $f_i$ disappears. We would also like to mention that, recently, there emerges some new gradient-based methods called `Specialized Multi-Task Optimizers' (SMTOs) \cite{hu2023revisiting}, such as MOO-SVGD \cite{liu2021profiling} or EPO \cite{mahapatra2020multi}.

We argue that these methods may not provide significant improvements for problems like ZDT4. The primary reason is that these methods require solving the gradients $\nabla f_i(\vx)$ for each objective and manipulating them using gradient manipulation techniques. However, if the norm of one of these gradient information, $\|\nabla f_i(\vx)\|$, equals zero, the resulting direction lacks any information to optimize the corresponding objective. Consequently, this objective remains trapped in a local optimum.

\section{Theoretical Results}
\label{app:sec:prof_mp_err}
In this section, we provide theoretical results of the benefits of the uniform Pareto objectives for estimating the PF model. Section \ref{app:risk} provides the preliminary tools of risk analysis which is served for Section \ref{app:sec:proof_thm2}. Section \ref{app:sec:proof_thm2} proves the main Theorem \ref{thm:generalization} in the main paper. The other sections provide the missing proofs in Section \ref{sec_41} and \ref{sec:mms}.

\subsection{Risk decomposition} \label{app:risk}
The excess risk of the prediction model, i.e., $R(\hat{\vh})$, can be decomposed in four parts according to \citet[Preface]{mjt_dlt}, 

\begin{equation}
    R(\hat{\vh}) \leq \underbrace{\abs{R(\hat{\vh}) - \hat{R}(\hat{\vh})}}_{\epsilon_1} + \underbrace{{{\hat{R}(\hat{\vh}) - \hat{R}(\overline{\vh})}}}_{\epsilon_2} + \underbrace{\abs{\hat{R}(\overline{\vh}) - R(\overline{\vh})}}_{\epsilon_3} + \underbrace{R(\overline{\vh})}_{\epsilon_4};
\end{equation}
Here $R(\cdot)$ denotes the true risk and $\hat{R}(\cdot)$ denotes the empirical risk with $N$ samples, $\hat{\vh}$ denotes the function achieved by SGD and $\bar{\vh}$ is some low-complexity approximation of $\vh_*(\cdot)$ in a function class $\mathcal{F}$ of all possible neural-networks predictors. 

The second term $\epsilon_2$ is the optimization loss, which can be optimized to global optimal in a polynomial time with the network depth and input size according to \citet[Th. 1]{allen2019convergence}. The forth term $\epsilon_4$ is the approximation error, both of which can be small due to the universal approximation theorem with an appropriately specified $\mathcal{F}$; see e.g., \citet[Preface]{mjt_dlt}.

In the next section, we give the bound of the first and third term $\epsilon_1$ and $\epsilon_3$ through controlling the more general generalization error. In particular, let $\tilde{\vh} = \overline{\vh} \; \text{or} \; \hat{\vh}$, then $\epsilon_1$ or $\epsilon_3$ can be expressed by the corresponding generalization error $\tilde{\epsilon}= |R(\tilde{\vh}) - \hat{R}(\tilde{\vh})|$. 

\subsection{Proof of Theorem \ref{thm:generalization}}
\label{app:sec:proof_thm2}
\begin{proof}
    We firstly decompose the error $\tilde{\epsilon}$ into two parts, $\varepsilon_1$ and $\varepsilon_2$, which can be formulated as,
    \begin{equation}
    \begin{split}
        \tilde{\epsilon}  & = \underbrace{\int_{\mathcal{T}} {\sbr{\norm{(\tilde{\vh} - \vh_*) \circ \vh_*^{-1}(\vy)}}}^2 d\vy}_{R(\tilde{\vh})} - \underbrace{\frac{1}{N} \sum_{i=1}^N {\sbr{\norm{(\tilde{\vh} - \vh_*) \circ \vh_*^{-1}(y_i)}}^2}}_{\hat{R}(\tilde{\vh})}, \\ & \leq \underbrace{\sum_{i=1}^N  \left ( \int_{ \mathcal{B}_i} \left ( {\sbr{\norm{(\tilde{\vh} - \vh_*) \circ \vh_*^{-1}(\vy)}}}^2 - {\sbr{\norm{(\tilde{\vh} - \vh_*) \circ \vh_*^{-1}(\vy^{(i)} )}}}^2 \right ) d\vy \right )}_{\varepsilon_1}
\\
& + \underbrace{ \sum_{i=1}^N \sbr{ \frac{\mathcal{H}_{m-1}(\mathcal{B}_i)}{\mathcal{H}_{m-1}(\mathcal{T})} - \frac{1}{N} } c_i}_{\varepsilon_2},
    \end{split}
    \label{eqn:tilde_decomp}
\end{equation}
where $\vh_*(\cdot)$ is the true mapping function from a weight angle to Pareto solution, $\vh_*^{-1}(\cdot)$ is the inverse function of $\vh_*(\cdot)$ and $c_i = \sbr{\norm{(\tilde{\vh} - \vh_*) \circ \vh_*^{-1}(\vy^{(i)})}}^2 \leq A^2$. We use notation $c$ to denote the maximal value of $c_i$, where $0<c = \max_{i \in [N]} \{ c_i\} \leq A^2$. In the second line in \Eqref{eqn:tilde_decomp}, $\mathcal{B}_i$ denotes the Voronoi cell of point $\vy^{(i)}$, where $\mathcal{B}_i = \{ \vy \; | \; \min_{\vy \in \mathcal{T}} \rho(\vy, \vy^{i}) \}$. The distance function $\rho(\cdot, \cdot)$ remain as the $\ell_2$ norm. We use $\delta_i$ to denote the diameter of set $\mathcal{B}_i$, where the formal definition of a diameter is provided in \Eqref{app:eqn:diam}. Similarly, the maximal diameter $\delta_v$ of all cells is defined as,

\begin{equation}
    \delta_v = \max_{i \in [N]} \{ \delta_i \}. 
    \label{app:eqn:delta_v}
\end{equation}
    
The proof primarily consists of two steps, where we separately bound $\varepsilon_1$ and $\varepsilon_2$. 
\paragraph{(The bound of $\varepsilon_1$)}
We first show that $\varepsilon_1$ can be bounded by the maximal diameter $\delta$ up to a constant.
\bee
& \sum_{i=1}^N \int_{ \mathcal{B}_i} {\sbr{\norm{(\tilde{\vh} - \vh_*) \circ \vh_*^{-1}(\vy)}}}^2 - {\sbr{\norm{(\tilde{\vh} - \vh_*) \circ \vh_*^{-1}(\vy^{(i)})}}}^2  d\vy
\\
= & \sum_{i=1}^N \int_{ \mathcal{B}_i} {\sbr{\norm{(\tilde{\vh} - \vh_*) \circ \vh_*^{-1}(\vy)} - \norm{(\tilde{\vh} - \vh_*) \circ \vh_*^{-1}(\vy^{(i)})}} \sbr{\norm{(\tilde{\vh} - \vh_*) \circ \vh_*^{-1}(\vy)} + \norm{(\tilde{\vh} - \vh_*) \circ \vh_*^{-1}(\vy^{(i)})}}} d\vy 
\\
\leq & \sum_{i=1}^N \int_{ \mathcal{B}_i} {\sbr{\norm{(\tilde{\vh} - \vh_*) \circ \vh_*^{-1}(\vy) - (\tilde{\vh} - \vh_*) \circ \vh_*^{-1}(\vy^{(i)})}}\sbr{\norm{(\tilde{\vh} - \vh_*) \circ \vh_*^{-1}(\vy)} + \norm{(\tilde{\vh} - \vh_*) \circ \vh_*^{-1}(\vy^{(i)})}}} d \vy 
\\
& \quad \text{(By function smoothness and  upper bound)} \\  
\leq & \sum_{i=1}^N \int_{\mathcal{B}_i} 2 A A^\prime L L^\prime \norm{\vy - \vy^{(i)} } d \vy \\
\leq & \sum_{i=1}^N 2 \mathcal{H}_{m-1} (\mathcal{B}_i) A A^\prime L L^\prime \delta_v \\ 
\leq & 2 \mathcal{H}_{m-1}(\mathcal{T}) A A^\prime L L^\prime \delta_v. 
\ee

\paragraph{(The bound of $\varepsilon_2$)} To bound $\varepsilon_2$, we introduce several distribution functions as summarized below.
\begin{itemize}
    \item $\mathcal{U} = \text{Unif}(\mathcal{T})$ denotes the uniform distribution over the PF $\mathcal{T}$. 
    \item $\widetilde{\mY}_N$ represents the category distribution over the set $\lbr{\vy^{(1)}, \ldots, \vy^{(N)}}$, where each discrete point has a probability of $\frac{1}{N}$: 
    \item $S_N$ is any distribution, that satisfies the following properties. (1) $\int_{\mathcal{B}_i} p_{S_N}(\vy) d\vy = \frac{1}{N}$ (2) $|{\partial p_{S_N}(\vy)}/{\partial \vy}|$ is zero at boundary and is bounded at other place, and (3) almost surely, the pointwise density of $S_N$ is large or smaller than the corresponding pointwise density of $\mathcal{U}$, at each $\mathcal{B}_i$. 
\end{itemize}
With the above distributions, we can bound $\varepsilon_2$ by the following derivation,
\begin{equation}
    \begin{split}
        \sum_{i=1}^N \left ( \frac{\mathcal{H}_{m-1}(\mathcal{B}_i)}{\mathcal{H}_{m-1}(\mathcal{T})} - \frac{1}{N} \right )c_i & \leq \sum_{i=1}^N {\abs{ \frac{\mathcal{H}_{m-1}(\mathcal{B}_i)}{\mathcal{H}_{m-1}(\mathcal{T})} - \frac{1}{N} }}A^2 \\
        & = C A^2\text{TV} (\mathcal{U}, S_N) \\
        & \leq C A^2 \sqrt{\mathcal{W}_1 (\mathcal{U}, S_N)} \\
        & \leq C A^2 \sqrt{\mathcal{W}_1 (\mathcal{U}, \widetilde{\mY}_N) +  \mathcal{W}_1(\widetilde{\mY}_N, S_N)} \\
        & \leq C A^2 \sqrt{\mathcal{W}_1 (\mathcal{U}, \widetilde{\mY}_N) + \delta_v}. 
    \end{split}
    \label{eqn:wdist_eqs}
\end{equation}

Here, $\mathcal{W}_1$ is the Wasserstein distance function. The second line is from the definition of the total variance (TV) distance. The third line is an adaptation of \citep[Theorem 2.1]{chae2020wasserstein} with $\alpha = 1$ therein. As shown by \citep[Theorem 2.1]{chae2020wasserstein}, $C> 0$ is determined by the Sobolev norms of $\mathcal{U}$ and $S_N$, which can be regarded as a universal constant, since $S_N$ has a smooth density function and $\mathcal{T}$ is compact. The quantity $\mathcal{W}_1(\widetilde{\mY}_N, S_N)$ can be bounded by the following expressions,
\begin{equation}
    \begin{split}
        \mathcal{W}_1(\widetilde{\mY}_N, S_N) & = \inf_{\gamma\in \Gamma} \int_{\mathcal{T}\times\mathcal{T} } \abs{\vy - \vy'} \gamma(\vy, \vy') d \vy d \vy' \\
        &\leq  \inf_{\gamma\in\Gamma}\sum_{i=1}^N  \int_{\mathcal{T}} \abs{\vy - \vy^{(i)}} \gamma(\vy, \vy^{(i)}) d \vy  \\
        & \leq  \delta_v\inf_{\gamma\in\Gamma}\sum_{i=1}^N\underbrace{\int_{\mathcal{T}} \gamma(\vy, \vy^{(i)}) d \vy}_{ = 1/N} = \delta_v.
    \end{split}
\end{equation}
Here $\Gamma$ is the set of all joint density function $\gamma$ over $\mathcal{T}\times\mathcal{T}$ such that,
$$
\sum_{i = 1}^N \gamma(\vy,\vy^{(i)}) = p_{S_N}(\vy), \quad \int_{\mathcal{T}} \gamma(\vy,\vy') d\vy = \frac{1}{N}\mathbb{I}(\vy' = y_i\text{ for some }i\in[n]),
$$ 
which also implies that $\gamma(\vy,\vy') = 0$ as long as $\vy'$ is not in $\{\vy^{(1)},\dots,\vy^{(N)}\}$.

\end{proof}

\subsection{Proof of \Cref{prop:uniform} } \label{sec:uniform_proof}
The asymptotic result follows \cite{borodachov2007asymptotics}[Theorem 2.2]. As for the non-asymptotic results, we make a proper assumption. 
\begin{ass} \label{ass:no_overlap}
    when the solution number $N_1 > N_2$, the packing distance solving \Eqref{eqn:pack_pf} is strictly decreasing, i.e., $\delta^*(N_1) < \delta^*(N_2)$.
\end{ass}

\begin{proof}
     Under this assumption, Let $N_1$ is the maximal packing number of $\delta^*(N_1)$. In such a case, the solution set $\mY_{N_1}^*$ is also a $\delta^*(N_1)$-covering. Since when it is not a $\delta^*(N_1)$-covering, then there exist a solution $\vy^\prime \in \Tau, y \neq y_i, i \in [N_1]$ such that $\rho(\vy^{(i)}, \vy^{(j)}) > \delta^*(N_1), i \neq j$, then $\mathcal{S}^\prime = \{\vy^\prime\} \cup \mY_{N_1}$ is a $(N_1+1)$ packing of $\mY_{N+1}$, which is a contradiction. 
\end{proof}

\subsection{Special Cases for Uniform weight Yielding Uniform Pareto Objectives} \label{app:sec:special}
The key is to show when the function $\vh(\vlam)$ is a constant mapping with respect to the input $\vlam$, which is a special case of affine mapping. We provide two special cases,
\begin{enumerate}
    \item The entire weight space $\vOmega$ is $\mathcal{S}^+_1$ or $\mathcal{S}^+_2$ . The objective function $f$ is ZDT2 or DTLZ2.
    \item The entire weight space $\vOmega$ is $\vDelta_2$. and the objective function $f$ is DTLZ1.
\end{enumerate}
We prove the first case as an example, and the second one can be proved similarly. 
\begin{proof}
    For each weight $\vlam = (\lambda_1, \ldots, \lambda_m), \vlam \in \mathcal{S}^{m-1}_+$, the corresponding solution $y = \vh(\vlam)$ can be expressed in the form of $(k^\prime(\vlam) \cdot \lambda_1, \ldots, k^\prime(\vlam) \cdot \lambda_m)$, since the function $\vh(\vlam)$ is an ``exact'' mapping function. Since $(k^\prime(\vlam) \cdot \lambda_1, \ldots, k^\prime(\vlam) \cdot \lambda_m)$ lies on the PF $k \cdot \mathcal{S}^{m-1}_+$, we have,
    \begin{equation}
        \left \{ \begin{split}
            & \sum_{i=1}^m \lambda_i^2 = 1, \\
            & {(k^\prime(\vlam))}^2 \sum \lambda_i^2 = k^2.
        \end{split} \right .
        \label{app:eqn:proof_prop1}
    \end{equation}
    \Eqref{app:eqn:proof_prop1} directly leads that $\vh(\vlam) = k \cdot \vlam, \forall \vlam \in \mathcal{S}_+^{m-1}$. Thus, $\vh(\vlam)$ maps an asymptotically uniform distribution up to a constant, which also is an asymptotically uniform distribution.
\end{proof}

\subsection{Proof of \Cref{thm:obj_dist}} \label{app:sec:proof_1}
\citet{blank2020generating} proposed to generate uniformly distributed weight set $\lambda_N$, by solving the following optimization problem,
\begin{equation}\label{eqn:pref_opt}
    \max_{(\lambda_N \subset \vOmega) } \min_{(i,j \in [N], i \neq j)} \rho(\vlam^{(i)}, \vlam^{(j)})
\end{equation} 
where $\vOmega$ is a compact connected set embedded of $\mathbb{R}^m$. The distance function $\rho(\cdot, \cdot)$ is adopted as the $\ell_2$ norm in this paper.  

To prove \Thref{thm:obj_dist}, we first need to introduce \Cref{lem:pref_asympt}. 
    According to \citet{borodachov2007asymptotics}, when $\vOmega$ is a rectifiable set \footnote{Any compact and connected set with a finite Hausdorff dimension is a rectifiable set. As we have assumed $\vOmega$ is compact and connected, then $\vOmega$ is rectifiable}, we have the following asymptotic uniformity of $\lambda_N$ when $\lambda_N$ solve \Probref{eqn:pref_opt},
\begin{lemma} [\cite{borodachov2007asymptotics}] \label{lem:pref_asympt}
For any fixed Borel subset $\mathcal{B}\subseteq \vOmega$, one has, when $N\rightarrow \infty$,
\begin{equation}
    \mathbb{P}\Big(\tilde{\vlam}_N \in \mathcal{B}\Big) = \frac{\mathtt{Card}(\lambda_N\cap\mathcal{B})}{\mathtt{Card}(\lambda_N)}\rightarrow \frac{\mathcal{H}_{m-1}(\mathcal{B})}{\mathcal{H}_{m-1}(\vOmega)} = \mathbb{P}\Big(\tilde{\vlam} \in \mathcal{B}\Big).    
\end{equation}
\end{lemma}
We use $\mathtt{Card}(\cdot)$ to represent the cardinality of a set. \Cref{lem:pref_asympt} directly follows from the proof of \citep[Theorem 2.2]{borodachov2007asymptotics}. Let $\tilde{\vlam}_N$ be a random variable sampled from the category distribution of the set $\lambda_N$, where each category has a probability of $\frac{1}{N}$. $\tilde{\vlam}$ is a random variable sampled from the uniform distribution on $\vOmega$, denoted as $\text{Unif}(\vOmega)$. See \Cref{app:sec:mtd:def} for more discussions about Hausdorff measure $\mathcal{H}_{m-1}(\cdot)$. 

\Lemref{lem:pref_asympt} is equivalent to say that $\tilde{\vlam}_N \xrightarrow{\text{d}} \text{Unif}(\vOmega)$. To prove Theorem \ref{thm:obj_dist}, according to the continuous mapping theorem \citep[Theorem 3.2.10]{durrett2019probability}, $\widetilde{\mY}_N$ = $\vh \circ \tilde{\vlam}_N \xrightarrow{\text{d}} \vh \circ \text{Unif}(\vOmega)$.

\end{document}